\documentclass[journal]{IEEEtran}

\usepackage[normalem]{ulem}
\usepackage{tikz,pgfplots}
\usepackage{multirow, booktabs}
\usepackage{graphicx}
\usepackage{mathtools, nccmath}
\DeclarePairedDelimiter{\nint}\lfloor\rceil
\usepackage[linesnumbered,ruled,vlined]{algorithm2e}
\usepackage{multirow}
\usepackage{longtable}
\usepackage{makecell}
\usepackage[export]{adjustbox}
\usepackage[hyphens]{url}
\usepackage{dingbat}
\usepackage{amssymb}
\usepackage{diagbox}
\usepackage{hhline}
\usepackage{booktabs}
\usepackage{pgfplotstable}
\usepackage{array,ragged2e}
\usepackage{hyperref}

\usepgfplotslibrary{fillbetween}

\usepackage{tikz,pgfplots}
\usepackage[caption=false,font=footnotesize]{subfig}
\usetikzlibrary{patterns}
\usepackage{siunitx}
\definecolor{rosso}{RGB}{220,57,18}
\definecolor{giallo}{RGB}{255,153,0}
\definecolor{blu}{RGB}{102,140,217}
\definecolor{verde}{RGB}{16,150,24}
\definecolor{viola}{RGB}{153,0,153}
\definecolor{awesome}{rgb}{1.0, 0.13, 0.32}
\definecolor{ref}{rgb}{0.65,0.65,0.65} 
\definecolor{darkgreen}{RGB}{47,109,79}
\definecolor{mydarkgreen}{rgb}{0.0, 0.39215686274509803, 0.0}
\definecolor{mygreen}{rgb}{0.0, 0.5019607843137255, 0.0}
\definecolor{darkblue}{RGB}{57,79,99}
\definecolor{verde}{RGB}{16,150,24}
\definecolor{viola}{RGB}{153,0,153}
\definecolor{americanrose}{rgb}{1.0, 0.01, 0.24}
\definecolor{bostonuniversityred}{rgb}{0.8, 0.0, 0.0}
\definecolor{chocolate(traditional)}{rgb}{0.48, 0.25, 0.0}
\definecolor{violet(web)}{rgb}{0.93, 0.51, 0.93}
\definecolor{airforceblue}{rgb}{0.36, 0.54, 0.66}

\definecolor{almond}{rgb}{0.94, 0.87, 0.8}
\definecolor{amethyst}{rgb}{0.6, 0.4, 0.8}
\definecolor{bazaar}{rgb}{0.6, 0.47, 0.48}
\definecolor{britishracinggreen}{rgb}{0.0, 0.26, 0.15}
\definecolor{byzantine}{rgb}{0.74, 0.2, 0.64}
\definecolor{cadetblue}{rgb}{0.37, 0.62, 0.63}
\definecolor{cambridgeblue}{rgb}{0.64, 0.76, 0.68}
\definecolor{candypink}{rgb}{0.89, 0.44, 0.48}
\definecolor{caputmortuum}{rgb}{0.35, 0.15, 0.13}
\definecolor{cerulean}{rgb}{0.0, 0.48, 0.65}
\definecolor{corn}{rgb}{0.98, 0.93, 0.36}
\definecolor{darkbyzantium}{rgb}{0.36, 0.22, 0.33}
\definecolor{darkgoldenrod}{rgb}{0.72, 0.53, 0.04}
\definecolor{darkseagreen}{rgb}{0.56, 0.74, 0.56}
\definecolor{darkturquoise}{rgb}{0.0, 0.81, 0.82}
\definecolor{dimgray}{rgb}{0.41, 0.41, 0.41}
\definecolor{eggplant}{rgb}{0.38, 0.25, 0.32}
\definecolor{darkorange}{rgb}{1.0, 0.55, 0.0}

\DeclareMathOperator{\sign}{sign}
\DeclareMathOperator*{\argmin}{arg\,min}
\DeclareMathOperator*{\argmax}{arg\,max}
\def\checkmark{\tikz\fill[scale=0.4](0,.35) -- (.25,0) -- (1,.7) -- (.25,.15) -- cycle;} 

\definecolor{BR}{rgb}{0,0,1}
\definecolor{MR}{RGB}{153,0,153}
\definecolor{JR}{RGB}{47,109,79}
\usepackage{colortbl}
\definecolor{R50}{RGB}{175,0,3}
\definecolor{R18}{RGB}{233,107,34}
\definecolor{AN}{RGB}{38,181,66}
\definecolor{mynicegreen}{RGB}{102,252,102}

\usepackage{ifthen}
\newboolean{showcomments}
\setboolean{showcomments}{true} 
\ifthenelse{\boolean{showcomments}}{
\newcommand{\mynote}[3]{\noindent{\color{#3}\textbf{#1:\xspace} #2}}}
{ \newcommand{\mynote}[3]{}
}

\usepackage{ulem}
\usepackage{mathtools}

\DeclarePairedDelimiter\ceil{\lceil}{\rceil}
\DeclarePairedDelimiter\floor{\lfloor}{\rfloor}

\pgfplotsset{compat=1.9}
\begin{document}

\title{Semantically Adversarial Learnable Filters}

\author{Ali Shahin Shamsabadi, Changjae Oh, Andrea Cavallaro
\thanks{This work was supported in part by the Alan Turing Institute, which is funded by the UK Engineering and Physical Sciences Research Council (EPSRC) under Grant EP/N510129/1, through the project PRIMULA. The authors are with Centre for Intelligent Sensing, Queen Mary University of London, UK. } 
\thanks{2022 IEEE. Personal use of this material is permitted. Permission from IEEE must be obtained for all other uses, in any current or future media, including reprinting/republishing this material for advertising or promotional purposes, creating new collective works, for resale or redistribution to servers or lists, or reuse of any copyrighted component of this work in other works.}
}

\maketitle

\begin{abstract}

We present an adversarial framework to craft perturbations that mislead classifiers by accounting for the image content and the semantics of the labels. The proposed framework combines a structure loss and a semantic adversarial loss in a multi-task objective function to train a fully convolutional neural network.
The structure loss helps generate perturbations whose type and magnitude are defined by a target image processing filter. 
The semantic adversarial loss considers groups of  (semantic) labels to craft perturbations that prevent the filtered image {from} being classified with a label in the same group. We validate our framework with three different target filters, namely  detail enhancement, log transformation and gamma correction filters; and evaluate the adversarially filtered images against three classifiers, ResNet50, ResNet18 and AlexNet, pre-trained on ImageNet. We show that the proposed framework generates filtered images with a high success rate, robustness, and transferability to unseen classifiers. {We also discuss objective and subjective evaluations of the adversarial perturbations.}

\end{abstract}

\IEEEpeerreviewmaketitle

\section{Introduction}

\IEEEPARstart{D}{eep} Neural Networks (DNNs) are vulnerable to  image perturbations that are crafted to cause misclassification~\cite{biggio2013evasion,szegedy2013intriguing}. These  adversarial perturbations can  be classified as norm-bounded or content-based. {\em Norm-bounded} perturbations are characterised by a limited $l_p$ distortion, with the aim of generating imperceptible changes~\cite{goodfellow2014,kurakin2016adversarial,dong2019evading,MoosaviDezfooli16,carlini2017towards,modas2018sparsefool,papernot2016limitations,Li2019}. However, this approach to craft adversarial examples limits their robustness to adversarial defences and hinder their transferability to unseen classifiers~\cite{sharif2018suitability, sen2019should}. Instead,  {\em content-based} perturbations, which  are crafted considering specific image properties, introduce unrestricted intensity changes~\cite{shamsabadi2019colorfool,shamsabadi2019edgefool,laidlaw2019functional,zhao2019towards,hosseini2018semantic,bhattad2019unrestricted}. Content-based perturbations may modify colours (SemanticAdv~\cite{hosseini2018semantic}, ColorFool~\cite{shamsabadi2019colorfool} and ACE~\cite{zhao2020ACE}) or image structures (EdgeFool~\cite{shamsabadi2019edgefool}). However, these perturbations may cause visible distortions (e.g.~unrealistic colours).

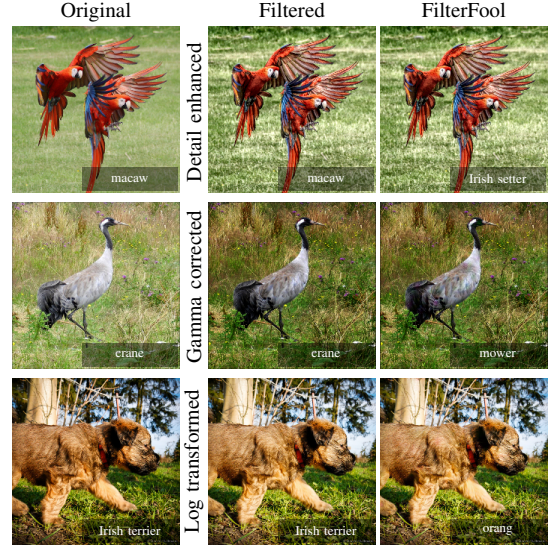
\begin{figure}[t]
    \centering
    \setlength\tabcolsep{1pt}
    \begin{tabular}{cccc}
        \footnotesize Original & &\footnotesize Filtered & \footnotesize FilterFool \\
        \begin{tikzpicture}
         \node[inner sep=0pt] (russell) at (0,0) {\includegraphics[width=0.25\columnwidth]{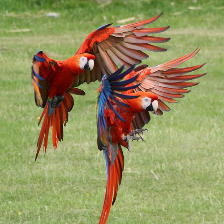}};
         \fill[fill=black, fill opacity=0.4] (-5pt,-22pt) rectangle (31pt,-31pt);  
         \node[align=right] at (.45,-0.92) {\tiny \textcolor{white}{macaw}};
         \end{tikzpicture}& 
         \raisebox{0.08\height}{\rotatebox{90}{\footnotesize Detail enhanced}}&
         \begin{tikzpicture}
         \node[inner sep=0pt] (russell) at (0,0) {\includegraphics[width=0.25\columnwidth]{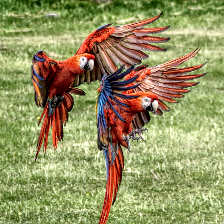}};
         \fill[fill=black, fill opacity=0.4] (-5pt,-22pt) rectangle (31pt,-31pt);  
          \node[align=right] at (.45,-0.92) {\tiny \textcolor{white}{macaw}};
         \end{tikzpicture}& 
         \begin{tikzpicture}
          \node[inner sep=0pt] (russell) at (0,0) {\includegraphics[width=0.25\columnwidth]{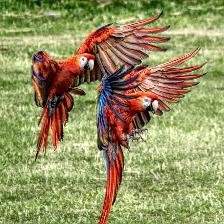}};
          \fill[fill=black, fill opacity=0.4] (-5pt,-22pt) rectangle (31pt,-31pt);  
           \node[align=right] at (.45,-0.92) {\tiny \textcolor{white}{Irish setter}};
          \end{tikzpicture} \\

         \begin{tikzpicture}
         \node[inner sep=0pt] (russell) at (0,0) {\includegraphics[width=0.25\columnwidth]{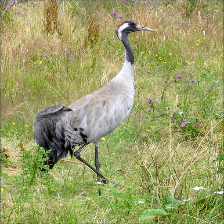}};
         \fill[fill=black, fill opacity=0.4] (-5pt,-22pt) rectangle (31pt,-31pt); 
          \node[align=right] at (.45,-0.92) {\tiny \textcolor{white}{crane}};
         \end{tikzpicture}& 
         \raisebox{0.036\height}{\rotatebox{90}{\footnotesize Gamma corrected}}&
         \begin{tikzpicture}
         \node[inner sep=0pt] (russell) at (0,0) {\includegraphics[width=0.25\columnwidth]{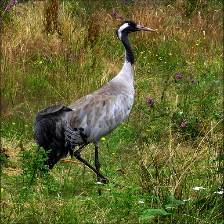}};
         \fill[fill=black, fill opacity=0.4] (-5pt,-22pt) rectangle (31pt,-31pt); 
          \node[align=right] at (.45,-0.92) {\tiny \textcolor{white}{crane}};
         \end{tikzpicture}& 
         \begin{tikzpicture}
         \node[inner sep=0pt] (russell) at (0,0) {\includegraphics[width=0.25\columnwidth]{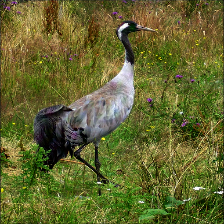}};
         \fill[fill=black, fill opacity=0.4] (-5pt,-22pt) rectangle (31pt,-31pt);  
          \node[align=right] at (.45,-0.92) {\tiny \textcolor{white}{mower}};
         \end{tikzpicture} \\
         
        \begin{tikzpicture}
         \node[inner sep=0pt] (russell) at (0,0) {\includegraphics[width=0.25\columnwidth]{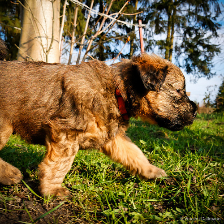}};
         \fill[fill=black, fill opacity=0.4] (-5pt,-22pt) rectangle (31pt,-31pt);  
         \node[align=right] at (.45,-0.92) {\tiny \textcolor{white}{Irish terrier}};
         \end{tikzpicture}& 
         \raisebox{0.08\height}{\rotatebox{90}{\footnotesize Log transformed}}&
         \begin{tikzpicture}
         \node[inner sep=0pt] (russell) at (0,0) {\includegraphics[width=0.25\columnwidth]{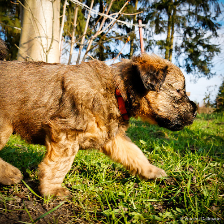}};
          \fill[fill=black, fill opacity=0.4] (-5pt,-22pt) rectangle (31pt,-31pt);
          \node[align=right] at (.45,-0.92) {\tiny \textcolor{white}{Irish terrier}};
         \end{tikzpicture}& 
         
         \begin{tikzpicture}
         \node[inner sep=0pt] (russell) at (0,0) {\includegraphics[width=0.25\columnwidth]{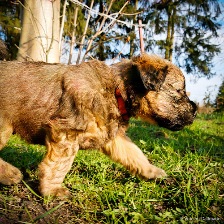}};
         \fill[fill=black, fill opacity=0.4] (-5pt,-22pt) rectangle (31pt,-31pt); 
         \node[align=right] at (.45,-0.92) {\tiny \textcolor{white}{orang}};
         \end{tikzpicture} \\
    \end{tabular}
    \caption{{Original and filtered images modified by three traditional image processing filters and by the proposed FilterFool framework.  FilterFool accounts for the image content to produce adversarial examples that mimic the target filter and mislead classifiers. The label shown with each image is the corresponding ResNet50 prediction.}}
    \label{fig:EnhAdvs}
\end{figure}

{To overcome this problem, we propose to adversarially manipulate an image while mimicking the effect of traditional image processing filters (see Figure~\ref{fig:EnhAdvs}).}
To this end, we propose an adversarial framework that learns to craft perturbations using a  multi-task objective function that combines a structure loss and a semantic adversarial loss to train, end-to-end, a fully convolutional neural network (FCNN). The structure loss supports the learning of the properties of a target image filter to control the structure of the perturbation. The semantic adversarial loss accounts for pre-defined groups of labels and prevents the classifier from predicting a label from the same group (i.e.~a semantically similar category). 
The proposed framework, FilterFool, extends our previous work~\cite{shamsabadi2019edgefool} by learning the residual of filters to generate different types of adversarial enhancement. We also introduce a semantic adversarial loss that improves the effectiveness of adversarial examples. We validate FilterFool on ImageNet~\cite{deng2009imagenet} with three classifiers, namely ResNet50, ResNet18~\cite{he2016deep} and AlexNet~\cite{krizhevsky2012imagenet} using Log transformation, Gamma correction, linear and non-linear detail enhancement as filters. The code implementing FilterFool is available at {\url{https://github.com/smartcameras/FilterFool}}.

\section{Problem Definition}
\label{sec:ProbForm}

Let $\mathbf{I}$ be an RGB image and ${\hat{y} \in \{1, ...,i, ..., D\}}$ its ground-truth class. Let a $D$-class classifier consist of a backbone of layers, {a softmax and an argmax operation}. The backbone  produces logit values for $\mathbf{I}$, ${\mathbf{z}=(z_i)_{i=1}^D}$, where ${z_{i} \in \mathbb{R}}$ is the logit value associated with class $i$. The  softmax normalises the logit values to predict the probability for all the classes, $\mathbf{p}= (p_i)_{i=1}^D$, where
\begin{equation}
p_i = \frac{{e}^{z_i}}{\Sigma_{d=1}^D {e}^{z_d}} \in [0,1]  
\end{equation}
represents the probability of $\mathbf{I}$ being associated with class $i$ and $\sum_{i=1}^Dp_i=1$. The  argmax predicts the  class $y \in \{1,...,D\}$ as:
\begin{equation}
     y = \argmax_{i=1,...,D}p_i.
\end{equation}
Note that $y \neq\hat{y}$ when the prediction of the classifier is {incorrect}.

An adversarial perturbation, $\boldsymbol{\delta}$, modifies $\mathbf{I}$ to generate an adversarial example, $\mathbf{\dot{I}}$, as $\dot{\mathbf{I}} = \mathbf{I} + \boldsymbol{\delta}$
so that its (adversarial) class, $\dot{y} \in \{1,...,D\}$, assigned by the classifier, differs from that of $\mathbf{I}$: $\dot{y}\neq y$.
Traditional adversarial examples  flip the categorical label of an image disregarding the semantics of the labels~\cite{mopuri2020adversarial}. However, labels may be synonymous (e.g.~screen and television), highly granular (e.g.~the ImageNet dataset~\cite{deng2009imagenet} includes labels for 130 breeds of dogs)~\cite{tsipras2020imagenet}, or may belong to the same semantic class (e.g.~the labels of the Scene365 dataset can be divided into two groups, namely private and non-private scene~\cite{Mediaeval2018}). 

Our goal is to group the $D$ categorical labels into $S$ classes and to prevent an attack from selecting a label within the same class.
Let $\mathbf{W}\in \{0,1\}^{D \times S}$ be a matrix that identifies categorical labels belonging to the same class $s$. In this paper, we focus on the $D=1000$ ImageNet labels and consider the $S=11$ classes generated through WordNet~\cite{tsipras2020imagenet}. These classes are dogs (containing 130 labels); other mammals (88 labels); birds (59 labels); reptiles, fish, amphibians (60 labels); invertebrates (61 labels); food, plants, fungi (63 labels); devices (172 labels); structures, furnishing (90 labels); clothes, covering (92 labels); implements, containers, misc.~objects (117 labels); and vehicles (68 labels). Other scenarios include the $D=365$ Scene365 labels and the $S=2$ classes defining private (60 labels) or non-private (305 labels) PixelPrivacy scenes~\cite{Mediaeval2018}; and the $D=30$ Cityscape labels and $S=8$ classes defined for urban  scenes: flat (4 labels), construction (6 labels), nature (2 labels), vehicle (8 labels), sky (1 label), object (4 labels), human (2 labels), and void (3 labels)~\cite{cordts2016cityscapes}.  

Adversarial attacks can be untargeted or targeted. The  perturbation of an {\em untargeted} attack is crafted to simply evade the original label, whereas the perturbation of a {\em targeted} attack is crafted to induce the prediction of a {\em specific} label, $\dot{y}_t\neq y$. In this paper, we focus on untargeted attacks, which are normally more efficient~\cite{carlini2017towards} than targeted attacks, {which generally require} larger distortions and longer time to craft a perturbation that reaches the desired (adversarial) label~\cite{kwon2020classification}. 

\section{Background}

In this section, we review norm-bounded attacks (FGSM~\cite{goodfellow2014}, BIM~\cite{kurakin2016adversarial}, CW~\cite{carlini2017towards}, DeepFool~\cite{MoosaviDezfooli16}, SparseFool~\cite{modas2018sparsefool}, RP-FGSM~\cite{sanchez2020exploiting}) and content-based attacks that consider colour (SemanticAdv~\cite{hosseini2018semantic} , ColorFool~\cite{shamsabadi2019colorfool}, ACE~\cite{zhao2020ACE}) and structure (EdgeFool~\cite{shamsabadi2019edgefool}).

FGSM~\cite{goodfellow2014} determines whether to increase or decrease the value of each pixel of $\mathbf{{I}}$ by defining $\boldsymbol{\delta} = \epsilon\sign\left(\nabla_{\mathbf{I}}J\left(\mathbf{I},y\right)\right)$, with a small $\epsilon\in \mathbb{N}$ based on the sign of $\nabla_{\mathbf{I}} J(\cdot)$, the gradient of the loss function with respect to  $\mathbf{I}$:

\begin{equation}
    \mathbf{\dot{I}}_{\mbox{\scriptsize FGSM}} =   \mathbf{I} \pm \boldsymbol{\delta}
\end{equation}

The adversarial image generated by FGSM, $\mathbf{\dot{I}}_{\mbox{\scriptsize FGSM}}$, within the $\epsilon$-neighbourhood of $\mathbf{I}$, can be the untargeted attack when $\mathbf{\dot{I}}_{\mbox{\scriptsize FGSM}}=\mathbf{I} +\boldsymbol{\delta}$ or can be the targeted attack when $\mathbf{\dot{I}}_{\mbox{\scriptsize FGSM}} = \mathbf{I} - \boldsymbol{\delta}$ with selecting $y$ as a specific label. BIM~\cite{kurakin2016adversarial} extends FGSM by iteratively generating the adversarial perturbation by aggregating ${N=\nint{\min(\epsilon+4,1.25\epsilon)}}$, where $\nint{\cdot}$ is a rounding operation to the nearest integer, perturbations $\boldsymbol{\delta}_n$ as

\begin{equation}
\label{eq:BIM}
    \mathbf{\dot{I}}_{\mbox{\scriptsize BIM}} = \mathcal{C}_{\mathbf{I},\epsilon}\left({\mathbf{\dot{I}}_N}\right),
 \end{equation}
where  $\mathcal{C}_{\mathbf{I},\epsilon}(\cdot)$ clips the pixel intensities of the adversarial image, ${\mathbf{\dot{I}}_N}= \mathbf{I} \pm  \sum_{n=1}^{N}\boldsymbol{\delta}_n$, with respect to three constant images, $\mathbf{E}$, $\mathbf{0}$ and $\mathbf{255}$, whose pixel intensities are $\epsilon$, $0$ and $255$, respectively:
\begin{equation}
\mathcal{C}_{\mathbf{I},\epsilon}(\mathbf{\dot{I}}_N)=\min\{\mathbf{255}, \mathbf{I}+\mathbf{E}, \max\{\mathbf{0}, \mathbf{I}-\mathbf{E},\mathbf{\dot{I}}_N\}\},
\end{equation}
and
\begin{equation}
    \boldsymbol{\delta}_n = \sign\left(\nabla_{\mathbf{I}+\boldsymbol{\delta}_{n-1}}J\left(\mathbf{I}+\boldsymbol{\delta}_{n-1},y\right)\right).
\end{equation}
Similar to FGSM, BIM can be a untargeted attack by {$ \mathbf{\dot{I}}_N = \mathbf{I} + \sum_{n=1}^{N}\boldsymbol{\delta}_n$} or an targeted attack by {$\mathbf{\dot{I}}_N = \mathbf{I} - \sum_{n=1}^{N}\boldsymbol{\delta}_n$} with selecting $y$ as a specific label. The iterative process helps BIM exploit finer perturbations that improve, compared to FGSM, the ability of the attack to mislead the classifier. 

CW~\cite{carlini2017towards} minimises the $l_0$, $l_2$ or $l_\infty$ norm of the difference between the original  image, ${\mathbf{I}}$, and the adversarial image, $\mathbf{\dot{I}}$, and the difference between the logit value, $\dot{z}_{y}$, of $\mathbf{\dot{I}}$ belonging to the same class as ${\mathbf{I}}$ and the maximum logit value among all the other classes:  
\begin{equation}
\label{eq:cw}
\mathbf{\dot{I}}_{\mbox{\scriptsize CW}}=\argmin_{\mathbf{\dot{I}}} \big( \|\mathbf{\dot{I}}-\mathbf{I}\|_p + c_{\mathbf{I}}(\dot{z}_{y} - \max_{i=1,...,D}\{{\dot{z}_i}; i\neq y\})\big), 
\end{equation}
where $p \in \{0,2,\infty\}$ and $c_{\mathbf{I}}>0$ is a constant selected via line search which makes the computations expensive~\cite{rony2019decoupling, yao2019trust}.

DeepFool~\cite{MoosaviDezfooli16} iteratively generates an adversarial perturbation whose $l_2$ norm is bounded. The adversarial perturbation at each iteration is the orthogonal projection of the adversarial image from the previous iteration onto the closest linearised boundary of class $y$ in the decision space. SparseFool~\cite{modas2018sparsefool} combines the DeepFool adversarial approach and the low mean curvature of DNNs in the neighbourhood of each image to perturb only a few pixels, thus resulting in sparse adversarial perturbations with a small $l_1$ norm.

Private FGSM (P-FGSM)~\cite{Li2019}, and its extension to multiple classifier and seen defences RP-FGSM~\cite{sanchez2020exploiting}, is an iterative norm-bounded attack that uses semantically pre-defined labels {({$S=2$})}. P-FGSM  follows the BIM's (Eq.~\ref{eq:BIM}) strategy to produce adversarial images  considering private and non-private groups of labels as pre-defined semantics of the categorical labels. P-FGSM chooses the target class randomly from a subset of classes defined based on the prediction probability sorted in a descending order, $\mathbf{p}'$, and a threshold on the cumulative probability, $\tau_p \in [0,1]$:
\begin{equation}
    \label{eq:P-BIM}
    y_{t}= R\left( \{ y_{j+1} : \sum_{i=1}^{j} {p'}_i > \tau_p, j \in \{1, ..., D-1\} \}\right),
\end{equation}
where $R(\cdot)$  randomly chooses a class from the set of non-private classes whose cumulative probability exceeds $\tau_p$.

SemanticAdv~\cite{hosseini2018semantic} generates adversarially colourised images in the HSV colour space by adding to the hue and saturation of $\mathbf{I}$ a random perturbation chosen uniformly from the range of valid values. SemanticAdv draws new perturbations until the classifier is misled (up to 1000 attempts). The SemanticAdv adversarial image, $\dot{\mathbf{I}}_{\text{SA}}$, is: 
\begin{equation}
    \dot{\mathbf{I}}_{\text{\scriptsize SA}}=\beta^{-1}\Big([\mathbf{I}_H+\boldsymbol{\delta}_H, \mathbf{I}_S+\boldsymbol{\delta}_S, \mathbf{I}_V]\Big),
\end{equation}
where $\beta(\cdot)$ converts $\mathbf{I}$ to its hue, $\mathbf{I}_H$, saturation, $\mathbf{I}_S$, and value, $\mathbf{I}_V$ components; $[\cdot,\cdot,\cdot]$ represents the channel-wise concatenation; and $\boldsymbol{\delta}_H $ and $\boldsymbol{\delta}_S$, where each component ranges $[0,1]$, are the final perturbations on the hue and saturation channels, respectively. Because $\mathbf{I}_H$ and  $\mathbf{I}_S$ are changed by the same amount, the colours of $\dot{\mathbf{I}}_{\text{\scriptsize SA}}$ may look unnatural.

ColorFool~\cite{shamsabadi2019colorfool} improves the naturalness of adversarial images compared to SemanticAdv~\cite{hosseini2018semantic} by  identifying  non-sensitive and sensitive regions through semantic segmentation. The  perturbations operate on the $a$ and $b$ channels of the  $Lab$ colour space: the perturbations of $N_{\bar{S}}$ non-sensitive regions are drawn randomly from the whole range of possible values, whereas the perturbations of $N_{S}$ sensitive regions are chosen randomly from pre-defined natural-colour ranges, defined based on human perception.  The ColorFool adversarial image, $\dot{\mathbf{I}}_{\text{CF}}$, is: 
\begin{equation}
    \dot{\mathbf{I}}_{\text{\scriptsize CF}}= \rho^{-1}\Big(\sum_{t=1}^{N_{S}}\big(\rho({\mathbf{S}_t})+\mathbf{N}_t\big) + \sum_{t=1}^{N_{\bar{S}}}\big(\rho({{\mathbf{\bar{S}}}_t})+\bar{\mathbf{N}}_t\big)\Big),
\end{equation}
where $\rho(\cdot)$ is the $RGB$-to-$Lab$ colour-space conversion and  $\mathbf{N}_t$ and $\bar{\mathbf{N}}_t$ are the colour perturbations of the $t$-th sensitive, $\mathbf{S}_t$, and non-sensitive, ${\mathbf{\bar{S}}}_t$, region, respectively. 

Unlike SemanticAdv and ColorFool that randomly change the colour, Adversarial Colour Enhancement (ACE)~\cite{zhao2020ACE} produces colourised adversarial images using a piecewise-linear colour adjustment filter, $\Gamma_{\Omega}(\cdot)$, where $\Omega$ represents the parameters of the filter, with $K$ pieces. Each pixel of the ACE adversarial image, $\dot{\mathbf{I}}_{\text{ACE}}$, is obtained by filtering  the original image, $I_x$, as follows:
 \begin{equation}
     \dot{I}_{\text{ACE},x}= \Gamma_{\Omega}(I_x) = \sum_{i=1}^{\floor{\frac{I_x}{s}}}\Omega_i+\frac{I_x (\mbox{mod}s)}{s}.\Omega_{\ceil{\frac{I_x}{s}}},
 \end{equation}
where $s$ is the size of each piece. ACE learns the parameters of the filter, $\Omega$, by optimising the combination of the CW loss function~\cite{carlini2017towards}, a colour adjustment constraint on the distance between each parameter $\Omega_i$ and its initial value $\frac{1}{K}$, corresponding to the unchanged image:
 \begin{equation}
\argmin_{\Omega} \big((\dot{z}_{y} - \max_{i=1,...,D}\{{\dot{z}_i}; i\neq y\})+\nu\sum_{i=1}^{\kappa}(\Omega_i-\frac{1}{K})^2\big), 
\end{equation}
where $\nu$ and $\kappa$ are a balance factor and size of $\Omega$, respectively. 
\begin{figure}[t]
    \centering
    \setlength\tabcolsep{1pt}
    \begin{tabular}{cccc}
        \scriptsize Original &\scriptsize BIM & \scriptsize DeepFool & \scriptsize SparseFool \\
         \includegraphics[width=0.23\columnwidth]{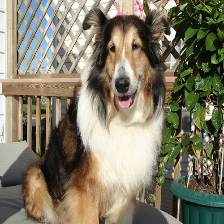}&
         \includegraphics[width=0.23\columnwidth]{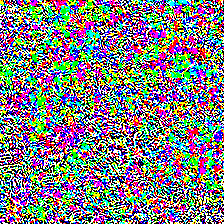}&
         \includegraphics[width=0.23\columnwidth]{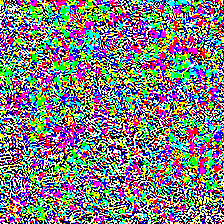}&
         \includegraphics[width=0.23\columnwidth]{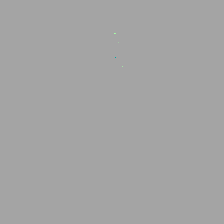}\\
         
         \scriptsize SemanticAdv & \scriptsize ColorFool & \scriptsize EdgeFool & \scriptsize FilterFool (LT)\\
         \includegraphics[width=0.23\columnwidth]{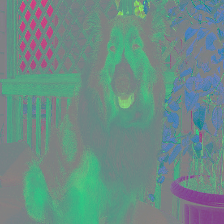}&
         \includegraphics[width=0.23\columnwidth]{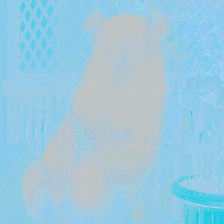}&
         \includegraphics[width=0.23\columnwidth]{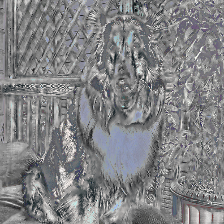}&
         \includegraphics[width=0.23\columnwidth]{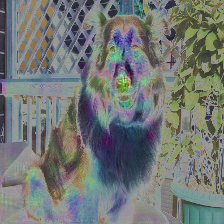}
    \end{tabular}
    \caption{Comparison of perturbations generated by selected adversarial methods. To facilitate visualisation, the values are scaled between 0 and 255. KEY -- BIM: Basic Iterative Method; LT:  log transformation.}
    \label{fig:VisPert}
\end{figure}

EdgeFool~\cite{shamsabadi2019edgefool} generates adversarial perturbations that enhance the image details. EdgeFool trains a FCNN, $R_{\theta_{\mbox{\scriptsize EF}}}(\cdot)$, where $\theta_{\mbox{\scriptsize EF}}$ represents all the parameters defining the FCNN, with a multi-task loss function:
\begin{equation}
\label{eq:EF}
\theta^*_{\mbox{\scriptsize EF}}=\argmin_{\theta_{\mbox{\scriptsize EF}}} \big( \underbrace{\|R_{\theta_{\mbox{\scriptsize EF}}}(\mathbf{I})-\mathbf{I}_g\|_2}_{\mathcal{L}_s(\cdot)} + c(\dot{z}_{y} - \max\{{\dot{z}_i};i\neq y\})\big),
\end{equation}
\begin{table}[t!]
\centering
\caption{Comparison of adversarial attacks in terms of information used to craft the perturbations ($l_p$ norm for bounded-perturbations; Col.: colour information; Obj.: information about objects in the scene; Str.: information about structures in the image) and consideration of semantic (Sem.) relationships between classes.} 
\setlength{\tabcolsep}{2pt}
\begin{tabular}{clcccccccc}
\Xhline{3\arrayrulewidth}
Ref. & Method   &  $l_0$ & $l_1$ & $l_2$ & $l_{\infty}$ & Col. & Obj. & Str. & Sem. \\
\Xhline{3\arrayrulewidth}
\cite{goodfellow2014}          & FGSM       & & & & \checkmark   & \cellcolor{gray!25} & \cellcolor{gray!25} & \cellcolor{gray!25} & \\
\cite{kurakin2016adversarial}  & BIM        & & & & \checkmark   & \cellcolor{gray!25} & \cellcolor{gray!25} & \cellcolor{gray!25} & \\
\cite{MoosaviDezfooli16}       & DeepFool   & & & \checkmark    & &\cellcolor{gray!25} & \cellcolor{gray!25} & \cellcolor{gray!25} &\\ 
\cite{modas2018sparsefool}     & SparseFool & & \checkmark &   & & \cellcolor{gray!25} & \cellcolor{gray!25} & \cellcolor{gray!25} & \\
\cite{carlini2017towards}      & CW         & \checkmark & & \checkmark  & \checkmark & \cellcolor{gray!25} & \cellcolor{gray!25} & \cellcolor{gray!25} & \\
\cite{sanchez2020exploiting}    & {RP-FGSM}   &  & &   & \checkmark & \cellcolor{gray!25} &  \cellcolor{gray!25} & \cellcolor{gray!25} & \checkmark \\
\hline
\cite{hosseini2018semantic}    & SemAdv     & \cellcolor{gray!25} & \cellcolor{gray!25} & \cellcolor{gray!25}  & \cellcolor{gray!25} &\checkmark & &  & \\
\cite{shamsabadi2019colorfool} & ColorFool  &\cellcolor{gray!25} & \cellcolor{gray!25} &\cellcolor{gray!25}  &  \cellcolor{gray!25} &\checkmark & \checkmark &  & \\
\cite{zhao2020ACE}             & {ACE}        &\cellcolor{gray!25} & \cellcolor{gray!25} &\cellcolor{gray!25}  &  \cellcolor{gray!25} &\checkmark &  &  &  \\
\cite{shamsabadi2019edgefool}  & EdgeFool   & \cellcolor{gray!25}& \cellcolor{gray!25} & \cellcolor{gray!25} & \cellcolor{gray!25}  & & & \checkmark & \\
                               & {\bf FilterFool}   &\cellcolor{gray!25} & \cellcolor{gray!25} & \cellcolor{gray!25} & \cellcolor{gray!25} & &  & \checkmark& \checkmark\\
\Xhline{3\arrayrulewidth}
\end{tabular}
\vspace{-9pt}
\label{tab:RelatedWork}
\end{table}

\noindent
where $\mathcal{L}_s(\cdot)$, the smoothing loss function, quantifies the difference between $\mathbf{I}_g$, the output of a smoothing filter, and  $\mathbf{I}_s=R_{\theta_{\mbox{\scriptsize EF}}}(\mathbf{I})$, the output of the FCNN. The adversarial perturbation operates on the $L$ channel of the  $Lab$ colour space by enhancing the image details, $\mathbf{I}_d=\mathbf{I}-\mathbf{I}_s$, using the sigmoid function, $f(a,b)=\left(1+e^{-ab}\right)^{-1}-0.5$~\cite{fan2018image}:
\begin{equation}
\dot{\mathbf{I}}^L = \Big(100f\left(\frac{\mathbf{I}^L_s-v_1}{100},v_2\right)+v_1\Big) + 100f\left(\frac{\mathbf{I}^L_d}{100},v_3\right),
\label{eq:EFImg2}
\end{equation}
where the input to the sigmoid is normalised by the maximum value of the $L$ channel (i.e.~100), and $v_1$, $v_2$ and $v_3$ are constants that adjust the midpoint and slope of the sigmoid. The second term in Eq.~\ref{eq:EF} guides the FCNN to smooth the image in a way that the EdgeFool adversarial image, $\dot{\mathbf{I}}_{\text{EF}}$:
\begin{equation}
\mathbf{\dot{I}}_{\text{\scriptsize EF}} =
\rho^{-1}\Big( [\dot{\mathbf{I}}^L, \mathbf{I}^a,\mathbf{I}^b]
\Big)
\label{eq:EFImg}
\end{equation}
causes misclassification.

Figure~\ref{fig:VisPert} compares sample perturbations generated by norm-bounded and content-based methods. It is possible to notice that the adversarial perturbations of BIM, DeepFool and SparseFool are unrelated to the content, whereas SemanticAdv modifies the colours of the whole image, ColorFool focuses on specific regions identified through semantic segmentation, and EdgeFool and FilterFool craft perturbations that relate to the structures in the image. 

Finally, Table~\ref{tab:RelatedWork} summarises the adversarial attacks discussed in this section and compares them with the proposed framework, FilterFool, which is detailed in the next section. 

\section{FilterFool}

\subsection{Learning framework}

We aim to generate adversarial perturbations that mimic the output of an image processing filter. To this end, we approximate the target filter with a DNN~\cite{chen2017fast} and, specifically, with a Fully Convolutional Neural Network (FCNN), which learns to produce perturbations that resemble that of an image filter. The parameters of the FCNN are optimised with a multi-objective loss: the {\em structure loss}, which accounts for the difference between the intensity changes introduced by the filter and the learned perturbation, and an {\em adversarial loss}, which induces misclassification by operating on pre-defined groups of labels. 

\begin{figure}[t]
    \centering
    \includegraphics[width=\columnwidth]{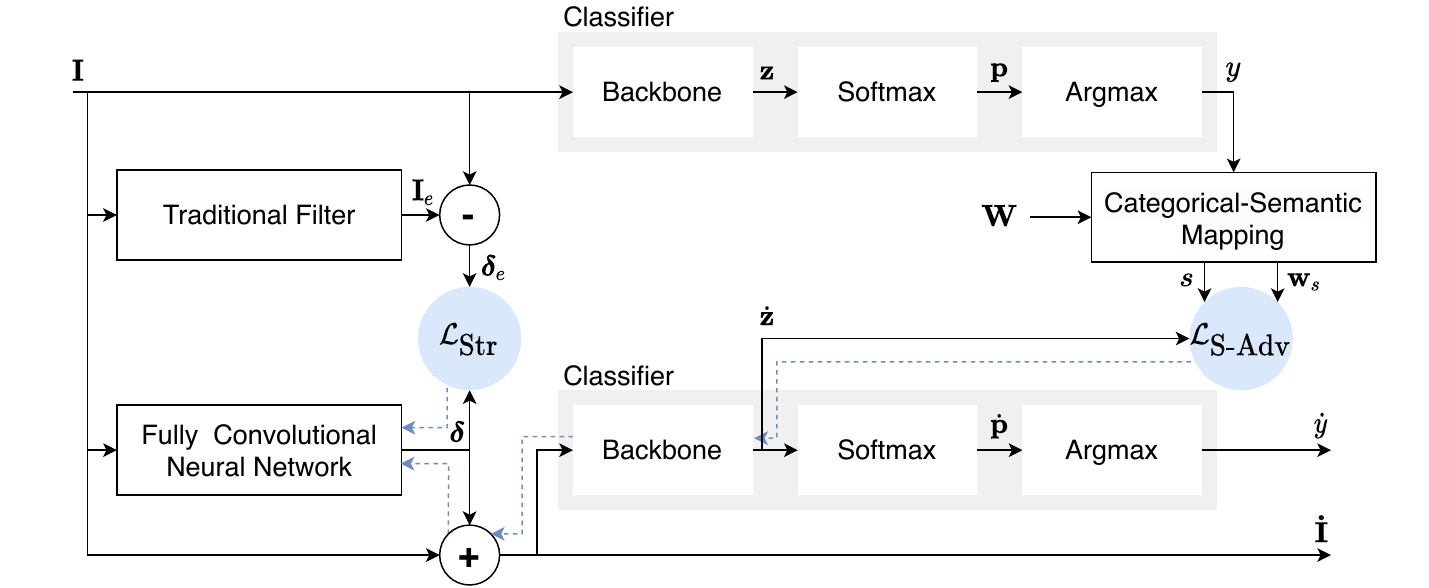}
    \caption{{Block diagram of the proposed framework, FilterFool, which integrates a traditional image processing filter with a Fully Convolutional Neural Network (FCNN) to output adversarial images that mislead a classifier with (adversarial) classes $\dot{y}$ that are semantically different  from the original classes (i.e.~ class predicted by the classifier for the original image, $y$). During training, the errors measured by  $\mathcal{L}_{\text{Str}}$ and $\mathcal{L}_{\text{S-Adv}}$ are backpropagated through the path indicated with  {\protect\raisebox{2pt}{\protect\tikz \protect\draw[black,line width=0.1mm,dotted] (0,0) -- (0.3,0);}} to optimise the parameters of FCNN. The Backbone represents all parts of the classifier before the Softmax layer. KEY -- $\mathbf{I}$ input image; $\mathbf{I}_e$: filtered image; $\boldsymbol{\delta}$: adversarial perturbation;   $\mathcal{L}_{\text{Str}}$: structure loss, which is the distance between the enhancement perturbation, $\boldsymbol{\delta}_e=\mathbf{I}_e-\mathbf{I}$, and the learned adversarial perturbation, $\boldsymbol{\delta}$;  $\mathbf{W}$ matrix that defines the mapping from categorical to semantic classes; $\mathcal{L}_{\text{S-Adv}}$: semantic adversarial loss; $\dot{\mathbf{I}}$: adversarial image.}}
    \label{fig:BlockDiagram}
\end{figure}

Figure~\ref{fig:BlockDiagram} shows the block diagram of FilterFool, a general framework to learn to generate adversarial images that resemble those obtained with an image processing filter. Given an image $\mathbf{I}$, the target filter produces $\mathbf{I}_e$. The FCNN learns an adversarial perturbation, $\boldsymbol{\delta}$, by optimising the structure loss, $\mathcal{L}_{\text{Str}}(\cdot,\cdot)$, which represents the error between the intensity changes produced by the target filter, $\boldsymbol{\delta}_e=\mathbf{I}_e-\mathbf{I}$, and the learned adversarial perturbation, $\boldsymbol{\delta}$. The semantic adversarial loss, $\mathcal{L}_{\text{S-Adv}}(\cdot,\cdot)$, causes the adversarial image, $\dot{\mathbf{I}}$, to be misclassified as class $\dot{y}$ that is not only categorically ($y \neq \dot{y}$) but also semantically different from that of the original class, $y$. The errors measured by $\mathcal{L}_{\text{Str}}(\cdot,\cdot)$ and $\mathcal{L}_{\text{S-Adv}}(\cdot,\cdot)$ are backpropagated to determine the parameters of the FCNN. 

In the rest of this section, we describe the details of structure loss $\mathcal{L}_{\text{Str}}$, the semantic adversarial loss $\mathcal{L}_{\text{S-Adv}}$  and the combined multi-task loss used for training the FCNN.
\subsection{Structure loss}
We use residual learning to  generate intensity changes for the original image, $\mathbf{I}$, to produce the desired filtered image, ${\mathbf{I}_e}$. 
We measure the difference between the residual, ${\boldsymbol{\delta}_e=\mathbf{I}_e-\mathbf{I}}$,  and the adversarial perturbation, $\boldsymbol{\delta}$, which is the output of the FCNN. To tailor the perturbation towards the output of the target filter, we define $\mathcal{L}_{l_2}(\cdot,\cdot)$, which penalises the squared error between $\boldsymbol{\delta}$ and $\boldsymbol{\delta}_e$:
\begin{equation}
\label{eq:loss_l2}
\mathcal{L}_{l_2}(\boldsymbol{\delta},\boldsymbol{\delta}_e )=
\|{ {\boldsymbol{\delta}-\boldsymbol{\delta}_e}}\|^2.
\end{equation}
\begin{figure}[t!]
    \centering
    \setlength\tabcolsep{1pt}
    \begin{tabular}{cccc}
        \scriptsize Original  &\scriptsize GC, $\gamma=0.5$ & \scriptsize  GC, $\gamma=1.5$& \scriptsize LT\\
         \includegraphics[width=0.18\columnwidth]{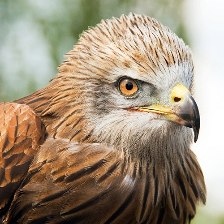}&
         \includegraphics[width=0.18\columnwidth]{ 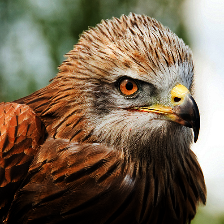}&
         \includegraphics[width=0.18\columnwidth]{ 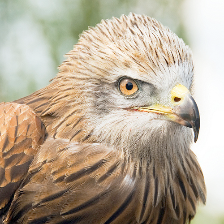}&
         \includegraphics[width=0.18\columnwidth]{ 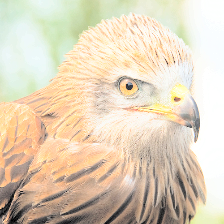}\\
        &         \scriptsize LD, $\alpha=0.1$& \scriptsize LD, $\alpha=0.5$& \scriptsize ND\\
         &
         \includegraphics[width=0.18\columnwidth]{ 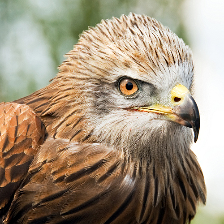}&
         \includegraphics[width=0.18\columnwidth]{ 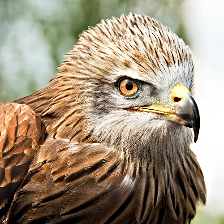}&
         \includegraphics[width=0.18\columnwidth]{ 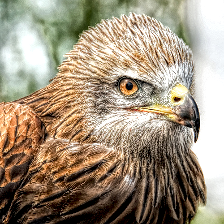}
    \end{tabular}
\caption{Sample images filtered by gamma correction (GC,  with strength $\gamma$), log transformation (LT), linear detail enhancement (LD, with strength $\alpha$) and nonlinear detail enhancement (ND). }
            \label{fig:FilterVisu}
\end{figure}
As using $\mathcal{L}_{l_2}(\cdot,\cdot)$ alone may cause artefacts in untextured regions~\cite{7797130}, we  consider another loss that accounts for structures in the image: 
\begin{equation}
    \mathcal{L}_{\text{SSIM}}(\boldsymbol{\delta},\boldsymbol{\delta}_e )=1-\text{SSIM}(\boldsymbol{\delta},\boldsymbol{\delta}_e),
\end{equation}
which is based on the Structural SIMilarity (SSIM) index~\cite{wang2004image}: 
\begin{equation}
\label{eq:loss_SSIM}
 \text{SSIM}(\boldsymbol{\delta},\boldsymbol{\delta}_e )= l(\boldsymbol{\delta},\boldsymbol{\delta}_e)\cdot c(\boldsymbol{\delta},\boldsymbol{\delta}_e)\cdot s(\boldsymbol{\delta},\boldsymbol{\delta}_e),
\end{equation}
where $l(\cdot,\cdot)$ is a function of the means:
\begin{equation}
    l(\boldsymbol{\delta},\boldsymbol{\delta}_e)=\frac{2\mu_{\boldsymbol{\delta}}\mu_{\boldsymbol{\delta}_e}+c_1}{\mu_{\boldsymbol{\delta}}^2+\mu_{\boldsymbol{\delta}_e}^2+c_1},
\end{equation}
where $\mu_{\boldsymbol{\delta}}=\frac{1}{M}\sum_{i=1}^{M}\delta_i$ and $\delta_i$ is the intensity of $i$-th element of the image (including the three colour planes) and hence $M$ is three times the resolution of the image; $c(\cdot,\cdot)$ is based on the standard deviations:
\begin{equation}
    c(\boldsymbol{\delta},\boldsymbol{\delta}_e)=\frac{2\sigma_{\boldsymbol{\delta}}\sigma_{\boldsymbol{\delta}_e}+c_2}{\sigma_{\boldsymbol{\delta}}^2+\sigma_{\boldsymbol{\delta}_e}^2+c_2},
\end{equation}
where $\sigma_{\boldsymbol{\delta}}=\big(\frac{1}{M-1}\sum_{i=1}^{M}(\delta_i-\mu_{\boldsymbol{\delta}})^2\big)^{1/2}$. The structure information, $s(\cdot,\cdot)$, is estimated based on the covariance of the perturbations: 
\begin{equation}
    s(\boldsymbol{\delta},\boldsymbol{\delta}_e)=\frac{\sigma_{\boldsymbol{\delta}\boldsymbol{\delta}_e}+c_3}{\sigma_{\boldsymbol{\delta}}\sigma_{\boldsymbol{\delta}_e}+c_3},
\end{equation}
where $\sigma_{\boldsymbol{\delta}\boldsymbol{\delta}_e}=\frac{1}{M-1}\sum_{i=1}^{M}(\delta_i-\mu_{\boldsymbol{\delta}})(\delta_i-\mu_{\boldsymbol{\delta_e}})$; and 
$c_1$, $c_2$ and $c_3$ are small constants that stabilise the division. ${\text{SSIM}(\boldsymbol{\delta},\boldsymbol{\delta}_e )\in[0,1]}$, and the closer to 1, the higher the similarity. 

The structure loss, $\mathcal{L}_{\text{Str}}(\cdot, \cdot)$, combines the two losses  as
\begin{equation}
\label{eq:loss_enh}
\mathcal{L}_{\text{Str}} (\boldsymbol{\delta},\boldsymbol{\delta}_e )=
\mathcal{L}_{l_2} (\boldsymbol{\delta},\boldsymbol{\delta}_e ) + \eta \mathcal{L}_{\text{SSIM}} (\boldsymbol{\delta},\boldsymbol{\delta}_e ),
\end{equation}
where the hyper-parameter $\eta$ is determined empirically. 

For the specific implementation of this paper, we consider four filters, namely linear and non-linear detail enhancement, log transformation and gamma correction (see Figure~\ref{fig:FilterVisu}).
For linear detail enhancement, we use an $l_0$ structure-preserving smoothing filter~\cite{xu2011image}, which  linearly scales the image details, $\mathbf{I}_d$, obtained by the difference between the original and the output of an $l_0$ smoothing filter,  as ${\mathbf{I}_e=\mathbf{I}+\alpha \mathbf{I}_d}$, where 
$\alpha \in (0,10]$. For non-linear detail enhancement, we enhance image details in the $Lab$ colour space with a sigmoid, following~\cite{fan2018image} and similarly to EdgeFool (see Eq.~\ref{eq:EFImg}).
The log transformation expands darker pixel values and compresses brighter pixel values~\cite{10.5555/1076432}. Finally, the exponent ${1/\gamma}$ of gamma correction darkens ($\gamma< 1$) or brightens ($\gamma > 1$) the image~\cite{10.5555/1076432}.

\subsection{Semantic adversarial loss}
\begin{figure}[t]
\centering
\begin{tikzpicture}
\begin{axis}[
    footnotesize,
    ybar,
    width=8.6cm,
    height=4cm,
    axis lines=left,
    bar width=.001cm,
    xmin=1,
    ymin=-10,
    ymax=15,
    enlarge y limits=0.01,
    ylabel={\scriptsize Original logit values},
    x axis line style=-,
    xtick={1,100,200,...,1000}
    ]
    \addplot+[green] table [x=x, y=y]{./VisMapping/o_0.txt};
\end{axis}
\begin{axis}[
    footnotesize,
    ybar,
    width=8.6cm,
    height=4cm,
    axis lines=left,
    bar width=.001cm,
    xmin=1,
    ymin=-10,
    ymax=15,
    enlarge y limits=0.01,
    yticklabels=\empty,
    xticklabels=\empty,
    x axis line style={draw=none},
    y axis line style={draw=none}
    ]
    \addplot+[red] table [x=x, y=y]{./VisMapping/o_1.txt};
\end{axis}
\begin{axis}[
    footnotesize,
    ybar,
    width=8.6cm,
    height=4cm,
    axis lines=left,
    bar width=.001cm,
    xmin=1,
    ymin=-10,
    ymax=15,
    enlarge y limits=0.01,
    yticklabels=\empty,
    xticklabels=\empty,
    x axis line style={draw=none},
    y axis line style={draw=none}
    ]
    \addplot+[blue] table [x=x, y=y]{./VisMapping/o_2.txt};
\end{axis}
\begin{axis}[
    footnotesize,
    ybar,
    width=8.6cm,
    height=4cm,
    axis lines=left,
    bar width=.001cm,
    xmin=1,
    ymin=-10,
    ymax=15,
    enlarge y limits=0.01,
    yticklabels=\empty,
    xticklabels=\empty,
    x axis line style={draw=none},
    y axis line style={draw=none}
    ]
    \addplot+[bazaar] table [x=x, y=y]{./VisMapping/o_3.txt};
\end{axis}
\begin{axis}[
    footnotesize,
    ybar,
    width=8.6cm,
    height=4cm,
    axis lines=left,
    bar width=.001cm,
    xmin=1,
    ymin=-10,
    ymax=15,
    enlarge y limits=0.01,
    yticklabels=\empty,
    xticklabels=\empty,
    x axis line style={draw=none},
    y axis line style={draw=none}
    ]
    \addplot+[cyan] table [x=x, y=y]{./VisMapping/o_4.txt};
\end{axis}
\begin{axis}[
    footnotesize,
    ybar,
    width=8.6cm,
    height=4cm,
    axis lines=left,
    bar width=.001cm,
    xmin=1,
    ymin=-10,
    ymax=15,
    enlarge y limits=0.01,
    yticklabels=\empty,
    xticklabels=\empty,
    x axis line style={draw=none},
    y axis line style={draw=none}
    ]
    \addplot+[darkgreen] table [x=x, y=y]{./VisMapping/o_5.txt};
\end{axis}
\begin{axis}[
    footnotesize,
    ybar,
    width=8.6cm,
    height=4cm,
    axis lines=left,
    bar width=.001cm,
    xmin=1,
    ymin=-10,
    ymax=15,
    enlarge y limits=0.01,
    yticklabels=\empty,
    xticklabels=\empty,
    x axis line style={draw=none},
    y axis line style={draw=none}
    ]
    \addplot+[chocolate(traditional)] table [x=x, y=y]{./VisMapping/o_6.txt};
\end{axis}
\begin{axis}[
    footnotesize,
    ybar,
    width=8.6cm,
    height=4cm,
    axis lines=left,
    bar width=.001cm,
    xmin=1,
    ymin=-10,
    ymax=15,
    enlarge y limits=0.01,
    yticklabels=\empty,
    xticklabels=\empty,
    x axis line style={draw=none},
    y axis line style={draw=none}
    ]
    \addplot+[orange] table [x=x, y=y]{./VisMapping/o_7.txt};
\end{axis}
\begin{axis}[
    footnotesize,
    ybar,
    width=8.6cm,
    height=4cm,
    axis lines=left,
    bar width=.001cm,
    xmin=1,
    ymin=-10,
    ymax=15,
    enlarge y limits=0.01,
    yticklabels=\empty,
    xticklabels=\empty,
    x axis line style={draw=none},
    y axis line style={draw=none}
    ]
    \addplot+[violet(web)] table [x=x, y=y]{./VisMapping/o_8.txt};
\end{axis}
\begin{axis}[
    footnotesize,
    ybar,
    width=8.6cm,
    height=4cm,
    axis lines=left,
    bar width=.001cm,
    xmin=1,
    ymin=-10,
    ymax=15,
    enlarge y limits=0.01,
    yticklabels=\empty,
    xticklabels=\empty,
    x axis line style={draw=none},
    y axis line style={draw=none}
    ]
    \addplot+[viola] table [x=x, y=y]{./VisMapping/o_9.txt};
\end{axis}
\begin{axis}[
    footnotesize,
    ybar,
    width=8.6cm,
    height=4cm,
    axis lines=left,
    bar width=.001cm,
    xmin=1,
    ymin=-10,
    ymax=15,
    enlarge y limits=0.01,
    yticklabels=\empty,
    xticklabels=\empty,
    x axis line style={draw=none},
    y axis line style={draw=none}
    ]
    \addplot+[black] table [x=x, y=y]{./VisMapping/o_10.txt};
\end{axis}

\node[inner sep=0pt] (whitehead) at (0.6,2.05)
    {\includegraphics[width=1cm]{Labels/imgs/ILSVRC2012_val_00031875.png}};

\end{tikzpicture}

\begin{tikzpicture}
\begin{axis}[
    footnotesize,
    ybar,
    width=8.6cm,
    height=4cm,
    axis lines=left,
    bar width=.001cm,
    xmin=1,
    ymin=-10,
    ymax=15,
    enlarge y limits=0.01,
    ylabel={\scriptsize Adversarial logit values},
    x axis line style=-,
    xtick={1,100,200,...,1000}
    ]
    \addplot+[green] table [x=x, y=y]{./VisMapping/a_0.txt};
\end{axis}
\begin{axis}[
    footnotesize,
    ybar,
    width=8.6cm,
    height=4cm,
    axis lines=left,
    bar width=.001cm,
    xmin=1,
    ymin=-10,
    ymax=15,
    enlarge y limits=0.01,
    yticklabels=\empty,
    xticklabels=\empty,
    x axis line style={draw=none},
    y axis line style={draw=none}
    ]
    \addplot+[red] table [x=x, y=y]{./VisMapping/a_1.txt};
\end{axis}
\begin{axis}[
    footnotesize,
    ybar,
    width=8.6cm,
    height=4cm,
    axis lines=left,
    bar width=.001cm,
    xmin=1,
    ymin=-10,
    ymax=15,
    enlarge y limits=0.01,
    yticklabels=\empty,
    xticklabels=\empty,
    x axis line style={draw=none},
    y axis line style={draw=none}
    ]
    \addplot+[blue] table [x=x, y=y]{./VisMapping/a_2.txt};
\end{axis}
\begin{axis}[
    footnotesize,
    ybar,
    width=8.6cm,
    height=4cm,
    axis lines=left,
    bar width=.001cm,
    xmin=1,
    ymin=-10,
    ymax=15,
    enlarge y limits=0.01,
    yticklabels=\empty,
    xticklabels=\empty,
    x axis line style={draw=none},
    y axis line style={draw=none}
    ]
    \addplot+[bazaar] table [x=x, y=y]{./VisMapping/a_3.txt};
\end{axis}
\begin{axis}[
    footnotesize,
    ybar,
    width=8.6cm,
    height=4cm,
    axis lines=left,
    bar width=.001cm,
    xmin=1,
    ymin=-10,
    ymax=15,
    enlarge y limits=0.01,
    yticklabels=\empty,
    xticklabels=\empty,
    x axis line style={draw=none},
    y axis line style={draw=none}
    ]
    \addplot+[cyan] table [x=x, y=y]{./VisMapping/a_4.txt};
\end{axis}
\begin{axis}[
    footnotesize,
    ybar,
    width=8.6cm,
    height=4cm,
    axis lines=left,
    bar width=.001cm,
    xmin=1,
    ymin=-10,
    ymax=15,
    enlarge y limits=0.01,
    yticklabels=\empty,
    xticklabels=\empty,
    x axis line style={draw=none},
    y axis line style={draw=none}
    ]
    \addplot+[darkgreen] table [x=x, y=y]{./VisMapping/a_5.txt};
\end{axis}
\begin{axis}[
    footnotesize,
    ybar,
    width=8.6cm,
    height=4cm,
    axis lines=left,
    bar width=.001cm,
    xmin=1,
    ymin=-10,
    ymax=15,
    enlarge y limits=0.01,
    yticklabels=\empty,
    xticklabels=\empty,
    x axis line style={draw=none},
    y axis line style={draw=none}
    ]
    \addplot+[chocolate(traditional)] table [x=x, y=y]{./VisMapping/a_6.txt};
\end{axis}
\begin{axis}[
    footnotesize,
    ybar,
    width=8.6cm,
    height=4cm,
    axis lines=left,
    bar width=.001cm,
    xmin=1,
    ymin=-10,
    ymax=15,
    enlarge y limits=0.01,
    yticklabels=\empty,
    xticklabels=\empty,
    x axis line style={draw=none},
    y axis line style={draw=none}
    ]
    \addplot+[orange] table [x=x, y=y]{./VisMapping/a_7.txt};
\end{axis}
\begin{axis}[
    footnotesize,
    ybar,
    width=8.6cm,
    height=4cm,
    axis lines=left,
    bar width=.001cm,
    xmin=1,
    ymin=-10,
    ymax=15,
    enlarge y limits=0.01,
    yticklabels=\empty,
    xticklabels=\empty,
    x axis line style={draw=none},
    y axis line style={draw=none}
    ]
    \addplot+[violet(web)] table [x=x, y=y]{./VisMapping/a_8.txt};
\end{axis}
\begin{axis}[
    footnotesize,
    ybar,
    width=8.6cm,
    height=4cm,
    axis lines=left,
    bar width=.001cm,
    xmin=1,
    ymin=-10,
    ymax=15,
    enlarge y limits=0.01,
    yticklabels=\empty,
    xticklabels=\empty,
    x axis line style={draw=none},
    y axis line style={draw=none}
    ]
    \addplot+[viola] table [x=x, y=y]{./VisMapping/a_9.txt};
\end{axis}
\begin{axis}[
    footnotesize,
    ybar,
    width=8.6cm,
    height=4cm,
    axis lines=left,
    bar width=.001cm,
    xmin=1,
    ymin=-10,
    ymax=15,
    enlarge y limits=0.01,
    yticklabels=\empty,
    xticklabels=\empty,
    x axis line style={draw=none},
    y axis line style={draw=none}
    ]
    \addplot+[black] table [x=x, y=y]{./VisMapping/a_10.txt};
\end{axis}
 
\node[inner sep=0pt] (whitehead) at (0.6,2.05)
    {\includegraphics[width=1cm]{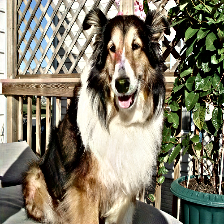}};     

\end{tikzpicture}
\caption{Logit values generated by ResNet50 for an original image (top) and for adversarial image (bottom) generated with the semantic adversarial loss. This loss minimises the positive logits of the labels that are semantically similar to the original label (e.g.~dogs) and increases the logit of a label that belongs to a different class. The horizontal axis represents the labels and colours represent the classes:
\textcolor{green}{dogs}, 
\textcolor{red}{other mammals}, 
\textcolor{blue}{birds}, 
\textcolor{bazaar}{reptiles,fish, amphibians}, 
\textcolor{cyan}{invertebrates}, 
\textcolor{darkgreen}{food, plants, fungi}, 
\textcolor{chocolate(traditional)}{devices}, 
\textcolor{orange}{structures, furnishing}, 
\textcolor{violet(web)}{clothes, covering}, 
\textcolor{viola}{implements, containers, misc. objects}, 
\textcolor{black}{vehicles}.}
\label{fig:SemAdvVis}
\end{figure}
We aim to devise an adversarial loss that operates on the groups of labels defined by $\mathbf{W}$ (see Sec.~\ref{sec:ProbForm}). To this end, we adapt the CW adversarial loss~\cite{carlini2017towards} and decrease the logits for the labels that share the same class $s$ with the label of the original image.

Let $\mathbf{w}_s \in \{0,1\}^D$ be the column of $\mathbf{W}$ that identifies the mapping of the labels to class $s$.
We apply a ReLU to the adversarial logit, $\dot{\mathbf{z}}$, and compute the dot product, $\langle\cdot,\cdot\rangle$, with ${\mathbf{w}}_s$ to focus the {\em same-class loss}, $\mathcal{L}_{\text{Sam}}(\cdot, \cdot)$, on the positive logits with the class $s$:
\begin{equation}
\label{eq:loss_sadv_1}
    \mathcal{L}_{\text{Sam}}({\dot{\mathbf{I}}},\mathbf{I})= \langle\text{ReLU}(\dot{\mathbf{z}}), \mathbf{w}_s\rangle.
\end{equation}

In each training iteration, we select the largest logit of a label not belonging to class $s$ using a {\em different-class loss} $\mathcal{L}_{\text{Diff}}(\cdot, \cdot)$: 
\begin{equation}
\label{eq:loss_sadv_2}
    \mathcal{L}_{\text{Diff}}({\dot{\mathbf{I}}},\mathbf{I})=   \max(\dot{\mathbf{z}} \odot \hat{\mathbf{w}}_s),
\end{equation}
where $\hat{\mathbf{w}}_s=\mathbf{1}-\mathbf{w}_s$ represents all the labels whose class differs from that of $y$, $\mathbf{1}=\{1\}^D$ and $\odot$ is the Hadamard product. 

The semantic adversarial loss, $\mathcal{L}_{\text{S-Adv}}(\cdot, \cdot)$, combines the same-class and different-class losses as:
\begin{equation}
\label{eq:loss_sadv}
    \mathcal{L}_{\text{S-Adv}}({\dot{\mathbf{I}}},\mathbf{I})=  \mathcal{L}_{\text{Sam}}({\dot{\mathbf{I}}},\mathbf{I}) - \mathcal{L}_{\text{Diff}}({\dot{\mathbf{I}}},\mathbf{I}).
\end{equation}

Figure~\ref{fig:SemAdvVis} shows an example of logit values for the ImageNet labels before and after using this semantic adversarial loss. 
\subsection{Multi-task loss}

We define our objective function, $\mathcal{L}$, as the combination of the structure loss, $\mathcal{L}_{\text{Str}}$, and the semantic adversarial loss, $\mathcal{L}_{\text{S-Adv}}$:
\begin{equation}
\label{eq:overall_loss}
\mathcal{L}=
\mathcal{L}_\text{Str} (\boldsymbol{\delta},\boldsymbol{\delta}_e ) +
\mathcal{L}_{\text{S-Adv}}({\dot{\mathbf{I}}},\mathbf{I}).
\end{equation}

During the iterative process that generates the perturbation, the FCNN learns to craft $\boldsymbol{\delta}$ by backpropagating both $\mathcal{L}_\text{Str}$ and $\mathcal{L}_\text{S-Adv}$ until $\dot{\mathbf{I}}$ misleads the classifier and $\mathcal{L}_\text{Str} < \tau$ (empirically set such that the adversarial image resembles an image enhanced with the target filter) or a maximum number of iterations is reached (3,000 in our case). 

\begin{figure}[t]

\centering
\setlength\tabcolsep{1.3pt}
\begin{tabular}{llll}
\multicolumn{1}{c}{\scriptsize Original} & 
\multicolumn{1}{c}{\scriptsize FilterFool (ND)} & 
\multicolumn{1}{c}{\scriptsize FilterFool (LD)} &  
\multicolumn{1}{c}{\scriptsize FilterFool (LT)}\\
\includegraphics[width=0.24\columnwidth]{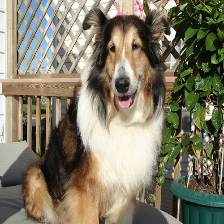}& 
\includegraphics[width=0.24\columnwidth]{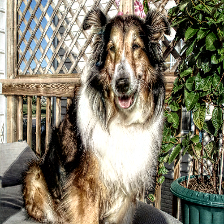} & 
\includegraphics[width=0.24\columnwidth]{imgs/VisComp/HF_LD10ILSVRC2012_val_00031875.png} & 
\includegraphics[width=0.24\columnwidth]{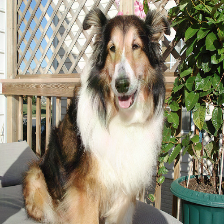}\\ 

\includegraphics[width=0.24\columnwidth]{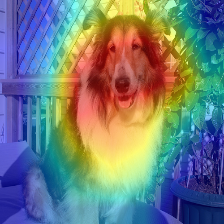}&
\includegraphics[width=0.24\columnwidth]{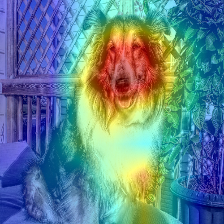}&
\includegraphics[width=0.24\columnwidth]{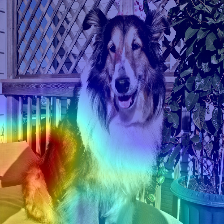}&
\includegraphics[width=0.24\columnwidth]{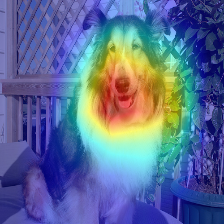}\\
\Gape[-5pt][-1pt]{\makecell[l]{\vspace{-0.1cm} \scriptsize Shetland sheepdog \\\vspace{-0.1cm} \scriptsize collie \\ \vspace{-0.1cm} \scriptsize German shepherd \\ \vspace{-0.1cm} \scriptsize Pembroke \\ \scriptsize Australian terrier}} &
\Gape[-5pt][-1pt]{\makecell[l]{\vspace{-0.1cm} \scriptsize peacock \\\vspace{-0.1cm} \scriptsize tarantula \\ \vspace{-0.1cm} \scriptsize window screen \\ \vspace{-0.1cm} \scriptsize wallaby \\ \scriptsize trash can}}  &
\Gape[-5pt][-1pt]{\makecell[l]{\vspace{-0.1cm} \scriptsize spatula \\\vspace{-0.1cm} \scriptsize cleaver \\ \vspace{-0.1cm} \scriptsize shovel \\ \vspace{-0.1cm} \scriptsize hammer \\ \scriptsize hatchet}} &
\Gape[-5pt][-1pt]{\makecell[l]{\vspace{-0.1cm} \scriptsize mask \\\vspace{-0.1cm} \scriptsize binoculars \\ \vspace{-0.1cm} \scriptsize sunscreen \\ \vspace{-0.1cm} \scriptsize koala \\ \scriptsize sunglasses}} \\
\end{tabular}
\caption{Comparison between the ResNet50 attention map on an original image and on three FilterFool examples, with the corresponding Top5 predictions. The colour of the heatmap ranging from blue to red indicates the importance of each image region (blue: least important; red: most important) in predicting the Top1 image label.}
\label{fig:adv_imgs}
\end{figure}

Figure~\ref{fig:adv_imgs} shows the Gradient-weighted Class Activation Maps (Grad-CAM)~\cite{selvaraju2017grad} of sample images alongside their Top5 predicted labels by ResNet50. Grad-CAM determines the importance of each neuron in a layer for the predicted Top1 label and the corresponding heatmap of the last convolutional layer is computed as  weighted sum of the neuron activations multiplied by their importance, followed by up-sampling to the original image size. It is possible to notice that in the original image ResNet50 focuses on the head, whereas in the FilterFool images the  focus is reduced in the face region  (ND, LT) or  shifted  towards another part of the image (LD). 

\section{Performance measures}

We evaluate adversarial attacks based on three main properties, namely  effectiveness,  robustness and transferability~\cite{Oh2021Visual}. 

{\em Effectiveness} (or success rate) is the degree to which the adversarial attack succeeds in misleading the classifier. Effectiveness can be measured as the accuracy of the classifier over a  dataset. The lower the accuracy, the higher the effectiveness of the attack. 

We measure effectiveness as {\em categorical} (label-level) and {\em semantic} (class-level) success rates. We also consider the Top$K$ success rate and, because of the study in this paper focuses on  ImageNet semantic labels, we also use the so called {semantic damage}~\cite{mopuri2020adversarial}. The semantic damage is  the average Wu-Palmar word similarity~\cite{wu1994verb}, $\text{Wu}(\cdot,\cdot)$, across the dataset. A damage occurs when the similarity between the  original label, $\lambda$, and adversarial label, $\dot{\lambda}$ is smaller than a threshold, $T_s$:
\begin{equation}
\text{Wu}_b =
\begin{cases}
1 \quad \text{Wu}(\lambda,\dot{\lambda}) < T_s \\
0 \quad \text{otherwise}. 
\end{cases}
\label{eq:SemD}
\end{equation}
where $\text{Wu}_b$ is the binarised  word similarity based on $T_s$.

\textit{Robustness} is the effectiveness of an adversarial attack in the presence of a defence (e.g.~median filtering, re-quantization, JPEG compression) that aims to remove the effect of a (potential) perturbation before the image is analysed by the classifier~\cite{xu2017feature}.   Robustness can be measured as the difference in accuracy of the classifier over a target dataset when a defence is used with respect to a setting when the defence is not used. The smaller this difference, the higher the robustness of the attack. A special case of defence is adversarial training, when the classifier itself is retrained with adversarial examples to increase its robustness~\cite{madry2018towards}.  

Finally, \textit{transferability} is the extent to which a perturbation crafted for a classifier is effective in misleading another classifier. Transferability can be measured as the difference in accuracy between the two classifiers over a target dataset. The smaller this difference, the higher the transferability of the attack.

\section{Validation}
\label{sec:val}
\subsection{Experimental setup}
\label{sec:para}

For the evaluation, we consider the following  adversarial attacks: Basic Iterative Method (BIM)~\cite{kurakin2016adversarial}, DeepFool~\cite{MoosaviDezfooli16}, SparseFool~\cite{modas2018sparsefool}, SemanticAdv~\cite{hosseini2018semantic}, ColorFool~\cite{shamsabadi2019colorfool}, EdgeFool~\cite{shamsabadi2019edgefool}, least-likely targeted~\cite{kurakin2016adversarial} and private targeted~\cite{Li2019} FGSM and BIM. 
As classifiers we use ResNet with 50 layers (ResNet50) and 18 layers (ResNet18)~\cite{he2016deep}, and AlexNet~\cite{krizhevsky2012imagenet}) trained for the image classification task on ImageNet~\cite{deng2009imagenet}, which includes $D=1,000$ classes. We use the same 3,000 images (3 images per class) that are randomly selected from the validation set by~\cite{shamsabadi2019colorfool,shamsabadi2019edgefool}. We consider $S=11$ WordNet semantic classes~\cite{tsipras2020imagenet}. 

The dimensions of $\mathbf{I}$ are $224\times224\times3$. We instantiate FCNN with the architecture of~\cite{wu2018fast}, which consists of 7 convolution layers with 24 intermediate feature maps and $3\times3$ kernels.
The last convolution layer applies a $1\times1$ convolution that generates $\mathbf{I}_s$.
The dilation factor of each layer is set to 1, 2, 4, 8, 16, 32, 1, and 1, respectively.
A leaky rectified linear unit is applied after padding and normalising each intermediate convolutional layer, except the last one. The hyper-parameter $\eta=0.01$ is chosen empirically to balance the errors of $\mathcal{L}_{l_2}$ and $\mathcal{L}_{\text{SSIM}}$ on the structure loss as $\mathcal{L}_{\text{ SSIM}}$ is bigger than $\mathcal{L}_{l_2}$. 
{Note that the hyper-parameter can affect the number of iterations needed to craft the perturbations. The structure loss and the semantic adversarial loss have equal contribution in the training. Important components of our optimisation process are the choice of the stopping criteria applied to the structure loss and the semantic adversarial loss.} The stopping threshold $\tau$ is set 0.04, 0.003 and 0.0005 for detail enhancement, log transformation and gamma correction, respectively. For linear detail enhancement and gamma correction, $\alpha$ and $\gamma$ range from 0.1 to 10. The parameters of the sigmoid in nonlinear detail enhancement are~\cite{fan2018image,shamsabadi2019edgefool}: $v_1=56$, $v_2=1$, and $v_3=15$. The parameters for bit reduction and median smoothing   are~\cite{xu2017feature}: 1 to 7 bits per colour channel in steps
of 1; kernel of $2\times2$, $3\times3$ and $5\times5$ (the
median values in odd and even kernels are the middle ones
and the mean of two middle ones, respectively). The quality
parameters we use for JPEG are 25, 50, 75 and 100.
As an adversarially trained classifier, we use ResNet50 
re-trained on adversarial images crafted with BIM~\cite{madry2018towards}.

\subsection{{FilterFool and traditional filters}}

{Figure~\ref{fig:GammaDetailCompFil} shows the categorical and semantic success rates of FilterFool and the target filters.} 
{As the filter effect becomes stronger, the success rates of Gamma correction and linear detail enhancement in traditional filters increase. However, the traditional filters have lower success rates than FilterFool, as expected. For example, the success rates of Log transformation, non-linear detail enhancement, Gamma correction ($\gamma=5$) and linear detail enhancement ($\alpha=5$) in misleading ResNet50 are 6\%, 29\%, 39\% and 51\%, respectively. FilterFool is successful in misleading ResNet50, ResNet18 and AlexNet with any strength of filters.}

{We use both objective and subjective evaluations to measure the differences between the FilterFool adversarial images and the target filtered images. As objective evaluation, we measure the SSIM between FilterFool and traditionally filtered images (see Table~\ref{tab:FilterFoolvsFilterQuality}). 
Because the adversarial perturbation of FilterFool is produced based on the structure loss function in Eq.~\ref{eq:loss_enh}, the SSIM values show that the outputs of FilterFool and the corresponding target filter are highly similar. The structure-aware, small  and adversarially guided differences  between the FilterFool adversarial images and the target filtered images enable FilterFool to mislead classifiers with high success rates.}

\begin{figure}[t!]
          \centering
          \subfloat{
          \begin{tikzpicture}
          \begin{axis}[cycle list name=color list,
                  footnotesize,
                  axis lines=left, 
                  width=4cm,
                  height=4cm,
                  ymin=0.00,
                  ymax=1.05,
                  title=\footnotesize Categorical,
                  enlarge y limits=0.01,
                  enlarge x limits=0.02,
                  xlabel={\footnotesize $\gamma$ (GC)},
                  ylabel={\footnotesize Success rate},
                  ytick={0,.2,.4,.6,.8, 1},
                  yticklabels={0,.2,.4,.6,.8, 1}
                  ]
          \addplot+[R50,line width=0.8pt,densely dashdotted]
              table[x=S,y=R50] {./tikz/ImageNet/FilterHF/GammaFilter.txt};
          \addplot+[R50,line width=0.8pt]
              table[x=S,y=R50] {./tikz/ImageNet/FilterHF/Gamma.txt};      

          \addplot+[R18, line width=0.8pt,densely dashdotted]
              table[x=S,y=R18] {./tikz/ImageNet/FilterHF/GammaFilter.txt};
          \addplot+[R18, dotted, line width=0.8pt]
              table[x=S,y=R18] {./tikz/ImageNet/FilterHF/Gamma.txt};           
              
          \addplot+[AN, line width=0.8pt,densely dashdotted]
              table[x=S,y=AN] {./tikz/ImageNet/FilterHF/GammaFilter.txt};
          \addplot+[AN, line width=0.8pt]
              table[x=S,y=AN] {./tikz/ImageNet/FilterHF/Gamma.txt};    
           \end{axis}
          \end{tikzpicture}} \hspace*{-0.6em} 
          \subfloat{\begin{tikzpicture}
          \begin{axis}[cycle list name=color list,
                  footnotesize,
                  axis lines=left, 
                  width=4cm,
                  height=4cm,
                  xmin=0.00,
                  ymin=0.00,
                  ymax=1.05,
                  title=\footnotesize Semantic,
                  enlarge y limits=0.01,
                  enlarge x limits=0.02,
                  xlabel={\footnotesize $\gamma$ (GC)},
                  ytick={0,.2,.4,.6,.8, 1},
                  yticklabels={0,.2,.4,.6,.8, 1}
                  ]
          \addplot+[R50,line width=0.8pt,densely dashdotted]
              table[x=S,y=R50S] {./tikz/ImageNet/FilterHF/GammaFilter.txt};
          \addplot+[R50,line width=0.8pt]
              table[x=S,y=R50S] {./tikz/ImageNet/FilterHF/Gamma.txt};      

          \addplot+[R18, line width=0.8pt,densely dashdotted]
              table[x=S,y=R18S] {./tikz/ImageNet/FilterHF/GammaFilter.txt};
          \addplot+[R18, dotted, line width=0.8pt]
              table[x=S,y=R18S] {./tikz/ImageNet/FilterHF/Gamma.txt};           
              
          \addplot+[AN, line width=0.8pt,densely dashdotted]
              table[x=S,y=ANS] {./tikz/ImageNet/FilterHF/GammaFilter.txt};
          \addplot+[AN, line width=0.8pt]
              table[x=S,y=ANS] {./tikz/ImageNet/FilterHF/Gamma.txt};  
           \end{axis}
          \end{tikzpicture}}\hspace*{-0.6em} 
          \subfloat{\begin{tikzpicture}
          \begin{axis}[cycle list name=color list,
                  footnotesize,
                  axis lines=left, 
                  width=2cm,
                  height=4cm,
                  ymin=0.00,
                  ymax=1.05,
                  title=\footnotesize C,
                  xlabel={\footnotesize LT},
                  enlarge y limits=0.01,
                  enlarge x limits=0.02,
                  ytick={0,.2,.4,.6,.8, 1},
                  yticklabels={0,.2,.4,.6,.8, 1},
                  xticklabels={1}
                  ]
          \addplot+[R50,only marks,mark=*,mark options={fill=white}]
              table[x=S,y=R50] {./tikz/ImageNet/FilterHF/LogFilter.txt};
          \addplot+[R50,only marks,mark=*]
              table[x=S,y=R50] {./tikz/ImageNet/FilterHF/Log.txt};

          \addplot+[R18,only marks,mark=*,mark options={fill=white}]
              table[x=S,y=R18] {./tikz/ImageNet/FilterHF/LogFilter.txt};
          \addplot+[R18,only marks,mark=*]
              table[x=S,y=R18] {./tikz/ImageNet/FilterHF/Log.txt};
              
          \addplot+[AN,only marks,mark=*,mark options={fill=white}]
               table[x=S,y=AN] {./tikz/ImageNet/FilterHF/LogFilter.txt};  
          \addplot+[AN,only marks, mark=*]
              table[x=S,y=AN] {./tikz/ImageNet/FilterHF/Log.txt};
           \end{axis}
          \end{tikzpicture}}\hspace*{-0.6em} 
          \subfloat{\begin{tikzpicture}
          \begin{axis}[cycle list name=color list,
                  footnotesize,
                  axis lines=left, 
                  width=2cm,
                  height=4cm,
                  ymin=0.00,
                  ymax=1.05,
                  title=\footnotesize S,
                  enlarge y limits=0.01,
                  enlarge x limits=0.02,
                  xlabel={\footnotesize LT},
                  ytick={0,.2,.4,.6,.8, 1},
                  yticklabels={0,.2,.4,.6,.8, 1},
                  xticklabels={1}
                  ]
          \addplot+[R50,only marks,mark=*,mark options={fill=white}]
              table[x=S,y=R50S] {./tikz/ImageNet/FilterHF/LogFilter.txt};
          \addplot+[R50,only marks,mark=*]
              table[x=S,y=R50S] {./tikz/ImageNet/FilterHF/Log.txt};

          \addplot+[R18,only marks,mark=*,mark options={fill=white}]
              table[x=S,y=R18S] {./tikz/ImageNet/FilterHF/LogFilter.txt};
          \addplot+[R18,only marks,mark=*]
              table[x=S,y=R18S] {./tikz/ImageNet/FilterHF/Log.txt};
              
          \addplot+[AN,only marks,mark=*,mark options={fill=white}]
               table[x=S,y=ANS] {./tikz/ImageNet/FilterHF/LogFilter.txt};  
          \addplot+[AN,only marks, mark=*]
              table[x=S,y=ANS] {./tikz/ImageNet/FilterHF/Log.txt};
           \end{axis}
          \end{tikzpicture} }\\
          \subfloat{
          \begin{tikzpicture}
          \begin{axis}[cycle list name=color list,
                  footnotesize,
                  axis lines=left, 
                  width=4cm,
                  height=4cm,
                  xmin=0.00,
                  ymin=0.00,
                  ymax=1.05,
                  title=\footnotesize Categorical,
                  enlarge y limits=0.01,
                  enlarge x limits=0.02,
                  xlabel={\footnotesize $\alpha$ (LD)},
                  ylabel={\footnotesize Success rate},
                  ytick={0,.2,.4,.6,.8, 1},
                  yticklabels={0,.2,.4,.6,.8, 1}
                  ]
          \addplot+[R50,line width=0.8pt,densely dashdotted]
              table[x=S,y=R50] {./tikz/ImageNet/FilterHF/Linear_DetailFilter.txt};
          \addplot+[R50,line width=0.8pt]
              table[x=S,y=R50] {./tikz/ImageNet/FilterHF/Linear_Detail.txt};      

          \addplot+[R18, line width=0.8pt,densely dashdotted]
              table[x=S,y=R18] {./tikz/ImageNet/FilterHF/Linear_DetailFilter.txt};
          \addplot+[R18, dotted, line width=0.8pt]
              table[x=S,y=R18] {./tikz/ImageNet/FilterHF/Linear_Detail.txt};           
              
          \addplot+[AN, line width=0.8pt,densely dashdotted]
              table[x=S,y=AN] {./tikz/ImageNet/FilterHF/Linear_DetailFilter.txt};
          \addplot+[AN, line width=0.8pt]
              table[x=S,y=AN] {./tikz/ImageNet/FilterHF/Linear_Detail.txt};  
           \end{axis}
          \end{tikzpicture}} \hspace*{-0.6em} 
          \subfloat{\begin{tikzpicture}
          \begin{axis}[cycle list name=color list,
                  footnotesize,
                  axis lines=left, 
                  width=4cm,
                  height=4cm,
                  xmin=0.00,
                  ymin=0.00,
                  ymax=1.05,
                  title=\footnotesize Semantic,
                  enlarge y limits=0.01,
                  enlarge x limits=0.02,
                  xlabel={\footnotesize $\alpha$ (LD)},
                  ytick={0,.2,.4,.6,.8, 1},
                  yticklabels={0,.2,.4,.6,.8, 1}
                  ]
          \addplot+[R50,line width=0.8pt,densely dashdotted]
              table[x=S,y=R50S] {./tikz/ImageNet/FilterHF/Linear_DetailFilter.txt};
          \addplot+[R50,line width=0.8pt]
              table[x=S,y=R50S] {./tikz/ImageNet/FilterHF/Linear_Detail.txt};      

          \addplot+[R18, line width=0.8pt,densely dashdotted]
              table[x=S,y=R18S] {./tikz/ImageNet/FilterHF/Linear_DetailFilter.txt};
          \addplot+[R18, dotted, line width=0.8pt]
              table[x=S,y=R18S] {./tikz/ImageNet/FilterHF/Linear_Detail.txt};           
              
          \addplot+[AN, line width=0.8pt,densely dashdotted]
              table[x=S,y=ANS] {./tikz/ImageNet/FilterHF/Linear_DetailFilter.txt};\label{cap:F-sr-l}
          \addplot+[AN, line width=0.8pt]
              table[x=S,y=ANS] {./tikz/ImageNet/FilterHF/Linear_Detail.txt}; \label{cap:FF-sr-l} 
           \end{axis}
          \end{tikzpicture}}\hspace*{-0.6em} 
          \subfloat{\begin{tikzpicture}
          \begin{axis}[cycle list name=color list,
                  footnotesize,
                  axis lines=left, 
                  width=2cm,
                  height=4cm,
                  ymin=0.00,
                  ymax=1.05,
                  title=\footnotesize C,
                  xlabel={\footnotesize ND},
                  enlarge y limits=0.01,
                  enlarge x limits=0.02,
                  ytick={0,.2,.4,.6,.8, 1},
                  yticklabels={0,.2,.4,.6,.8, 1},
                  xticklabels={1},
                  xtick={1}
                  ]
          \addplot+[R50,only marks,mark=*,mark options={fill=white}]
              table[x=S,y=R50] {./tikz/ImageNet/FilterHF/Nonlinear_DetailFilter.txt};
          \addplot+[R50,only marks,mark=*]
              table[x=S,y=R50] {./tikz/ImageNet/FilterHF/Nonlinear_Detail.txt};      

          \addplot+[R18,only marks,mark=*,mark options={fill=white}]
              table[x=S,y=R18] {./tikz/ImageNet/FilterHF/Nonlinear_DetailFilter.txt};
          \addplot+[R18,only marks,mark=*]
              table[x=S,y=R18] {./tikz/ImageNet/FilterHF/Nonlinear_Detail.txt};           
              
          \addplot+[AN,only marks,mark=*,mark options={fill=white}]
              table[x=S,y=AN] {./tikz/ImageNet/FilterHF/Nonlinear_DetailFilter.txt};
          \addplot+[AN,only marks,mark=*]
              table[x=S,y=AN] {./tikz/ImageNet/FilterHF/Nonlinear_Detail.txt};  
           \end{axis}
          \end{tikzpicture}}\hspace*{-0.6em} 
          \subfloat{\begin{tikzpicture}
          \begin{axis}[cycle list name=color list,
                  footnotesize,
                  axis lines=left, 
                  width=2cm,
                  height=4cm,
                  ymin=0.00,
                  ymax=1.05,
                  title=\footnotesize S,
                  enlarge y limits=0.01,
                  enlarge x limits=0.02,
                  xlabel={\footnotesize ND},
                  ytick={0,.2,.4,.6,.8, 1},
                  yticklabels={0,.2,.4,.6,.8, 1},
                  xticklabels={1},
                  xtick={1}
                  ]
          \addplot+[R50,only marks,mark=*,mark options={fill=white}]
              table[x=S,y=R50S] {./tikz/ImageNet/FilterHF/Nonlinear_DetailFilter.txt};
          \addplot+[R50,only marks,mark=*]
              table[x=S,y=R50S] {./tikz/ImageNet/FilterHF/Nonlinear_Detail.txt};     

          \addplot+[R18,only marks,mark=*,mark options={fill=white}]
              table[x=S,y=R18S] {./tikz/ImageNet/FilterHF/Nonlinear_DetailFilter.txt};
          \addplot+[R18,only marks,mark=*]
              table[x=S,y=R18S] {./tikz/ImageNet/FilterHF/Nonlinear_Detail.txt};           
              
          \addplot+[AN,only marks,mark=*,mark options={fill=white}]
              table[x=S,y=ANS] {./tikz/ImageNet/FilterHF/Nonlinear_DetailFilter.txt};\label{cap:F-sr-p}
          \addplot+[AN,only marks,mark=*]
              table[x=S,y=ANS] {./tikz/ImageNet/FilterHF/Nonlinear_Detail.txt}; \label{cap:FF-sr-p} 
           \end{axis}
          \end{tikzpicture} 
          }
           \caption{Comparing the categorical (C) and semantic (S) success rates of FilterFool {(\ref{cap:FF-sr-l}~or~\ref{cap:FF-sr-p})}  with its corresponding traditional filters {(\ref{cap:F-sr-l}~or~\ref{cap:F-sr-p})}: Nonlinear Detail enhancement (ND), Log Transformation (LT), Gamma Correction (GC), Linear Detail enhancement (LD) with strengths varying from 0.1 to 10 on the ImageNet dataset against {\color{R50} ResNet50}, {\color{R18} ResNet18} and {\color{AN} AlexNet}. 
           }
            \label{fig:GammaDetailCompFil}
\end{figure}
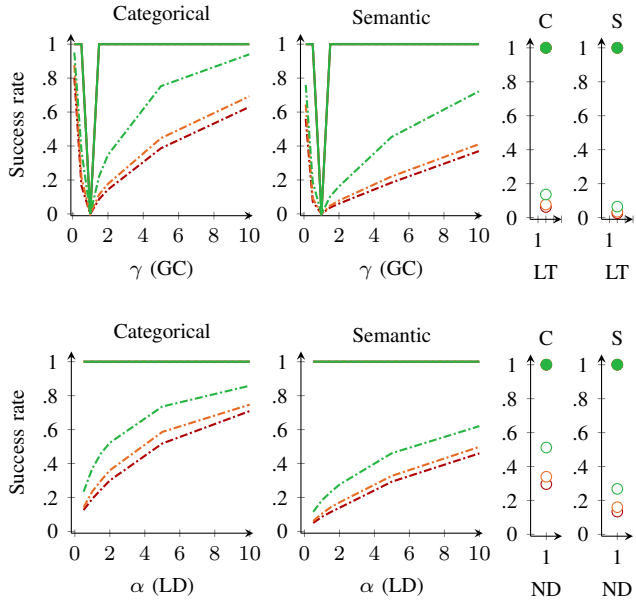
\begin{table}[t!]
\centering

\caption{{Similarity  of  the FilterFool adversarial images and  perturbations and their corresponding traditionally filtered images and pixel changes.}}

\begin{tabular}{|l|l|cc|}
    \Xhline{3\arrayrulewidth}
\multirow{2}{*}{Attack} & \multirow{2}{*}{Model} &  \multicolumn{2}{c|}{SSIM}\\
          &    & Adv. images & Adv. perturbation          \\
    
    \Xhline{3\arrayrulewidth} 
    \multirow{3}{*}{FF (ND)} &   R50 & $0.98\pm0.02$ & $0.97\pm0.02$ \\
                            &   R18 &  $0.98\pm0.01$ & $0.97\pm0.01$ \\
                            &   A   &  $0.97\pm0.02$ &  $0.96\pm0.02$\\
                            
    \hline
    \multirow{3}{*}{FF (Log)} &   R50 & $0.99\pm0.01$ & $0.99\pm0.00$\\
                            &   R18 &  $0.99\pm0.01$  & $0.99\pm0.00$\\
                            &   A   &  $0.98\pm0.02$  & $0.98\pm0.02$\\
    \hline                        
\multirow{3}{*}{FF (LD1)} &   R50 & $0.99\pm0.01$ & $0.99\pm0.00$\\
                            &   R18 &  $0.99\pm0.01$ & $0.99\pm0.00$\\
                            &   A   &  $0.98\pm0.02$ & $0.99\pm0.01$\\  

      \hline
\multirow{3}{*}{FF (GC.5)} &   R50 & $0.99\pm0.02$ & $1.00\pm0.01$\\
                            &   R18 & $0.99\pm0.02$ & $1.00\pm0.01$\\
                            &   A   &$0.98\pm0.02$ & $0.99\pm0.01$\\

    \Xhline{3\arrayrulewidth} 

    \end{tabular}

    \label{tab:FilterFoolvsFilterQuality}
\end{table}

{We assessed the similarity of the FilterFool images with a panel of observers (Queen Mary Ethics of Research Committee reference number: QMERC20.452). We considered all four filters: linear detail enhancement ($\alpha=1$), non-linear detail enhancement, Log transformation and Gamma correction ($\gamma=0.5$).
We first ranked the 3,000 output images from FilterFool based on their SSIM values with respect to the filtered images and then divided the images into three groups, namely low, middle, and high SSIM values (the average and standard deviation of SSIM values are reported in Table~\ref{tab:FilterFoolvsFilterQuality}). 
Next, we randomly chose 5 images from each group to obtain a total of 15 images for each filter.
Next 33 human subjects were shown a pair of images from FilterFool and the corresponding target filter with an original image, and asked: \emph{Are there any differences between these two filtered images?} Each pair of images was shown for 7 seconds, after which the answer (\textit{Yes} or \textit{No}) was provided. We randomised the display order of the image pairs and  used the Mean Opinion Score (MOS) to quantify the percentage of answers that perceive no differences between the FilterFool adversarial images and the target filtered images. 
The resulting MOS is $83.41\pm 9.83\%$, which suggests a high fidelity by FilterFool in mimicking the target filters.}

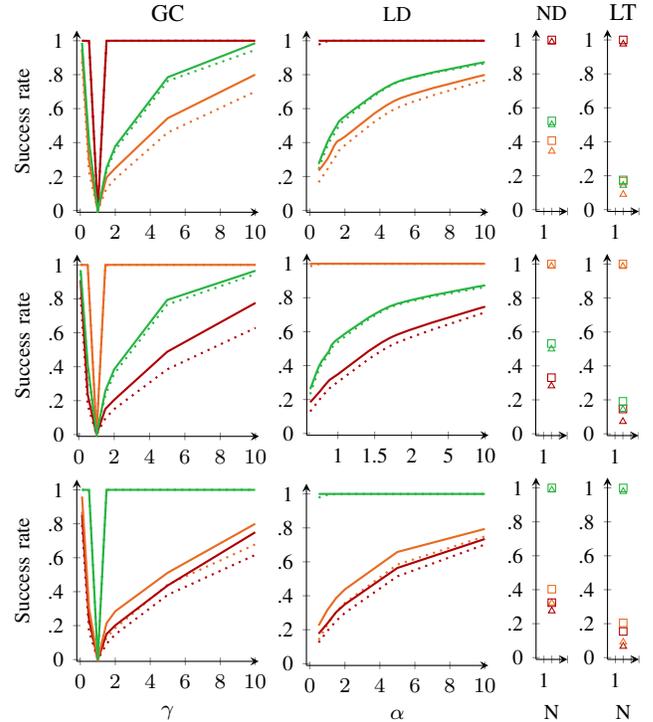
\begin{figure}[t!]
          \centering
          
          \begin{tikzpicture}
          \begin{axis}[cycle list name=color list,
                  footnotesize,
                  axis lines=left, 
                  width=4cm,
                  height=4cm,
                  xmin=0.00,
                  ymin=0.00,
                  ymax=1.05,
                  title=GC,
                  enlarge y limits=0.01,
                  enlarge x limits=0.02,
                  ylabel={\footnotesize Success rate},
                  ytick={0,.2,.4,.6,.8, 1},
                  yticklabels={0,.2,.4,.6,.8, 1}
                  ]
          \addplot+[R50,dotted,line width=0.8pt]
              table[x expr=\thisrow{S},y=R50] {./tikz/ImageNet/Transfer/GammaDNN_resnet50.txt};

          \addplot+[R18,dotted,line width=0.8pt]
              table[x expr=\thisrow{S},y=R18] {./tikz/ImageNet/Transfer/GammaDNN_resnet50.txt}; 
              
          \addplot+[AN,dotted,line width=0.8pt]
              table[x expr=\thisrow{S},y=AN] {./tikz/ImageNet/Transfer/GammaDNN_resnet50.txt}; 
              
         \addplot+[R50,line width=0.8pt]
              table[x expr=\thisrow{S},y=R50] {./tikz/ImageNet/Transfer/SemHF/Gamma_resnet50.txt};
          
          \addplot+[R18,line width=0.8pt]
              table[x expr=\thisrow{S},y=R18] {./tikz/ImageNet/Transfer/SemHF/Gamma_resnet50.txt}; 
              
          \addplot+[AN,line width=0.8pt]
              table[x expr=\thisrow{S},y=AN] {./tikz/ImageNet/Transfer/SemHF/Gamma_resnet50.txt};         
           \end{axis}
          \end{tikzpicture}\hspace*{-0.6em}  
          \begin{tikzpicture}
          \begin{axis}[cycle list name=color list,
                  footnotesize,
                  axis lines=left, 
                  width=4cm,
                  height=4cm,
                  xmin=0.00,
                  ymin=0.00,
                  ymax=1.05,
                  title=\footnotesize LD,
                  enlarge y limits=0.01,
                  smooth,
                  enlarge x limits=0.02,
                  ytick={0,.2,.4,.6,.8, 1},
                  yticklabels={0,.2,.4,.6,.8, 1}
                  ]
          \addplot+[R50,dotted,line width=0.8pt]
              table[x expr=\thisrow{S},y=R50] {./tikz/ImageNet/Transfer/UnsharpDNN_resnet50.txt};
          
          \addplot+[R18,dotted,line width=0.8pt]
              table[x expr=\thisrow{S},y=R18] {./tikz/ImageNet/Transfer/UnsharpDNN_resnet50.txt}; 
              
          \addplot+[AN,dotted,line width=0.8pt]
              table[x expr=\thisrow{S},y=AN] {./tikz/ImageNet/Transfer/UnsharpDNN_resnet50.txt}; 
              
         \addplot+[R50,line width=0.8pt]
              table[x expr=\thisrow{S},y=R50] {./tikz/ImageNet/Transfer/SemHF/Linear_Detail_resnet50.txt};
          
          \addplot+[R18,line width=0.8pt]
              table[x expr=\thisrow{S},y=R18] {./tikz/ImageNet/Transfer/SemHF/Linear_Detail_resnet50.txt}; 
              
          \addplot+[AN,line width=0.8pt]
              table[x expr=\thisrow{S},y=AN] {./tikz/ImageNet/Transfer/SemHF/Linear_Detail_resnet50.txt};       
           \end{axis}
          \end{tikzpicture}\hspace*{-0.6em}  
        \begin{tikzpicture}
          \begin{axis}[cycle list name=color list,
                  footnotesize,
                  axis lines=left, 
                  width=2cm,
                  height=4cm,
                  ymin=0.00,
                  ymax=1.05,
                  title=\footnotesize ND,
                  enlarge y limits=0.01,
                  enlarge x limits=0.02,
                  ytick={0,.2,.4,.6,.8, 1},
                  yticklabels={0,.2,.4,.6,.8, 1},
                  xticklabels={1}
                  ]
          \addplot+[R50, only marks,
            mark=triangle, mark size=0.5mm]
              table[x expr=\thisrow{S},y=R50] {./tikz/ImageNet/Transfer/DetailDNN_resnet50.txt};
          
          \addplot+[R18, only marks,
            mark=triangle, mark size=0.5mm]
              table[x expr=\thisrow{S},y=R18] {./tikz/ImageNet/Transfer/DetailDNN_resnet50.txt}; 
              
          \addplot+[AN, only marks,
            mark=triangle, mark size=0.5mm]
              table[x expr=\thisrow{S},y=AN] {./tikz/ImageNet/Transfer/DetailDNN_resnet50.txt}; 
              
        \addplot+[R50, only marks,
            mark=square, mark size=0.5mm]
              table[x expr=\thisrow{S},y=R50] {./tikz/ImageNet/Transfer/SemHF/Nonlinear_Detail_resnet50.txt};
          
          \addplot+[R18, only marks,
            mark=square, mark size=0.5mm]
              table[x expr=\thisrow{S},y=R18] {./tikz/ImageNet/Transfer/SemHF/Nonlinear_Detail_resnet50.txt}; 
              
          \addplot+[AN, only marks,
            mark=square, mark size=0.5mm]
              table[x expr=\thisrow{S},y=AN] {./tikz/ImageNet/Transfer/SemHF/Nonlinear_Detail_resnet50.txt};       
           \end{axis}
          \end{tikzpicture}\hspace*{-0.6em} 
          \begin{tikzpicture}
          \begin{axis}[cycle list name=color list,
                  footnotesize,
                  axis lines=left, 
                  width=2cm,
                  height=4cm,
                  ymin=0.00,
                  ymax=1.05,
                  title=LT,
                  enlarge y limits=0.01,
                  enlarge x limits=0.02,
                  ytick={0,.2,.4,.6,.8, 1},
                  yticklabels={0,.2,.4,.6,.8, 1},
                  xticklabels={1}
                  ]
          \addplot+[R50, only marks,
            mark=triangle, mark size=0.5mm]
              table[x expr=\thisrow{S},y=R50] {./tikz/ImageNet/Transfer/LogDNN_resnet50.txt};
        \addplot+[R50, only marks,
            mark=square, mark size=0.5mm]
              table[x expr=\thisrow{S},y=R50] {./tikz/ImageNet/Transfer/SemHF/Log_resnet50.txt};      
          
          \addplot+[R18, only marks,
            mark=triangle, mark size=0.5mm]
              table[x expr=\thisrow{S},y=R18] {./tikz/ImageNet/Transfer/LogDNN_resnet50.txt}; 
           \addplot+[R18, only marks,
            mark=square, mark size=0.5mm]
              table[x expr=\thisrow{S},y=R18] {./tikz/ImageNet/Transfer/SemHF/Log_resnet50.txt};

          \addplot+[AN, only marks,
            mark=triangle, mark size=0.5mm]
              table[x expr=\thisrow{S},y=AN] {./tikz/ImageNet/Transfer/LogDNN_resnet50.txt}; 
          \addplot+[AN, only marks,
            mark=square, mark size=0.5mm]
              table[x expr=\thisrow{S},y=AN] {./tikz/ImageNet/Transfer/SemHF/Log_resnet50.txt};       
           \end{axis}
          \end{tikzpicture}

          \begin{tikzpicture}
          \begin{axis}[cycle list name=color list,
                  footnotesize,
                  axis lines=left, 
                  width=4cm,
                  height=4cm,
                  ymin=0.00,
                  ymax=1.05,
                  enlarge y limits=0.01,
                  enlarge x limits=0.02,
                  ylabel={\footnotesize Success rate},
                  ytick={0,.2,.4,.6,.8, 1},
                  yticklabels={0,.2,.4,.6,.8, 1}
                  ]
          \addplot+[R50,dotted,line width=0.8pt]
              table[x expr=\thisrow{S},y=R50] {./tikz/ImageNet/Transfer/GammaDNN_resnet18.txt};
          
          \addplot+[R18,dotted,line width=0.8pt]
              table[x expr=\thisrow{S},y=R18] {./tikz/ImageNet/Transfer/GammaDNN_resnet18.txt}; 
              
          \addplot+[AN,dotted,line width=0.8pt]
              table[x expr=\thisrow{S},y=AN] {./tikz/ImageNet/Transfer/GammaDNN_resnet18.txt}; 
              
          \addplot+[R50,line width=0.8pt]
              table[x expr=\thisrow{S},y=R50] {./tikz/ImageNet/Transfer/SemHF/Gamma_resnet18.txt};
          
          \addplot+[R18,line width=0.8pt]
              table[x expr=\thisrow{S},y=R18] {./tikz/ImageNet/Transfer/SemHF/Gamma_resnet18.txt}; 
              
          \addplot+[AN,line width=0.8pt]
              table[x expr=\thisrow{S},y=AN] {./tikz/ImageNet/Transfer/SemHF/Gamma_resnet18.txt};        
           \end{axis}
          \end{tikzpicture}\hspace*{-0.6em} 
          \begin{tikzpicture}
          \begin{axis}[cycle list name=color list,
                  footnotesize,
                  axis lines=left, 
                  width=4cm,
                  height=4cm,
                  ymin=0.00,
                  ymax=1.05,
                  smooth,
                  enlarge y limits=0.01,
                  enlarge x limits=0.02,
                  ytick={0,.2,.4,.6,.8, 1},
                  yticklabels={0,.2,.4,.6,.8, 1},
                  xticklabels={.1,.5,1,1.5,2, 5, 10}
                  ]
          \addplot+[R50, dotted,line width=0.8pt]
              table[x expr=\thisrow{S},y=R50] {./tikz/ImageNet/Transfer/UnsharpDNN_resnet18.txt};
          
          \addplot+[R18, dotted,line width=0.8pt]
              table[x expr=\thisrow{S},y=R18] {./tikz/ImageNet/Transfer/UnsharpDNN_resnet18.txt}; 
              
          \addplot+[AN, dotted,line width=0.8pt]
              table[x expr=\thisrow{S},y=AN] {./tikz/ImageNet/Transfer/UnsharpDNN_resnet18.txt}; 
              
          \addplot+[R50,line width=0.8pt]
              table[x expr=\thisrow{S},y=R50] {./tikz/ImageNet/Transfer/SemHF/Linear_Detail_resnet18.txt};
          
          \addplot+[R18,line width=0.8pt]
              table[x expr=\thisrow{S},y=R18] {./tikz/ImageNet/Transfer/SemHF/Linear_Detail_resnet18.txt}; 
              
          \addplot+[AN,line width=0.8pt]
              table[x expr=\thisrow{S},y=AN] {./tikz/ImageNet/Transfer/SemHF/Linear_Detail_resnet18.txt};       
           \end{axis}
          \end{tikzpicture}\hspace*{-0.6em}  
        \begin{tikzpicture}
          \begin{axis}[cycle list name=color list,
                  footnotesize,
                  axis lines=left, 
                  width=2cm,
                  height=4cm,
                  ymin=0.00,
                  ymax=1.05,
                  enlarge y limits=0.01,
                  enlarge x limits=0.02,
                  ytick={0,.2,.4,.6,.8, 1},
                  yticklabels={0,.2,.4,.6,.8, 1},
                  xticklabels={1}
                  ]
          \addplot+[R50, only marks,
            mark=triangle, mark size=0.5mm]
              table[x expr=\thisrow{S},y=R50] {./tikz/ImageNet/Transfer/DetailDNN_resnet18.txt};
          
          \addplot+[R18, only marks,
            mark=triangle, mark size=0.5mm]
              table[x expr=\thisrow{S},y=R18] {./tikz/ImageNet/Transfer/DetailDNN_resnet18.txt}; 
              
          \addplot+[AN, only marks,
            mark=triangle, mark size=0.5mm]
              table[x expr=\thisrow{S},y=AN] {./tikz/ImageNet/Transfer/DetailDNN_resnet18.txt}; 
             
         \addplot+[R50, only marks,
            mark=square, mark size=0.5mm]
              table[x expr=\thisrow{S},y=R50] {./tikz/ImageNet/Transfer/SemHF/Nonlinear_Detail_resnet18.txt};
          
          \addplot+[R18, only marks,
            mark=square, mark size=0.5mm]
              table[x expr=\thisrow{S},y=R18] {./tikz/ImageNet/Transfer/SemHF/Nonlinear_Detail_resnet18.txt}; 
              
          \addplot+[AN, only marks,
            mark=square, mark size=0.5mm]
              table[x expr=\thisrow{S},y=AN] {./tikz/ImageNet/Transfer/SemHF/Nonlinear_Detail_resnet18.txt};       
           \end{axis}
          \end{tikzpicture}\hspace*{-0.6em}  
          \begin{tikzpicture}
          \begin{axis}[cycle list name=color list,
                  footnotesize,
                  axis lines=left, 
                  width=2cm,
                  height=4cm,
                  ymin=0.00,
                  ymax=1.05,
                  enlarge y limits=0.01,
                  enlarge x limits=0.02,
                  ytick={0,.2,.4,.6,.8, 1},
                  yticklabels={0,.2,.4,.6,.8, 1},
                  xticklabels={1}
                  ]
          \addplot+[R50, only marks,
            mark=triangle, mark size=0.5mm]
              table[x expr=\thisrow{S},y=R50] {./tikz/ImageNet/Transfer/LogDNN_resnet18.txt};
          \addplot+[R50, only marks,
            mark=square, mark size=0.5mm]
              table[x expr=\thisrow{S},y=R50] {./tikz/ImageNet/Transfer/SemHF/Log_resnet18.txt};

          \addplot+[R18, only marks,
            mark=triangle, mark size=0.5mm]
              table[x expr=\thisrow{S},y=R18] {./tikz/ImageNet/Transfer/LogDNN_resnet18.txt};
          \addplot+[R18, only marks,
            mark=square, mark size=0.5mm]
              table[x expr=\thisrow{S},y=R18] {./tikz/ImageNet/Transfer/SemHF/Log_resnet18.txt};
              
          \addplot+[AN, only marks,
            mark=triangle, mark size=0.5mm]
              table[x expr=\thisrow{S},y=AN] {./tikz/ImageNet/Transfer/LogDNN_resnet18.txt}; 
          \addplot+[AN, only marks,
            mark=square, mark size=0.5mm]
              table[x expr=\thisrow{S},y=AN] {./tikz/ImageNet/Transfer/SemHF/Log_resnet18.txt};       
           \end{axis}
          \end{tikzpicture}

                    \begin{tikzpicture}
          \begin{axis}[cycle list name=color list,
                  footnotesize,
                  axis lines=left, 
                  width=4cm,
                  height=4cm,
                  xmin=0.00,
                  ymin=0.00,
                  ymax=1.05,
                  enlarge y limits=0.01,
                  enlarge x limits=0.02,
                  xlabel={\footnotesize $\gamma$},
                  ylabel={\footnotesize Success rate},
                  y tick label style={
                  /pgf/number format/.cd,
                  fixed,
                  fixed zerofill,
                  precision=2,
                  /tikz/.cd
                  },
                  ytick={0,.2,.4,.6,.8, 1},
                  yticklabels={0,.2,.4,.6,.8, 1},
                  ]
          \addplot+[R50,dotted,line width=0.8pt]
              table[x expr=\thisrow{S},y=R50] {./tikz/ImageNet/Transfer/GammaDNN_alexnet.txt};
          
          \addplot+[R18,dotted,line width=0.8pt]
              table[x expr=\thisrow{S},y=R18] {./tikz/ImageNet/Transfer/GammaDNN_alexnet.txt}; 
              
          \addplot+[AN,dotted,line width=0.8pt]
              table[x expr=\thisrow{S},y=AN] {./tikz/ImageNet/Transfer/GammaDNN_alexnet.txt}; 
        
        \addplot+[R50,line width=0.8pt]
              table[x expr=\thisrow{S},y=R50] {./tikz/ImageNet/Transfer/SemHF/Gamma_alexnet.txt};
          
          \addplot+[R18,line width=0.8pt]
              table[x expr=\thisrow{S},y=R18] {./tikz/ImageNet/Transfer/SemHF/Gamma_alexnet.txt}; 
              
          \addplot+[AN,line width=0.8pt]
              table[x expr=\thisrow{S},y=AN] {./tikz/ImageNet/Transfer/SemHF/Gamma_alexnet.txt};       
           \end{axis}
          \end{tikzpicture}\hspace*{-0.6em}  
          \begin{tikzpicture}
          \begin{axis}[cycle list name=color list,
                  footnotesize,
                  axis lines=left, 
                  width=4cm,
                  height=4cm,
                  xmin=0.00,
                  ymin=0.00,
                  ymax=1.05,
                  enlarge y limits=0.01,
                  enlarge x limits=0.02,
                  xlabel={\footnotesize $\alpha$},
                  ytick={0,.2,.4,.6,.8, 1},
                  yticklabels={0,.2,.4,.6,.8, 1},
                  ]
          \addplot+[R50, dotted,line width=0.8pt]
              table[x expr=\thisrow{S},y=R50] {./tikz/ImageNet/Transfer/UnsharpDNN_alexnet.txt};
          
          \addplot+[R18, dotted,line width=0.8pt]
              table[x expr=\thisrow{S},y=R18] {./tikz/ImageNet/Transfer/UnsharpDNN_alexnet.txt}; 
              
          \addplot+[AN, dotted,line width=0.8pt]
              table[x expr=\thisrow{S},y=AN] {./tikz/ImageNet/Transfer/UnsharpDNN_alexnet.txt}; \label{cap:FF-c-l}
        
        \addplot+[R50,line width=0.8pt]
              table[x expr=\thisrow{S},y=R50] {./tikz/ImageNet/Transfer/SemHF/Linear_Detail_alexnet.txt};
          
          \addplot+[R18,line width=0.8pt]
              table[x expr=\thisrow{S},y=R18] {./tikz/ImageNet/Transfer/SemHF/Linear_Detail_alexnet.txt}; 
              
          \addplot+[AN,line width=0.8pt]
              table[x expr=\thisrow{S},y=AN] {./tikz/ImageNet/Transfer/SemHF/Linear_Detail_alexnet.txt}; \label{cap:FF-l}

           \end{axis}
          \end{tikzpicture}\hspace*{-0.6em}  
        \begin{tikzpicture}
          \begin{axis}[cycle list name=color list,
                  footnotesize,
                  axis lines=left, 
                  width=2cm,
                  height=4cm,
                  ymin=0.00,
                  ymax=1.05,
                  enlarge y limits=0.01,
                  enlarge x limits=0.02,
                  xlabel={\footnotesize N},
                  ytick={0,.2,.4,.6,.8, 1},
                  yticklabels={0,.2,.4,.6,.8, 1},
                  xticklabels={1}
                  ]
          \addplot+[R50, only marks,
            mark=triangle, mark size=0.5mm]
              table[x expr=\thisrow{S},y=R50] {./tikz/ImageNet/Transfer/DetailDNN_alexnet.txt};
          
          \addplot+[R18, only marks,
            mark=triangle, mark size=0.5mm]
              table[x expr=\thisrow{S},y=R18] {./tikz/ImageNet/Transfer/DetailDNN_alexnet.txt}; 
              
          \addplot+[AN, only marks,
            mark=triangle, mark size=0.5mm]
              table[x expr=\thisrow{S},y=AN] {./tikz/ImageNet/Transfer/DetailDNN_alexnet.txt}; 
              
          \addplot+[R50, only marks,
            mark=square, mark size=0.5mm]
              table[x expr=\thisrow{S},y=R50] {./tikz/ImageNet/Transfer/SemHF/Nonlinear_Detail_alexnet.txt};
          
          \addplot+[R18, only marks,
            mark=square, mark size=0.5mm]
              table[x expr=\thisrow{S},y=R18] {./tikz/ImageNet/Transfer/SemHF/Nonlinear_Detail_alexnet.txt}; 
              
          \addplot+[AN, only marks,
            mark=square, mark size=0.5mm]
              table[x expr=\thisrow{S},y=AN] {./tikz/ImageNet/Transfer/SemHF/Nonlinear_Detail_alexnet.txt}; 
     
           \end{axis}
          \end{tikzpicture}\hspace*{-0.6em}  
          \begin{tikzpicture}
          \begin{axis}[cycle list name=color list,
                  footnotesize,
                  axis lines=left, 
                  width=2cm,
                  height=4cm,
                  ymin=0.00,
                  ymax=1.05,
                  enlarge y limits=0.01,
                  enlarge x limits=0.02,
                  xlabel={\footnotesize N},
                  ytick={0,.2,.4,.6,.8, 1},
                  yticklabels={0,.2,.4,.6,.8, 1},
                  xticklabels={1}
                  ]
          \addplot+[R50, only marks,
            mark=triangle, mark size=0.5mm]
              table[x expr=\thisrow{S},y=R50] {./tikz/ImageNet/Transfer/LogDNN_alexnet.txt};
          \addplot+[R50, only marks,
            mark=square, mark size=0.5mm]
              table[x expr=\thisrow{S},y=R50] {./tikz/ImageNet/Transfer/SemHF/Log_alexnet.txt};      
          
          \addplot+[R18, only marks,
            mark=triangle, mark size=0.5mm]
              table[x expr=\thisrow{S},y=R18] {./tikz/ImageNet/Transfer/LogDNN_alexnet.txt}; 
          \addplot+[R18, only marks,
            mark=square, mark size=0.5mm]
              table[x expr=\thisrow{S},y=R18] {./tikz/ImageNet/Transfer/SemHF/Log_alexnet.txt};      
              
          \addplot+[AN, only marks,
            mark=triangle, mark size=0.5mm]
              table[x expr=\thisrow{S},y=AN] {./tikz/ImageNet/Transfer/LogDNN_alexnet.txt}; \label{cap:FF-c-p}
          \addplot+[AN, only marks,
            mark=square, mark size=0.5mm]
              table[x expr=\thisrow{S},y=AN] {./tikz/ImageNet/Transfer/SemHF/Log_alexnet.txt};  \label{cap:FF-p}     
           \end{axis}
          \end{tikzpicture} 
          
           \caption{The categorical success rate and transferability of FilterFool (\ref{cap:FF-l}~or~\ref{cap:FF-p}) and FilterFool-c (\ref{cap:FF-c-l}~or~\ref{cap:FF-c-p}) to evaluate the effect of type and strength of selected filters, namely Nonlinear Detail enhancement (ND), Log Transformation (LT), Gamma Correction (GC) with strength $\gamma$ and Linear Detail enhancement (LD) with strength $\alpha$, as well as the proposed semantic adversarial loss function. Adversarial images are generated against {\color{R50} ResNet50}, {\color{R18} ResNet18} and {\color{AN} AlexNet} in the first, second and third rows, respectively. Hence, in each plot, the top two overlapped lines (or points) show the categorical success rate when the classifier is seen, while other four bottom lines (or points) show the transferability against unseen classifiers. Note that $\gamma=1$ in the Gamma correction plots corresponds to the original images.}
            \label{fig:EnhTransAnalyse}
\end{figure}

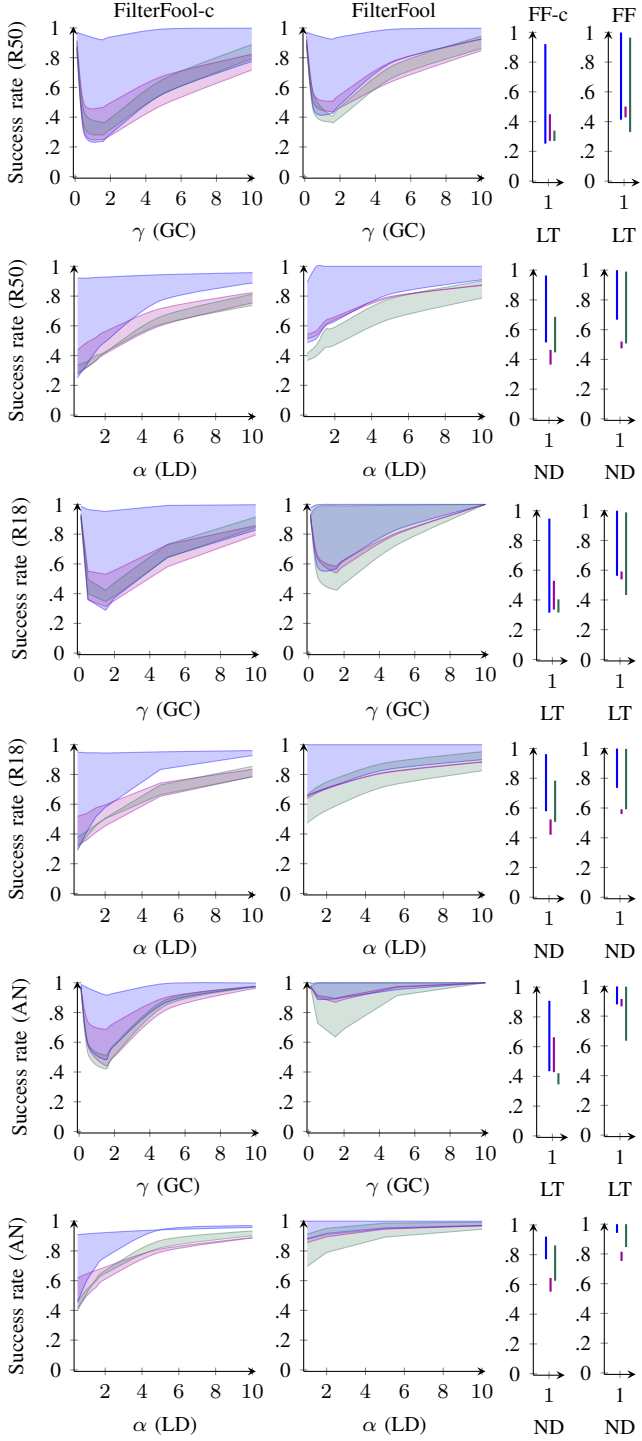
\begin{figure}[t!]
          \centering
          
          \begin{tikzpicture}
          \begin{axis}[cycle list name=color list,
                  footnotesize,
                  axis lines=left, 
                  width=4cm,
                  height=3.6cm,
                  ymin=0.00,
                  ymax=1.00,
                  title=\footnotesize FilterFool-c,
                  enlarge y limits=0.01,
                  enlarge x limits=0.02,
                  smooth,
                  xlabel={\footnotesize{$\gamma$ (GC)}},
                  ylabel={\footnotesize Success rate (R50)},
                  y tick label style={
                  /pgf/number format/.cd,
                  fixed,
                  fixed zerofill,
                  precision=2,
                  /tikz/.cd
                  },
                  ytick={0,.2,.4,.6,.8, 1},
                  yticklabels={0,.2,.4,.6,.8, 1}
                  ]
            
              \addplot+[BR,opacity=0.4,name path=B1] table[x=S,y={B1}]{./tikz/ImageNet/Robust/GammaDNN_resnet50.txt};\label{AnalysisRobustB}
            \addplot+[BR,opacity=0.4,name path=B4] table[x=S,y={B4}]{./tikz/ImageNet/Robust/GammaDNN_resnet50.txt};
            \addplot[fill=BR,opacity=0.2] fill between[of=B1 and B4];
            
              \addplot+[MR,opacity=0.4,name path=M2] table[x expr=\thisrow{S},y={M2}]{./tikz/ImageNet/Robust/GammaDNN_resnet50.txt};\label{AnalysisRobustM}
             \addplot+[MR,opacity=0.4,name path=M5] table[x expr=\thisrow{S},y={M5}]{./tikz/ImageNet/Robust/GammaDNN_resnet50.txt};
            \addplot[fill=MR,opacity=0.2] fill between[of=M2 and M5];

              \addplot+[JR,opacity=0.4,name path=J25] table[x expr=\thisrow{S},y={J25}]{./tikz/ImageNet/Robust/GammaDNN_resnet50.txt};\label{AnalysisRobustJ}
              \addplot+[JR,opacity=0.4,name path=J75] table[x expr=\thisrow{S},y={J75}]{./tikz/ImageNet/Robust/GammaDNN_resnet50.txt};
              \addplot[fill=JR,opacity=0.2] fill between[of=J25 and J75];

           \end{axis}
          \end{tikzpicture}\hspace*{-0.6em}  
        \begin{tikzpicture}
          \begin{axis}[cycle list name=color list,
                  footnotesize,
                  axis lines=left, 
                  width=4cm,
                  height=3.6cm,
                  ymin=0.00,
                  ymax=1.00,
                  title=\footnotesize FilterFool,
                  enlarge y limits=0.01,
                  enlarge x limits=0.02,
                  xlabel={\footnotesize $\gamma$ (GC)},
                  smooth,
                  y tick label style={
                  /pgf/number format/.cd,
                  fixed,
                  fixed zerofill,
                  precision=2,
                  /tikz/.cd
                  },
                  ytick={0,.2,.4,.6,.8, 1},
                  yticklabels={0,.2,.4,.6,.8, 1}
                  ]
            
              \addplot+[BR,name path=B1,opacity=0.4] table[x=S,y={B1}]{./tikz/ImageNet/Robust/SemHF/Gamma_resnet50.txt};
              \addplot+[BR,name path=B3,opacity=0.4] table[x=S,y={B3}]{./tikz/ImageNet/Robust/SemHF/Gamma_resnet50.txt};
              \addplot[fill=BR,opacity=0.2] fill between[of=B3 and B1];

              \addplot+[MR,name path=M3,opacity=0.4] table[x expr=\thisrow{S},y={M3}]{./tikz/ImageNet/Robust/SemHF/Gamma_resnet50.txt};
              \addplot+[MR,name path=M5,opacity=0.4] table[x expr=\thisrow{S},y={M5}]{./tikz/ImageNet/Robust/SemHF/Gamma_resnet50.txt};
            \addplot[fill=MR,opacity=0.2] fill between[of=M3 and M5];
            
             \addplot+[JR,name path=J100,opacity=0.4] table[x expr=\thisrow{S},y={J100}]{./tikz/ImageNet/Robust/SemHF/Gamma_resnet50.txt};
             \addplot+[JR,name path=J75,opacity=0.4] table[x expr=\thisrow{S},y={J75}]{./tikz/ImageNet/Robust/SemHF/Gamma_resnet50.txt};
             \addplot[fill=JR,opacity=0.2] fill between[of=J100 and J75];
             
           \end{axis}
          \end{tikzpicture}\hspace*{-0.6em}  
        \begin{tikzpicture}
          \begin{axis}[cycle list name=color list,
                  footnotesize,
                  axis lines=left, 
                  width=2cm,
                  height=3.6cm,
                  xmin=0.5,
                  xmax=1.5,
                  ymin=0.00,
                  ymax=1.00,
                  title=\footnotesize FF-c,
                  xlabel={\footnotesize LT},
                  enlarge y limits=0.01,
                  enlarge x limits=0.02,
                  y tick label style={
                  /pgf/number format/.cd,
                  fixed,
                  fixed zerofill,
                  precision=2,
                  /tikz/.cd
                  },
                  ytick={0,.2,.4,.6,.8, 1},
                  yticklabels={0,.2,.4,.6,.8, 1},
                  xtick={1}
                  ]
            \addplot +[BR,mark=none,line width=0.8pt] coordinates {(0.9, 0.2517) (0.9, 0.9210)};
            \addplot +[MR,mark=none,line width=0.8pt] coordinates {(1.05, 0.2703) (1.05, 0.4497)};
            \addplot +[JR,mark=none,line width=0.8pt] coordinates {(1.20, 0.2697) (1.20,  0.3393)};
           \end{axis}
          \end{tikzpicture}\hspace*{-0.6em} 
          \begin{tikzpicture}
          \begin{axis}[cycle list name=color list,
                  footnotesize,
                  axis lines=left, 
                  width=2cm,
                  height=3.6cm,
                  xmin=0.5,
                  xmax=1.5,
                  ymin=0.00,
                  ymax=1.00,
                  title=\footnotesize FF,
                  xlabel={\footnotesize LT},
                  enlarge y limits=0.01,
                  enlarge x limits=0.02,
                  y tick label style={
                  /pgf/number format/.cd,
                  fixed,
                  fixed zerofill,
                  precision=2,
                  /tikz/.cd
                  },
                  ytick={0,.2,.4,.6,.8, 1},
                  yticklabels={0,.2,.4,.6,.8, 1},
                  xtick={1}
                  ]
            \addplot +[BR,mark=none,line width=0.8pt] coordinates {(0.9, 0.4128) (0.9, 0.9989)};
            \addplot +[MR,mark=none,line width=0.8pt] coordinates {(1.05, 0.4283) (1.05, 0.5011)};
            \addplot +[JR,mark=none,line width=0.8pt] coordinates {(1.20, 0.3296) (1.20,  0.9645)};
           \end{axis}
          \end{tikzpicture}

          \begin{tikzpicture}
          \begin{axis}[cycle list name=color list,
                  footnotesize,
                  axis lines=left, 
                  width=4cm,
                  height=3.6cm,
                  ymin=0.00,
                  ymax=1.00,
                  enlarge y limits=0.01,
                  enlarge x limits=0.02,
                  smooth,
                  xlabel={\footnotesize $\alpha$ (LD)},
                  ylabel={\footnotesize Success rate (R50)},
                  y tick label style={
                  /pgf/number format/.cd,
                  fixed,
                  fixed zerofill,
                  precision=2,
                  /tikz/.cd
                  },
                  ytick={0,.2,.4,.6,.8, 1},
                  yticklabels={0,.2,.4,.6,.8, 1}
                  ]
            
              \addplot+[BR,name path=B1,opacity=0.4] table[x=S,y={B1}]{./tikz/ImageNet/Robust/UnsharpDNN_resnet50.txt};
              \addplot+[BR,name path=B4,opacity=0.4] table[x=S,y={B4}]{./tikz/ImageNet/Robust/UnsharpDNN_resnet50.txt};
              \addplot[fill=BR,opacity=0.2] fill between[of=B1 and B4];

              \addplot+[MR,name path=M2,opacity=0.4] table[x expr=\thisrow{S},y={M2}]{./tikz/ImageNet/Robust/UnsharpDNN_resnet50.txt};
              \addplot+[MR,name path=M5,opacity=0.4] table[x expr=\thisrow{S},y={M5}]{./tikz/ImageNet/Robust/UnsharpDNN_resnet50.txt};
            \addplot[fill=MR,opacity=0.2] fill between[of=M2 and M5];
            
             \addplot+[JR,name path=J25,opacity=0.4] table[x expr=\thisrow{S},y={J25}]{./tikz/ImageNet/Robust/UnsharpDNN_resnet50.txt};
             \addplot+[JR,name path=J75,opacity=0.4] table[x expr=\thisrow{S},y={J75}]{./tikz/ImageNet/Robust/UnsharpDNN_resnet50.txt};
             \addplot[fill=JR,opacity=0.2] fill between[of=J25 and J75];

           \end{axis}
          \end{tikzpicture}\hspace*{-0.6em}  
        \begin{tikzpicture}
          \begin{axis}[cycle list name=color list,
                  footnotesize,
                  axis lines=left, 
                  width=4cm,
                  height=3.6cm,
                  ymin=0.00,
                  ymax=1.00,
                  enlarge y limits=0.01,
                  enlarge x limits=0.02,
                  xlabel={\footnotesize $\alpha$ (LD)},
                  smooth,
                  y tick label style={
                  /pgf/number format/.cd,
                  fixed,
                  fixed zerofill,
                  precision=2,
                  /tikz/.cd
                  },
                  ytick={0,.2,.4,.6,.8, 1},
                  yticklabels={0,.2,.4,.6,.8, 1}
                  ]
            
              \addplot+[BR,name path=B1,opacity=0.4] table[x=S,y={B3}]{./tikz/ImageNet/Robust/SemHF/Linear_Detail_resnet50.txt};
              \addplot+[BR,name path=B4,opacity=0.4] table[x=S,y={B7}]{./tikz/ImageNet/Robust/SemHF/Linear_Detail_resnet50.txt};
              \addplot[fill=BR,opacity=0.2] fill between[of=B1 and B4];

              \addplot+[MR,name path=M2,opacity=0.4] table[x expr=\thisrow{S},y={M3}]{./tikz/ImageNet/Robust/SemHF/Linear_Detail_resnet50.txt};
              \addplot+[MR,name path=M5,opacity=0.4] table[x expr=\thisrow{S},y={M5}]{./tikz/ImageNet/Robust/SemHF/Linear_Detail_resnet50.txt};
            \addplot[fill=MR,opacity=0.2] fill between[of=M2 and M5];
            
             \addplot+[JR,name path=J25,opacity=0.4] table[x expr=\thisrow{S},y={J25}]{./tikz/ImageNet/Robust/SemHF/Linear_Detail_resnet50.txt};
             \addplot+[JR,name path=J75,opacity=0.4] table[x expr=\thisrow{S},y={J75}]{./tikz/ImageNet/Robust/SemHF/Linear_Detail_resnet50.txt};
             \addplot[fill=JR,opacity=0.2] fill between[of=J25 and J75];
             
           \end{axis}
          \end{tikzpicture}\hspace*{-0.6em}  
        \begin{tikzpicture}
          \begin{axis}[cycle list name=color list,
                  footnotesize,
                  axis lines=left, 
                  width=2cm,
                  height=3.6cm,
                  xmin=0.5,
                  xmax=1.5,
                  ymin=0.00,
                  ymax=1.00,
                  xlabel={\footnotesize ND},
                  enlarge y limits=0.01,
                  enlarge x limits=0.02,
                  y tick label style={
                  /pgf/number format/.cd,
                  fixed,
                  fixed zerofill,
                  precision=2,
                  /tikz/.cd
                  },
                  ytick={0,.2,.4,.6,.8, 1},
                  yticklabels={0,.2,.4,.6,.8, 1},
                  xtick={1}
                  ]
            \addplot +[BR,mark=none,line width=0.8pt] coordinates {(0.9, 0.5143) (0.9, 0.9643)};
            \addplot +[MR,mark=none,line width=0.8pt] coordinates {(1.05, 0.3647) (1.05, 0.4643)};
            \addplot +[JR,mark=none,line width=0.8pt] coordinates {(1.20, 0.4473) (1.20,  0.6863)};
           \end{axis}
          \end{tikzpicture}\hspace*{-0.6em} 
          \begin{tikzpicture}
          \begin{axis}[cycle list name=color list,
                  footnotesize,
                  axis lines=left, 
                  width=2cm,
                  height=3.6cm,
                  xmin=0.5,
                  xmax=1.5,
                  ymin=0.00,
                  ymax=1.00,
                  xlabel={\footnotesize ND},
                  enlarge y limits=0.01,
                  enlarge x limits=0.02,
                  y tick label style={
                  /pgf/number format/.cd,
                  fixed,
                  fixed zerofill,
                  precision=2,
                  /tikz/.cd
                  },
                  ytick={0,.2,.4,.6,.8, 1},
                  yticklabels={0,.2,.4,.6,.8, 1},
                  xtick={1}
                  ]
            \addplot +[BR,mark=none,line width=0.8pt] coordinates {(0.9, 0.6670) (0.9, 0.9997)};
            \addplot +[MR,mark=none,line width=0.8pt] coordinates {(1.05, 0.4741) (1.05, 0.5202)};
            \addplot +[JR,mark=none,line width=0.8pt] coordinates {(1.20, 0.5072) (1.20,  0.9900)};
           \end{axis}
          \end{tikzpicture}          
          
\begin{tikzpicture}
          \begin{axis}[cycle list name=color list,
                  footnotesize,
                  axis lines=left, 
                  width=4cm,
                  height=3.6cm,
                  ymin=0.00,
                  ymax=1.00,
                  enlarge y limits=0.01,
                  enlarge x limits=0.02,
                  xlabel={\footnotesize $\gamma$ (GC}),
                  ylabel={\footnotesize Success rate (R18)},
                  y tick label style={
                  /pgf/number format/.cd,
                  fixed,
                  fixed zerofill,
                  precision=2,
                  /tikz/.cd
                  },
                  ytick={0,.2,.4,.6,.8, 1},
                  yticklabels={0,.2,.4,.6,.8, 1}
                  ]
            
              \addplot+[BR,name path=B1,opacity=0.4] table[x=S,y={B1}]{./tikz/ImageNet/Robust/GammaDNN_resnet18.txt};
              \addplot+[BR,name path=B4,opacity=0.4] table[x=S,y={B4}]{./tikz/ImageNet/Robust/GammaDNN_resnet18.txt};
              \addplot[fill=BR,opacity=0.2] fill between[of=B1 and B4];
            
              \addplot+[MR,name path=M2,opacity=0.4] table[x expr=\thisrow{S},y={M2}]{./tikz/ImageNet/Robust/GammaDNN_resnet18.txt};
              \addplot+[MR,name path=M5,opacity=0.4] table[x expr=\thisrow{S},y={M5}]{./tikz/ImageNet/Robust/GammaDNN_resnet18.txt};
               \addplot[fill=MR,opacity=0.2] fill between[of=M2 and M5];
            
              \addplot+[JR,name path=J25,opacity=0.4] table[x expr=\thisrow{S},y={J25}]{./tikz/ImageNet/Robust/GammaDNN_resnet18.txt};
              \addplot+[JR,name path=J75,opacity=0.4] table[x expr=\thisrow{S},y={J75}]{./tikz/ImageNet/Robust/GammaDNN_resnet18.txt};
              \addplot[fill=JR,opacity=0.2] fill between[of=J25 and J75];
              
           \end{axis}
          \end{tikzpicture}\hspace*{-0.6em} 
        \begin{tikzpicture}
          \begin{axis}[cycle list name=color list,
                  footnotesize,
                  axis lines=left, 
                  width=4cm,
                  height=3.6cm,
                  ymin=0.00,
                  ymax=1.00,
                  enlarge y limits=0.01,
                  enlarge x limits=0.02,
                  smooth,
                  xlabel={\footnotesize $\gamma$ (GC}),
                  y tick label style={
                  /pgf/number format/.cd,
                  fixed,
                  fixed zerofill,
                  precision=2,
                  /tikz/.cd
                  },
                  ytick={0,.2,.4,.6,.8, 1},
                  yticklabels={0,.2,.4,.6,.8, 1}
                  ]
             \addplot+[BR,name path=B3,opacity=0.4] table[x=S,y={B3}]{./tikz/ImageNet/Robust/SemHF/Gamma_resnet18.txt};
              \addplot+[BR,name path=B7,opacity=0.4] table[x=S,y={B7}]{./tikz/ImageNet/Robust/SemHF/Gamma_resnet18.txt};
              \addplot[fill=BR,opacity=0.2] fill between[of=B3 and B7];
            
              \addplot+[MR,name path=M3,opacity=0.4] table[x expr=\thisrow{S},y={M3}]{./tikz/ImageNet/Robust/SemHF/Gamma_resnet18.txt};
              \addplot+[MR,name path=M5,opacity=0.4] table[x expr=\thisrow{S},y={M5}]{./tikz/ImageNet/Robust/SemHF/Gamma_resnet18.txt};
               \addplot[fill=MR,opacity=0.2] fill between[of=M3 and M5];
            
              \addplot+[JR,name path=J50,opacity=0.4] table[x expr=\thisrow{S},y={J50}]{./tikz/ImageNet/Robust/SemHF/Gamma_resnet18.txt};
              \addplot+[JR,name path=J100,opacity=0.4] table[x expr=\thisrow{S},y={J100}]{./tikz/ImageNet/Robust/SemHF/Gamma_resnet18.txt};
              \addplot[fill=JR,opacity=0.2] fill between[of=J50 and J100];
           \end{axis}
          \end{tikzpicture}\hspace*{-0.6em}  
        \begin{tikzpicture}
          \begin{axis}[cycle list name=color list,
                  footnotesize,
                  axis lines=left, 
                  width=2cm,
                  height=3.6cm,
                  xmin=0.5,
                  xmax=1.5,
                  ymin=0.00,
                  ymax=1.00,
                  xlabel={\footnotesize LT},
                  enlarge y limits=0.01,
                  enlarge x limits=0.02,
                  y tick label style={
                  /pgf/number format/.cd,
                  fixed,
                  fixed zerofill,
                  precision=2,
                  /tikz/.cd
                  },
                  ytick={0,.2,.4,.6,.8, 1},
                  yticklabels={0,.2,.4,.6,.8, 1},
                  xtick={1}
                  ]
            \addplot +[BR,mark=none,line width=0.8pt] coordinates {(0.9, 0.3143) (0.9, 0.9470)};
            \addplot +[MR,mark=none,line width=0.8pt] coordinates {(1.05, 0.3353) (1.05, 0.5287)};
            \addplot +[JR,mark=none,line width=0.8pt] coordinates {(1.20, 0.3157) (1.20, 0.4040)};
           \end{axis}
          \end{tikzpicture}\hspace*{-0.6em} 
          \begin{tikzpicture}
          \begin{axis}[cycle list name=color list,
                  footnotesize,
                  axis lines=left, 
                  width=2cm,
                  height=3.6cm,
                  xmin=0.5,
                  xmax=1.5,
                  ymin=0.00,
                  ymax=1.00,
                  xlabel={\footnotesize LT},
                  enlarge y limits=0.01,
                  enlarge x limits=0.02,
                  y tick label style={
                  /pgf/number format/.cd,
                  fixed,
                  fixed zerofill,
                  precision=2,
                  /tikz/.cd
                  },
                  ytick={0,.2,.4,.6,.8, 1},
                  yticklabels={0,.2,.4,.6,.8, 1},
                  xtick={1}
                  ]
            \addplot +[BR,mark=none,line width=0.8pt] coordinates {(0.9, 0.5623) (0.9, 0.9997)};
            \addplot +[MR,mark=none,line width=0.8pt] coordinates {(1.05, 0.5380) (1.05, 0.5913)};
            \addplot +[JR,mark=none,line width=0.8pt] coordinates {(1.20, 0.4333) (1.20, 0.9890)};
           \end{axis}
          \end{tikzpicture}

\begin{tikzpicture}
          \begin{axis}[cycle list name=color list,
                  footnotesize,
                  axis lines=left, 
                  width=4cm,
                  height=3.6cm,
                  ymin=0.00,
                  ymax=1.00,
                  enlarge y limits=0.01,
                  enlarge x limits=0.02,
                  xlabel={\footnotesize $\alpha$ (LD)},
                  ylabel={\footnotesize Success rate (R18)},
                  y tick label style={
                  /pgf/number format/.cd,
                  fixed,
                  fixed zerofill,
                  precision=2,
                  /tikz/.cd
                  },
                  ytick={0,.2,.4,.6,.8, 1},
                  yticklabels={0,.2,.4,.6,.8, 1}
                  ]
            
             \addplot+[BR,name path=B1,opacity=0.4] table[x=S,y={B1}]{./tikz/ImageNet/Robust/UnsharpDNN_resnet18.txt};
              \addplot+[BR,name path=B4,opacity=0.4] table[x=S,y={B4}]{./tikz/ImageNet/Robust/UnsharpDNN_resnet18.txt};
              \addplot[fill=BR,opacity=0.2] fill between[of=B1 and B4];
            
              \addplot+[MR,name path=M2,opacity=0.4] table[x expr=\thisrow{S},y={M2}]{./tikz/ImageNet/Robust/UnsharpDNN_resnet18.txt};
              \addplot+[MR,name path=M5,opacity=0.4] table[x expr=\thisrow{S},y={M5}]{./tikz/ImageNet/Robust/UnsharpDNN_resnet18.txt};
               \addplot[fill=MR,opacity=0.2] fill between[of=M2 and M5];
            
              \addplot+[JR,name path=J25,opacity=0.4] table[x expr=\thisrow{S},y={J25}]{./tikz/ImageNet/Robust/UnsharpDNN_resnet18.txt};
              \addplot+[JR,name path=J75,opacity=0.4] table[x expr=\thisrow{S},y={J75}]{./tikz/ImageNet/Robust/UnsharpDNN_resnet18.txt};
              \addplot[fill=JR,opacity=0.2] fill between[of=J25 and J75];
              
           \end{axis}
          \end{tikzpicture}\hspace*{-0.6em} 
        \begin{tikzpicture}
          \begin{axis}[cycle list name=color list,
                  footnotesize,
                  axis lines=left, 
                  width=4cm,
                  height=3.6cm,
                  ymin=0.00,
                  ymax=1.00,
                  enlarge y limits=0.01,
                  enlarge x limits=0.02,
                  smooth,
                  xlabel={\footnotesize $\alpha$ (LD)},
                  y tick label style={
                  /pgf/number format/.cd,
                  fixed,
                  fixed zerofill,
                  precision=2,
                  /tikz/.cd
                  },
                  ytick={0,.2,.4,.6,.8, 1},
                  yticklabels={0,.2,.4,.6,.8, 1}
                  ]
             \addplot+[BR,name path=B2,opacity=0.4] table[x=S,y={B2}]{./tikz/ImageNet/Robust/SemHF/Linear_Detail_resnet18.txt};
              \addplot+[BR,name path=B7,opacity=0.4] table[x=S,y={B7}]{./tikz/ImageNet/Robust/SemHF/Linear_Detail_resnet18.txt};
              \addplot[fill=BR,opacity=0.2] fill between[of=B2 and B7];
            
              \addplot+[MR,name path=M3,opacity=0.4] table[x expr=\thisrow{S},y={M3}]{./tikz/ImageNet/Robust/SemHF/Linear_Detail_resnet18.txt};
              \addplot+[MR,name path=M5,opacity=0.4] table[x expr=\thisrow{S},y={M5}]{./tikz/ImageNet/Robust/SemHF/Linear_Detail_resnet18.txt};
               \addplot[fill=MR,opacity=0.2] fill between[of=M3 and M5];
            
              \addplot+[JR,name path=J25,opacity=0.4] table[x expr=\thisrow{S},y={J25}]{./tikz/ImageNet/Robust/SemHF/Linear_Detail_resnet18.txt};
              \addplot+[JR,name path=J75,opacity=0.4] table[x expr=\thisrow{S},y={J75}]{./tikz/ImageNet/Robust/SemHF/Linear_Detail_resnet18.txt};
              \addplot[fill=JR,opacity=0.2] fill between[of=J25 and J75];
           \end{axis}
          \end{tikzpicture}\hspace*{-0.6em}  
        \begin{tikzpicture}
          \begin{axis}[cycle list name=color list,
                  footnotesize,
                  axis lines=left, 
                  width=2cm,
                  height=3.6cm,
                  xmin=0.5,
                  xmax=1.5,
                  ymin=0.00,
                  ymax=1.00,
                  xlabel={\footnotesize ND},
                  enlarge y limits=0.01,
                  enlarge x limits=0.02,
                  y tick label style={
                  /pgf/number format/.cd,
                  fixed,
                  fixed zerofill,
                  precision=2,
                  /tikz/.cd
                  },
                  ytick={0,.2,.4,.6,.8, 1},
                  yticklabels={0,.2,.4,.6,.8, 1},
                  xtick={1}
                  ]
            \addplot +[BR,mark=none,line width=0.8pt] coordinates {(0.9, 0.5783) (0.9, 0.9620)};
            \addplot +[MR,mark=none,line width=0.8pt] coordinates {(1.05, 0.4203) (1.05, 0.5233)};
            \addplot +[JR,mark=none,line width=0.8pt] coordinates {(1.20, 0.507) (1.20, 0.7840)};
           \end{axis}
          \end{tikzpicture}\hspace*{-0.6em} 
          \begin{tikzpicture}
          \begin{axis}[cycle list name=color list,
                  footnotesize,
                  axis lines=left, 
                  width=2cm,
                  height=3.6cm,
                  xmin=0.5,
                  xmax=1.5,
                  ymin=0.00,
                  ymax=1.00,
                  xlabel={\footnotesize ND},
                  enlarge y limits=0.01,
                  enlarge x limits=0.02,
                  y tick label style={
                  /pgf/number format/.cd,
                  fixed,
                  fixed zerofill,
                  precision=2,
                  /tikz/.cd
                  },
                  ytick={0,.2,.4,.6,.8, 1},
                  yticklabels={0,.2,.4,.6,.8, 1},
                  xtick={1}
                  ]
            \addplot +[BR,mark=none,line width=0.8pt] coordinates {(0.9, 0.7354) (0.9, 0.9996)};
            \addplot +[MR,mark=none,line width=0.8pt] coordinates {(1.05, 0.5601) (1.05, 0.5918)};
            \addplot +[JR,mark=none,line width=0.8pt] coordinates {(1.20, 0.5918) (1.20, 0.9978)};
           \end{axis}
          \end{tikzpicture}          
          
\begin{tikzpicture}
          \begin{axis}[cycle list name=color list,
                  footnotesize,
                  axis lines=left, 
                  width=4cm,
                  height=3.6cm,
                  ymin=0.00,
                  ymax=1.00,
                  smooth,
                  enlarge y limits=0.01,
                  enlarge x limits=0.02,
                  xlabel={\footnotesize $\gamma$ (GC)},
                  ylabel={\footnotesize Success rate (AN)},
                  y tick label style={
                  /pgf/number format/.cd,
                  fixed,
                  fixed zerofill,
                  precision=2,
                  /tikz/.cd
                  },
                  ytick={0,.2,.4,.6,.8, 1},
                  yticklabels={0,.2,.4,.6,.8, 1}
                  ]
             \addplot+[BR,name path=B1,opacity=0.4] table[x=S,y={B1}]{./tikz/ImageNet/Robust/GammaDNN_alexnet.txt};
              \addplot+[BR,name path=B4,opacity=0.4] table[x=S,y={B4}]{./tikz/ImageNet/Robust/GammaDNN_alexnet.txt};
              \addplot[fill=BR,opacity=0.2] fill between[of=B1 and B4];
            
              \addplot+[MR,name path=M2,opacity=0.4] table[x expr=\thisrow{S},y={M2}]{./tikz/ImageNet/Robust/GammaDNN_alexnet.txt};
              \addplot+[MR,name path=M5,opacity=0.4] table[x expr=\thisrow{S},y={M5}]{./tikz/ImageNet/Robust/GammaDNN_alexnet.txt};
               \addplot[fill=MR,opacity=0.2] fill between[of=M2 and M5];
            
              \addplot+[JR,name path=J25,opacity=0.4] table[x expr=\thisrow{S},y={J25}]{./tikz/ImageNet/Robust/GammaDNN_alexnet.txt};
              \addplot+[JR,name path=J75,opacity=0.4] table[x expr=\thisrow{S},y={J75}]{./tikz/ImageNet/Robust/GammaDNN_alexnet.txt};
              \addplot[fill=JR,opacity=0.2] fill between[of=J25 and J75];
           \end{axis}
          \end{tikzpicture}\hspace*{-0.6em}  
        \begin{tikzpicture}
          \begin{axis}[cycle list name=color list,
                  footnotesize,
                  axis lines=left, 
                  width=4cm,
                  height=3.6cm,
                  ymin=0.00,
                  ymax=1.00,
                  enlarge y limits=0.01,
                  enlarge x limits=0.02,
                  xlabel={\footnotesize $\gamma$ (GC)},
                  y tick label style={
                  /pgf/number format/.cd,
                  fixed,
                  fixed zerofill,
                  precision=2,
                  /tikz/.cd
                  },
                  ytick={0,.2,.4,.6,.8, 1},
                  yticklabels={0,.2,.4,.6,.8, 1}
                  ]
              \addplot+[BR,name path=B2,opacity=0.4] table[x=S,y={B2}]{./tikz/ImageNet/Robust/SemHF/Gamma_alexnet.txt};
              \addplot+[BR,name path=B7,opacity=0.4] table[x=S,y={B7}]{./tikz/ImageNet/Robust/SemHF/Gamma_alexnet.txt};
              \addplot[fill=BR,opacity=0.2] fill between[of=B2 and B7];
            
              \addplot+[MR,name path=M2,opacity=0.4] table[x expr=\thisrow{S},y={M2}]{./tikz/ImageNet/Robust/SemHF/Gamma_alexnet.txt};
              \addplot+[MR,name path=M3,opacity=0.4] table[x expr=\thisrow{S},y={M3}]{./tikz/ImageNet/Robust/SemHF/Gamma_alexnet.txt};
               \addplot[fill=MR,opacity=0.2] fill between[of=M2 and M3];
            
              \addplot+[JR,name path=J25,opacity=0.4] table[x expr=\thisrow{S},y={J25}]{./tikz/ImageNet/Robust/SemHF/Gamma_alexnet.txt};
              \addplot+[JR,name path=J100,opacity=0.4] table[x expr=\thisrow{S},y={J100}]{./tikz/ImageNet/Robust/SemHF/Gamma_alexnet.txt};
              \addplot[fill=JR,opacity=0.2] fill between[of=J25 and J100];
           \end{axis}
          \end{tikzpicture}\hspace*{-0.6em}  
        \begin{tikzpicture}
          \begin{axis}[cycle list name=color list,
                  footnotesize,
                  axis lines=left, 
                  width=2cm,
                  height=3.6cm,
                  xmin=0.5,
                  xmax=1.5,
                  ymin=0.00,
                  ymax=1.00,
                  xlabel=\footnotesize LT,
                  enlarge y limits=0.01,
                  enlarge x limits=0.02,
                  y tick label style={
                  /pgf/number format/.cd,
                  fixed,
                  fixed zerofill,
                  precision=2,
                  /tikz/.cd
                  },
                  ytick={0,.2,.4,.6,.8, 1},
                  yticklabels={0,.2,.4,.6,.8, 1},
                  xtick={1}
                  ]
            \addplot +[BR,mark=none,line width=0.8pt] coordinates {(0.9, 0.4350) (0.9, 0.9087)};
            \addplot +[MR,mark=none,line width=0.8pt] coordinates {(1.05, 0.4287) (1.05, 0.6633)};
            \addplot +[JR,mark=none,line width=0.8pt] coordinates {(1.20, 0.3463) (1.20, 0.4217)};
           \end{axis}
          \end{tikzpicture}\hspace*{-0.6em} 
          \begin{tikzpicture}
          \begin{axis}[cycle list name=color list,
                  footnotesize,
                  axis lines=left, 
                  width=2cm,
                  height=3.6cm,
                  xmin=0.5,
                  xmax=1.5,
                  ymin=0.00,
                  ymax=1.00,
                  xlabel= \footnotesize LT,
                  enlarge y limits=0.01,
                  enlarge x limits=0.02,
                  y tick label style={
                  /pgf/number format/.cd,
                  fixed,
                  fixed zerofill,
                  precision=2,
                  /tikz/.cd
                  },
                  ytick={0,.2,.4,.6,.8, 1},
                  yticklabels={0,.2,.4,.6,.8, 1},
                  xticklabels={1},
                  xtick={1}
                  ]
            \addplot +[BR,mark=none,line width=0.8pt] coordinates {(0.9, 0.8790) (0.9, 0.9996)};
            \addplot +[MR,mark=none,line width=0.8pt] coordinates {(1.05, 0.8656) (1.05, 0.9157)};
            \addplot +[JR,mark=none,line width=0.8pt] coordinates {(1.20, 0.6347) (1.20, 0.9986)};

           \end{axis}
          \end{tikzpicture}
          
          \begin{tikzpicture}
          \begin{axis}[cycle list name=color list,
                  footnotesize,
                  axis lines=left, 
                  width=4cm,
                  height=3.6cm,
                  ymin=0.00,
                  ymax=1.00,
                  smooth,
                  enlarge y limits=0.01,
                  enlarge x limits=0.02,
                  xlabel={\footnotesize $\alpha$ (LD)},
                  ylabel={\footnotesize Success rate (AN)},
                  y tick label style={
                  /pgf/number format/.cd,
                  fixed,
                  fixed zerofill,
                  precision=2,
                  /tikz/.cd
                  },
                  ytick={0,.2,.4,.6,.8, 1},
                  yticklabels={0,.2,.4,.6,.8, 1}
                  ]
              \addplot+[BR,name path=B1,opacity=0.4] table[x=S,y={B1}]{./tikz/ImageNet/Robust/UnsharpDNN_alexnet.txt};
              \addplot+[BR,name path=B4,opacity=0.4] table[x=S,y={B4}]{./tikz/ImageNet/Robust/UnsharpDNN_alexnet.txt};
              \addplot[fill=BR,opacity=0.2] fill between[of=B1 and B4];
            
              \addplot+[MR,name path=M2,opacity=0.4] table[x expr=\thisrow{S},y={M2}]{./tikz/ImageNet/Robust/UnsharpDNN_alexnet.txt};
              \addplot+[MR,name path=M5,opacity=0.4] table[x expr=\thisrow{S},y={M5}]{./tikz/ImageNet/Robust/UnsharpDNN_alexnet.txt};
               \addplot[fill=MR,opacity=0.2] fill between[of=M2 and M5];
            
              \addplot+[JR,name path=J25,opacity=0.4] table[x expr=\thisrow{S},y={J25}]{./tikz/ImageNet/Robust/UnsharpDNN_alexnet.txt};
              \addplot+[JR,name path=J75,opacity=0.4] table[x expr=\thisrow{S},y={J75}]{./tikz/ImageNet/Robust/UnsharpDNN_alexnet.txt};
              \addplot[fill=JR,opacity=0.2] fill between[of=J25 and J75];
           \end{axis}
          \end{tikzpicture}\hspace*{-0.6em}  
        \begin{tikzpicture}
          \begin{axis}[cycle list name=color list,
                  footnotesize,
                  axis lines=left, 
                  width=4cm,
                  height=3.6cm,
                  ymin=0.00,
                  ymax=1.00,
                  enlarge y limits=0.01,
                  enlarge x limits=0.02,
                  xlabel={\footnotesize $\alpha$ (LD)},
                  y tick label style={
                  /pgf/number format/.cd,
                  fixed,
                  fixed zerofill,
                  precision=2,
                  /tikz/.cd
                  },
                  ytick={0,.2,.4,.6,.8, 1},
                  yticklabels={0,.2,.4,.6,.8, 1}
                  ]
              \addplot+[BR,name path=B2,opacity=0.4] table[x=S,y={B2}]{./tikz/ImageNet/Robust/SemHF/Linear_Detail_alexnet.txt};
              \addplot+[BR,name path=B7,opacity=0.4] table[x=S,y={B7}]{./tikz/ImageNet/Robust/SemHF/Linear_Detail_alexnet.txt};
              \addplot[fill=BR,opacity=0.2] fill between[of=B2 and B7];
            
              \addplot+[MR,name path=M2,opacity=0.4] table[x expr=\thisrow{S},y={M2}]{./tikz/ImageNet/Robust/SemHF/Linear_Detail_alexnet.txt};
              \addplot+[MR,name path=M3,opacity=0.4] table[x expr=\thisrow{S},y={M3}]{./tikz/ImageNet/Robust/SemHF/Linear_Detail_alexnet.txt};
               \addplot[fill=MR,opacity=0.2] fill between[of=M2 and M3];
            
              \addplot+[JR,name path=J25,opacity=0.4] table[x expr=\thisrow{S},y={J25}]{./tikz/ImageNet/Robust/SemHF/Linear_Detail_alexnet.txt};
              \addplot+[JR,name path=J75,opacity=0.4] table[x expr=\thisrow{S},y={J75}]{./tikz/ImageNet/Robust/SemHF/Linear_Detail_alexnet.txt};
              \addplot[fill=JR,opacity=0.2] fill between[of=J25 and J75];
           \end{axis}
          \end{tikzpicture}\hspace*{-0.6em}  
        \begin{tikzpicture}
          \begin{axis}[cycle list name=color list,
                  footnotesize,
                  axis lines=left, 
                  width=2cm,
                  height=3.6cm,
                  xmin=0.5,
                  xmax=1.5,
                  ymin=0.00,
                  ymax=1.00,
                  xlabel=\footnotesize ND,
                  enlarge y limits=0.01,
                  enlarge x limits=0.02,
                  y tick label style={
                  /pgf/number format/.cd,
                  fixed,
                  fixed zerofill,
                  precision=2,
                  /tikz/.cd
                  },
                  ytick={0,.2,.4,.6,.8, 1},
                  yticklabels={0,.2,.4,.6,.8, 1},
                  xtick={1}
                  ]
            \addplot +[BR,mark=none,line width=0.8pt] coordinates {(0.9, 0.7697) (0.9, 0.9217)};
            \addplot +[MR,mark=none,line width=0.8pt] coordinates {(1.05, 0.5517) (1.05, 0.6437)};
            \addplot +[JR,mark=none,line width=0.8pt] coordinates {(1.20, 0.6250) (1.20, 0.8623)};
           \end{axis}
          \end{tikzpicture}\hspace*{-0.6em} 
          \begin{tikzpicture}
          \begin{axis}[cycle list name=color list,
                  footnotesize,
                  axis lines=left, 
                  width=2cm,
                  height=3.6cm,
                  xmin=0.5,
                  xmax=1.5,
                  ymin=0.00,
                  ymax=1.00,
                  xlabel= \footnotesize ND,
                  enlarge y limits=0.01,
                  enlarge x limits=0.02,
                  y tick label style={
                  /pgf/number format/.cd,
                  fixed,
                  fixed zerofill,
                  precision=2,
                  /tikz/.cd
                  },
                  ytick={0,.2,.4,.6,.8, 1},
                  yticklabels={0,.2,.4,.6,.8, 1},
                  xticklabels={1},
                  xtick={1}
                  ]
            \addplot +[BR,mark=none,line width=0.8pt] coordinates {(0.9, 0.9415) (0.9, 1.0000)};
            \addplot +[MR,mark=none,line width=0.8pt] coordinates {(1.05, 0.7529) (1.05, 0.8149)};
            \addplot +[JR,mark=none,line width=0.8pt] coordinates {(1.2, 0.8462) (1.2, 1.0000)};

           \end{axis}
          \end{tikzpicture}
           \caption{Robustness of FilterFool-c and FilterFool to defence frameworks. The coloured areas are the categorical success rate of FilterFool-c and FilterFool in the presence of the defence frameworks that use (1-7) {\color{BR}bit} reduction, (2,3,5) {\color{MR} median} smoothing and (25,50,75,100) {\color{JR} JPEG} compression against ResNet50 (R50), ResNet18 (R18) and AlexNet (AN) for all four filters; Nonlinear Detail enhancement (ND), Log Transformation (LT), Gamma Correction (GC), Linear Detail enhancement (LD) with different strengths. $\gamma$ and $\alpha$ are the strengths of GC and LD that range from 0.1 to 10.}
            \label{fig:EnhRobustAnalyse}
\end{figure}

Figure~\ref{fig:EnhTransAnalyse} shows the effect of different filters used in FilterFool and the influence of their strengths on the categorical success rate and {\em transferability}. 
In each plot, the two top lines (or points) are the categorical success rate against the classifier that was used to generate the adversarial images. 
Four bottom lines (or points) show the categorical transferability to unseen classifiers.
FilterFool achieves a high categorical success rate for all three classifiers. As expected, the transferability increases with the strength of the filtering effect~\cite{hosseini2018semantic,shamsabadi2019colorfool}. 
We also consider here FilterFool-c, the categorical version of FilterFool, for which each categorical label is considered as a class on its own, i.e.~ $S=D$ and hence $\mathcal{L}_{\text{S-Adv}}(\cdot, \cdot)$ becomes the adversarial CW loss. The use of the semantic adversarial loss improves the categorical transferability by 8\%, as unseen classifiers may misclassify an image to a categorically different label which is, however, semantically similar. 

Figure~\ref{fig:EnhRobustAnalyse} shows the impact of the filters in FilterFool on the {\em robustness} against defences. Coloured regions represent the range of categorical success rate in the presence of various parameters of bit reduction, median filtering and JPEG compression, which are defined in Sec.~\ref{sec:para}. As for transferability, from the coloured region covered by linear detail enhancement and gamma correction, the robustness improves with the strength of filters. 
FilterFool is more robust than FilterFool-c as the semantic adversarial loss reduces the group of logits that are sematically related to the predicted label.

\begin{table}[t]
    \centering
    \caption{Categorical success rate and transferability on ImageNet.  }
        \setlength{\tabcolsep}{3pt}
       \begin{tabular}{|l|l|ccc|}
       
        \Xhline{3\arrayrulewidth}
        
        Attack & Classifier & $\rightarrow$ResNet50& $\rightarrow$ResNet18 & $\rightarrow$AlexNet\\

        \Xhline{3\arrayrulewidth} 
        \multirow{3}{*}{{BIM~\cite{kurakin2016adversarial}}} & {ResNet50} & 1.000 & 0.127 & 0.037  \\
         & {ResNet18} & 0.109 & 1.000 &0.041 \\
         & {AlexNet} & 0.038 &  0.050 &1.000 \\
         \hline
         \multirow{3}{*}{LL-FGSM~\cite{kurakin2016adversarial}} & ResNet50 & 0.850 & 0.483 & 0.414  \\
         & ResNet18 & 0.418 & 0.966 & 0.434 \\
         & AlexNet & 0.266 & 0.362 & 0.993  \\
         \hline
         \multirow{3}{*}{LL-BIM~\cite{kurakin2016adversarial}} & ResNet50 & 1.000 & 0.173 & 0.097  \\
         & ResNet18 & 0.129 & 1.000 & 0.100 \\
         & AlexNet & 0.117 & 0.153 & 0.998  \\
         \hline
         \multirow{3}{*}{P-FGSM~\cite{Li2019}} & ResNet50 & 1.000 & 0.233 & 0.172  \\
         & ResNet18 & 0.203 & 1.000 & 0.183 \\
         & AlexNet & 0.151 & 0.189 & 1.000  \\
         \hline
        \multirow{3}{*}{{DeepFool~\cite{MoosaviDezfooli16}}} & {ResNet50} & 0.983 & 0.071 & 0.018  \\
        & {ResNet18} &  0.055 & 0.991 & 0.017\\
        & {AlexNet} & 0.019   & 0.031 &0.967 \\
         \hline
        \multirow{3}{*}{{SparseFool~\cite{modas2018sparsefool}}} & {ResNet50}  &  0.990 & 0.167 & 0.176 \\
        & {ResNet18} & 0.086 & 0.997 & 0.134 \\
        & {AlexNet} & 0.062 & 0.079  & 1.000 \\
         \hline
        \multirow{3}{*}{{SemanticAdv~\cite{hosseini2018semantic}}} & {ResNet50}  &  0.890 & 0.540 &0.770 \\
        & {ResNet18} & 0.422 & 0.931 & 0.757 \\
        & {AlexNet}  & 0.359 &0.431  &0.994 \\
         \hline
        \multirow{3}{*}{{ColorFool~\cite{shamsabadi2019colorfool}}} & {ResNet50} & 0.917 & 0.346 &0.592  \\
         & {ResNet18} &  0.223 & 0.934 & 0.541\\
        & {AlexNet} &  0.114 & 0.147 & 0.995\\
        \hline
        \multirow{3}{*}{EdgeFool~\cite{shamsabadi2019edgefool}} & {ResNet50} & 0.981 & 0.357 & 0.512 \\
         & {ResNet18} & 0.278 & 0.989 & 0.510 \\
        & {AlexNet} & 0.272 & 0.333 &0.995  \\
        \hline
        \multirow{3}{*}{FilterFool (LT)} & {ResNet50} & 1.000 & 0.168 & 0.175 \\
         & {ResNet18} & 0.146 & 1.000 &0.191 \\
        & {AlexNet} &  0.154 & 0.205 & 1.000 \\
         \hline
        \multirow{3}{*}{FilterFool (ND)} & {ResNet50} & 1.000& 0.407 & 0.523  \\
         & {ResNet18} & 0.331 & 1.000 & 0.532 \\
        & {AlexNet} & 0.324 & 0.403 & 1.000  \\ 
        \hline
        \multirow{3}{*}{FilterFool (GC)} & {ResNet50} & 0.999 & 0.292 &0.402 \\
         & {ResNet18} & 0.237 & 1.000 & 0.414 \\
        \multicolumn{1}{|r|}{0.5} & {AlexNet} & 0.237 & 0.303 & 1.000  \\
        \hline
        \multirow{3}{*}{FilterFool (LD)} & {ResNet50} & 1.000 & 0.303 & 0.402 \\
         & {ResNet18} &  0.245 & 1.000 &0.402 \\
        \multicolumn{1}{|r|}{1.0} & {AlexNet} & 0.237 & 0.316 &1.000  \\
        \hline
        \Xhline{3\arrayrulewidth} 
        %
    \end{tabular}
    \label{tab:TR-PC}
\end{table}



\begin{figure}[t]
          \centering
          \begin{tikzpicture}
          \begin{axis}[
                  footnotesize,
                  axis lines=left, 
                  width=4cm,
                  height=5cm,
                  cycle list name=color list,
                  ymax=1.01,
                  ymin=0.0,
                  xmin=0.5,
                  xmax=1.01,
                  xlabel= {\footnotesize SSIM},
                  xtick={.5,.6,.7,.8,.9,1},
                  xticklabels={.5,.6,.7,.8,.9,1},
                  ytick={0,.1,.2,.3,.4,.5,.6,.7,.8,.9,1},
                  yticklabels={0,.1,.2,.3,.4,.5,.6,.7,.8,.9,1},
                  enlarge y limits=0.01,
                  enlarge x limits=0.01,
                  ylabel={\footnotesize Categorical success rate},
                  ylabel shift=-5pt
                  ]

\addplot[
    scatter/classes={ bim={orange}, df={americanrose}, sf={chocolate(traditional)}, cf={violet(web)}, sa={viola}, ef={darkgreen}, nd={blue}, log={yellow}, gc={green}, ld={red}},
    scatter,
    only marks,
    mark=*,
    scatter src=explicit symbolic,
    ]
    plot [error bars/.cd, x dir = both, x explicit, y dir = both, y explicit]
    table [meta=Class, x=x,y=y,x error=ex,y error=ey] {
  y      x     ey    ex      Class    
1.00 	0.95    0	 0.03      bim
0.983	1.00    0	 0.01      df
0.990	0.98	0	 0.03      sf
0.917	0.78    0	 0.19	   cf
0.890	0.77	0	 0.19      sa
0.981	0.73    0	 0.08      ef
1.000	0.73	0	 0.07      nd
1.000	0.97	0	 0.02      log
1.000	0.84    0	 0.04      ld
0.999	0.64    0	 0.13      gc
};
\addplot[
    scatter/classes={nd={blue}, log={yellow}, gc={green}, ld={red}},
    scatter,
    only marks,
    mark=square*,
    scatter src=explicit symbolic,
    ]
    plot [error bars/.cd, x dir = both, x explicit, y dir = both, y explicit]
    table [meta=Class, x=x,y=y,x error=ex,y error=ey] {
  y      x     ey    ex      Class    
0.2963	0.71688	   0	 0.07453      nd
0.0620	0.96974    0	 0.02395      log
0.1920	0.84024    0	 0.03762      ld
0.1667	0.64790    0	 0.13098      gc
};
\end{axis}
\end{tikzpicture}\hspace*{-0.1em}
\begin{tikzpicture}
    \begin{axis}[
                  footnotesize,
                  width=4cm,
                  height=5cm,
                  cycle list name=color list,
                  ymax=1.01,
                  ymin=0.0,
                  xmin=0.0,
                  xmax=1.01,
                  xlabel= {\footnotesize $l_{\infty}$},
                 xtick={0,.2,.4,.6,.8,1},
                 xticklabels={0,.2,.4,.6,.8,1},
                  yticklabels={},
                  ytick=\empty,
                  y axis line style={draw opacity=0},
                  enlarge y limits=0.01,
                  enlarge x limits=0.01,
                  axis x line=bottom
                  ]
            
            \addplot[
                scatter/classes={ bim={orange}, df={americanrose}, sf={chocolate(traditional)}, cf={violet(web)}, sa={viola}, ef={darkgreen}, nd={blue}, log={yellow}, gc={green}, ld={red}},
                scatter,
                only marks,
                mark=*,
                scatter src=explicit symbolic,
                ]
                plot [error bars/.cd, x dir = both, x explicit, y dir = both, y explicit]
                table [meta=Class, x=x,y=y,x error=ex,y error=ey] {
              y      x     ey    ex      Class    
            1.00 	0.03137    0	 0.00      bim
            0.983	0.08059    0	 0.25801      df
            0.990	0.87496	   0	 0.21105      sf
            0.917	0.46583    0	 0.32433	  cf
            0.890	0.63967	   0	 0.24213      sa
            0.981	0.29663    0	 0.03731      ef
            1.000	0.33030	   0	 0.03270      nd
            1.000	0.18312	   0	 0.06179      log
            1.000	0.41466    0	 0.06150      ld
            0.999	0.31991    0	 0.04154      gc
            };
            \addplot[
                scatter/classes={nd={blue}, log={yellow}, gc={green}, ld={red}},
                scatter,
                only marks,
                mark=square*,
                scatter src=explicit symbolic,
                ]
                plot [error bars/.cd, x dir = both, x explicit, y dir = both, y explicit]
                table [meta=Class, x=x,y=y,x error=ex,y error=ey] {
              y      x     ey    ex      Class    
            0.2963	0.29841	   0	 0.03512      nd
            0.0620	0.10528    0	 0.05669      log
            0.1920	0.43186    0	 0.06586      ld
            0.1667	0.25098    0	 0.03492      gc
            };
    \end{axis}
\end{tikzpicture}\hspace*{-0.1em}
\begin{tikzpicture}
      \begin{axis}[
                  footnotesize,
                  width=4cm,
                  height=5cm,
                  cycle list name=color list,
                  ymax=1.01,
                  ymin=0.0,
                 xmin=0.0,
                 xmax=0.1,
                  xlabel= {\footnotesize $l_{2}$},
                 xtick={0,.02,.04,.06,.08,.10},
                 xticklabels={0,.02,.04,.06,.08,.10},
                  yticklabels={},
                  ytick=\empty,
                  y axis line style={draw opacity=0},
                  enlarge y limits=0.01,
                  enlarge x limits=0.01,
                  scaled x ticks = false,
                  axis x line=bottom
                  ]

\addplot[
    scatter/classes={ bim={orange}, df={americanrose}, sf={chocolate(traditional)}, cf={violet(web)}, sa={viola}, ef={darkgreen}, nd={blue}, log={yellow}, gc={green}, ld={red}},
    scatter,
    only marks,
    mark=*,
    scatter src=explicit symbolic,
    ]
    plot [error bars/.cd, x dir = both, x explicit, y dir = both, y explicit]
    table [meta=Class, x=x,y=y,x error=ex,y error=ey] {
  y      x     ey    ex      Class    
1.00 	0.00019    0	 0.00002      bim
0.983	0.00003    0	 0.00035      df
0.990	0.00040	   0	 0.00056      sf
0.917	0.04246    0	 0.05469	   cf
0.890	0.03470	   0	 0.04582      sa
0.981	0.00737    0	 0.00223      ef
1.000	0.00823	   0	 0.00661      nd
1.000	0.00488	   0	 0.00547      log
1.000	0.00623    0	 0.00305      ld
0.999	0.03571    0	 0.01217      gc
};
\addplot[
    scatter/classes={nd={blue}, log={yellow}, gc={green}, ld={red}},
    scatter,
    only marks,
    mark=square*,
    scatter src=explicit symbolic,
    ]
    plot [error bars/.cd, x dir = both, x explicit, y dir = both, y explicit]
    table [meta=Class, x=x,y=y,x error=ex,y error=ey] {
  y      x     ey    ex      Class    
0.2963	0.00862	   0	 0.00230      nd
0.0620	0.00512	   0	 0.00572      log
0.1920	0.00638    0	 0.00314      ld
0.1667	0.03492    0	 0.01034      gc
};

\end{axis}
\end{tikzpicture}
\caption{{SSIM, $l_2$ and $l_{\infty}$ values and categorical success rate of BIM\protect\tikz \protect\fill[orange] (0.5ex,0.5ex) circle (0.5ex);, DeepFool\protect\tikz \protect\fill[americanrose] (0.5ex,0.5ex) circle (0.5ex);, SparseFool\protect\tikz \protect\fill[chocolate(traditional)] (0.5ex,0.5ex) circle (0.5ex);, SemanticAdv\protect\tikz \protect\fill[viola] (0.5ex,0.5ex) circle (0.5ex);, ColorFool\protect\tikz \protect\fill[violet(web)] (0.5ex,0.5ex) circle (0.5ex);, EdgeFool\protect\tikz \protect\fill[darkgreen] (0.5ex,0.5ex) circle (0.5ex);, FilterFool (ND)\protect\tikz \protect\fill[blue] (0.5ex,0.5ex) circle (0.5ex);, FilterFool (LD1)\protect\tikz \protect\fill[red] (0.5ex,0.5ex) circle (0.5ex);, FilterFool (Log)\protect\tikz \protect\fill[yellow] (0.5ex,0.5ex) circle (0.5ex);, FilterFool (GC.5)\protect\tikz \protect\fill[green] (0.5ex,0.5ex) circle (0.5ex);, Traditional filter (GC.5)\protect\tikz \protect\fill[green] (0,0) rectangle (0.12cm,0.12cm);, Traditional filter (ND)\protect\tikz \protect\fill[blue] (0,0) rectangle (0.12cm,0.12cm);, Traditional filter (LD1)\protect\tikz \protect\fill[red] (0,0) rectangle (0.12cm,0.12cm); and Traditional filter (Log)\protect\tikz \protect\fill[yellow] (0,0) rectangle (0.12cm,0.12cm);. }}
\label{fig:q-sr}
\end{figure}
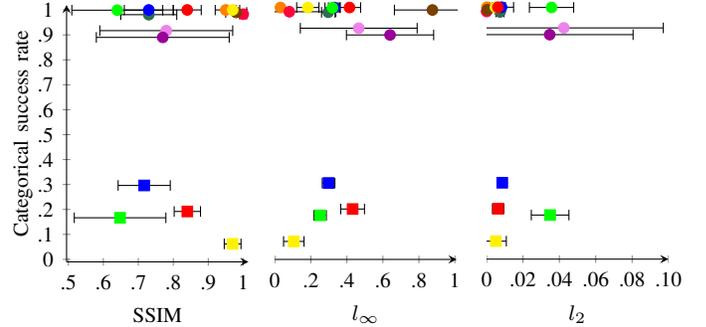

\begin{table}[h]
    \centering
    \caption{Semantic success rate and transferability on ImageNet. }
    %
    \setlength{\tabcolsep}{3pt}
   \begin{tabular}{|l|l|ccc|}
        \Xhline{3\arrayrulewidth}
        
        Attack & Classifier & $\rightarrow$ResNet50& $\rightarrow$ResNet18 & $\rightarrow$AlexNet\\
        \Xhline{3\arrayrulewidth} 
        \multirow{3}{*}{{BIM~\cite{kurakin2016adversarial}}} & {ResNet50} & 0.405&0.052&0.015  \\
         & {ResNet18} & 0.045&0.425&0.019 \\
         & {AlexNet} &0.015&0.021&0.480 \\
         \hline
         \multirow{3}{*}{{LL-FGSM~\cite{kurakin2016adversarial}}} & {ResNet50} & {0.526} & {0.244} & {0.211}  \\
         & {ResNet18} & {0.217} & {0.727} & {0.225} \\
         & {AlexNet} & {0.122} & {0.179} & {0.823}  \\
         \hline
         \multirow{3}{*}{{LL-BIM~\cite{kurakin2016adversarial}}} & {ResNet50} & {0.969} & {0.112} & {0.078}  \\
         & {ResNet18} & {0.095} & {0.975} & {0.088} \\
         & {AlexNet} & {0.074} & {0.085} & {0.973}  \\
         \hline
         \multirow{3}{*}{{P-FGSM~\cite{Li2019}}} & {ResNet50} & {0.898} & {0.103} & {0.083} \\
         & {ResNet18} & {0.095} & {0.890} & {0.084}\\
         & AlexNet & 0.070 & 0.086 & 0.897  \\
         \hline
        \multirow{3}{*}{{DeepFool~\cite{MoosaviDezfooli16}}} & {ResNet50} &0.358&0.030&0.009  \\
        & {ResNet18} &  0.023&0.384&0.008\\
        & {AlexNet} &0.006&0.012&0.413 \\
         \hline
        \multirow{3}{*}{{SparseFool~\cite{modas2018sparsefool}}} & {ResNet50}  &  0.390 &0.073 & 0.087 \\
        & {ResNet18} & 0.034 &0.408 & 0.065 \\
        & {AlexNet} &0.024 & 0.030 & 0.441 \\
         \hline
        \multirow{3}{*}{{SemanticAdv~\cite{hosseini2018semantic}}} & {ResNet50}  & 0.406 &0.274 &0.496 \\
        & {ResNet18} & 0.198 & 0.442 &0.472 \\
        & {AlexNet} &0.169 &0.209 &0.563 \\
         \hline
        \multirow{3}{*}{{ColorFool~\cite{shamsabadi2019colorfool}}} & {ResNet50} &0.400 & 0.170 & 0.380  \\
         & {ResNet18} &  0.100 & 0.426 &0.331\\
        & {AlexNet} &  0.048 & 0.062 & 0.476\\
        \hline
        \multirow{3}{*}{EdgeFool~\cite{shamsabadi2019edgefool}} & {ResNet50} &0.374 & 0.165 & 0.273 \\
         & {ResNet18} &0.126 &0.407 &0.262 \\
        & {AlexNet} &0.119 &0.149 & 0.483  \\
        \hline

        \Xhline{3\arrayrulewidth} 
        %
    \end{tabular}
    \label{tab:STR-SOA}
\end{table}

\begin{table}[!t]
    \centering
     \setlength{\tabcolsep}{3pt}
    \caption{Semantic success rate and transferability of FilterFool on ImageNet.  }
   \begin{tabular}{|l|l|ccc|
    }
        \Xhline{3\arrayrulewidth}
        
        Attack & Classifier & $\rightarrow$ResNet50& $\rightarrow$ResNet18& $\rightarrow$AlexNet\\
        \Xhline{3\arrayrulewidth} 
        \multirow{3}{*}{FilterFool (LT)} & {ResNet50} & 1.000 & 0.085 & 0.082 \\
         & {ResNet18} & 0.069 & 1.000 &0.093 \\
        & {AlexNet} & 0.065 & 0.094 & 1.000 \\
         \hline
        \multirow{3}{*}{FilterFool (ND)} & {ResNet50} & 1.000 & 0.221 & 0.287  \\
         & {ResNet18} & 0.173 & 1.000 & 0.299 \\
        & {AlexNet} & 0.148 & 0.204 &1.000  \\ 
        \hline
        \multirow{3}{*}{FilterFool (GC)} & {ResNet50} & 0.999 &0.149 & 0.206 \\
         & {ResNet18} & 0.118 & 1.000 & 0.211 \\
        \multicolumn{1}{|r|}{0.5} & {AlexNet} & 0.111 & 0.145 & 1.000  \\
        \hline
       
        \multirow{3}{*}{FilterFool (LD)} & {ResNet50} & 1.000 & 0.153 & 0.209 \\
         & {ResNet18} &  0.121 & 1.000 & 0.212 \\
        \multicolumn{1}{|r|}{1.0} & {AlexNet} & 0.103 & 0.151 & 1.000  \\
        \hline
        \hline
        
        \Xhline{3\arrayrulewidth} 
        %
    \end{tabular}
    \label{tab:STR-FF}
\end{table}

\subsection{Comparisons}

The tables in this section show the success rate on  {\em on-diagonal} elements of each sub-table and the transferability on {\em  off-diagonal} elements of each sub-table. The second column of each table shows the classifiers used to craft the adversarial images, whereas the first row shows the classifiers (with $\rightarrow$) used for testing.

Table~\ref{tab:TR-PC} reports the categorical success rate and the transferability  of several state-of-the-art adversarial attacks. Limiting the $l_p$ norm of adversarial perturbations also limits the categorical transferability of adversarial images. For example, although DeepFool adversarial images on ResNet50 achieve 98.3\% categorical success rate, only 7.1\% and 1.8\% of them are transferable to  ResNet18 and AlexNet, respectively.

\begin{figure}[t!]
    \centering
    \setlength{\tabcolsep}{1.2pt}
    \begin{tabular}{ccc}
         \footnotesize{BIM~\cite{kurakin2016adversarial}} & \footnotesize{DeepFool~\cite{MoosaviDezfooli16}} & \footnotesize{SparseFool~\cite{modas2018sparsefool}}\\
         \includegraphics[width=0.30\columnwidth,height=0.25\columnwidth]{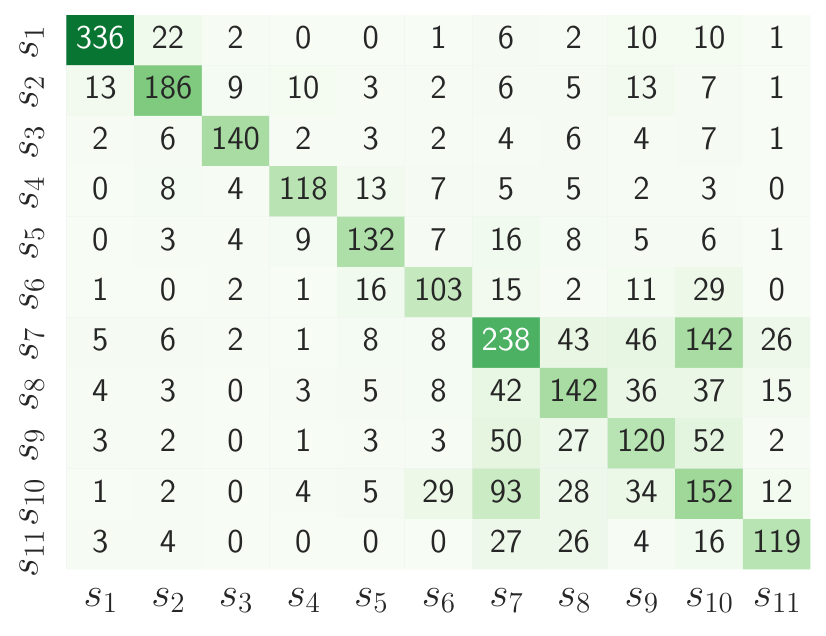}&
         \includegraphics[width=0.30\columnwidth,height=0.25\columnwidth]{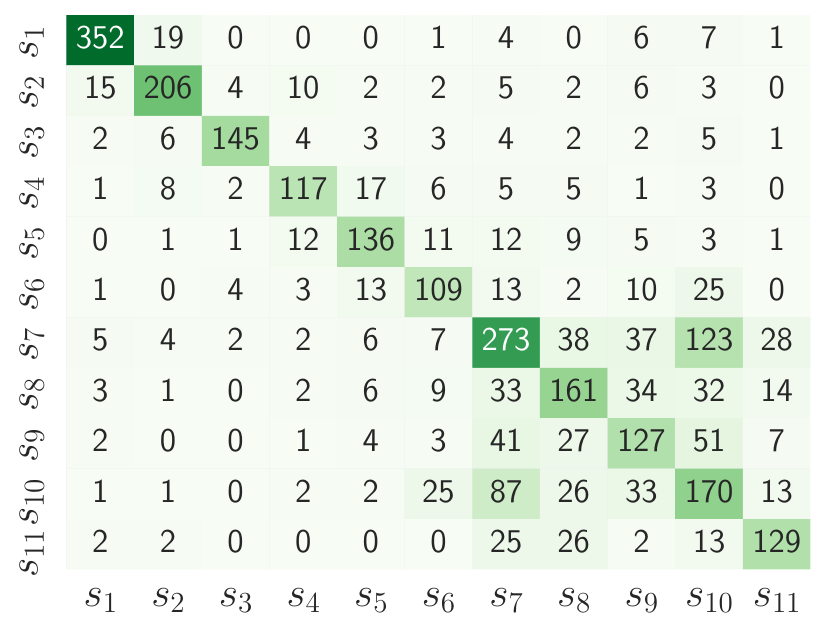}&
         \includegraphics[width=0.30\columnwidth,height=0.25\columnwidth]{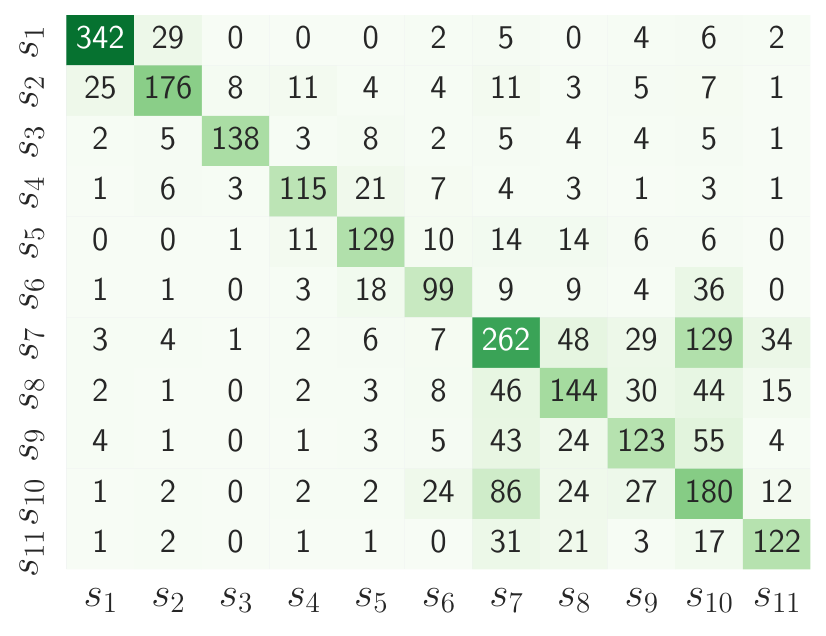}\\
        \footnotesize{{LL-BIM~\cite{kurakin2016adversarial}}}& \footnotesize{{LL-FGSM~\cite{kurakin2016adversarial}}} & \footnotesize{{P-FGSM~\cite{Li2019}}}\\
        \includegraphics[width=0.30\columnwidth,height=0.25\columnwidth]{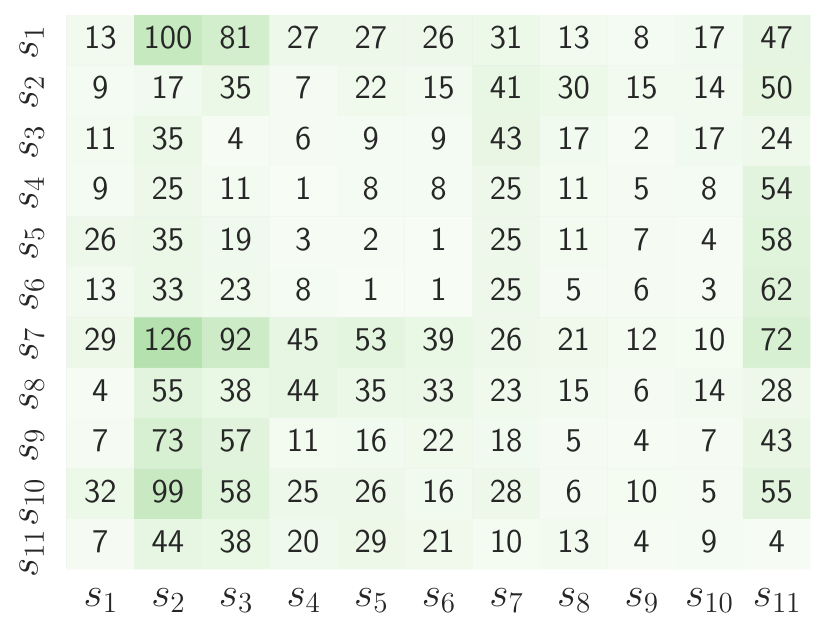}&
        \includegraphics[width=0.30\columnwidth,height=0.25\columnwidth]{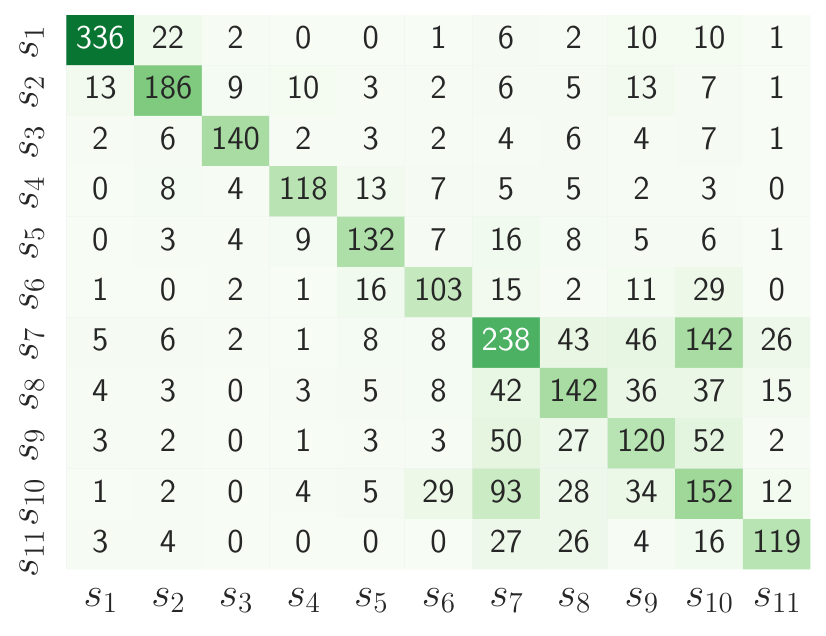}&
        \includegraphics[width=0.30\columnwidth,height=0.25\columnwidth]{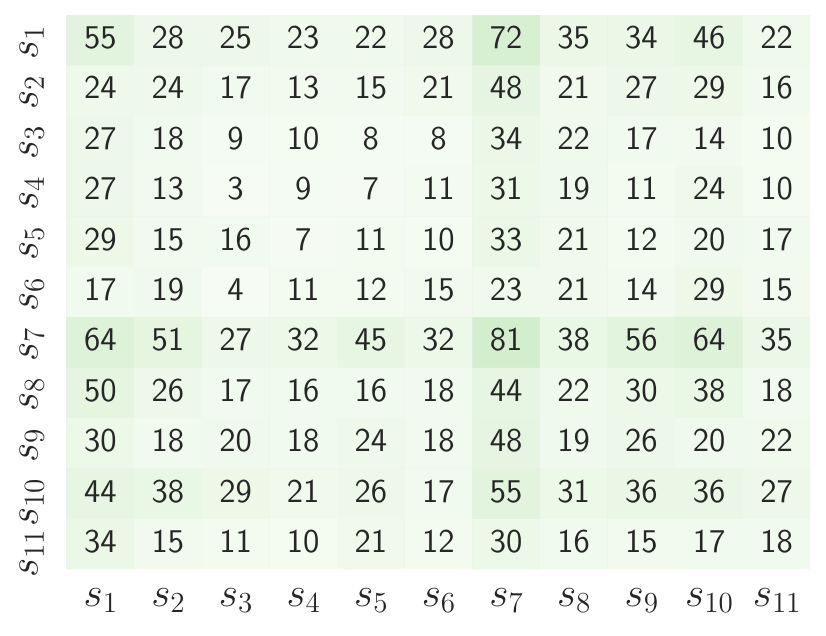}
        \\
         \footnotesize{EdgeFool~\cite{shamsabadi2019edgefool}} & \footnotesize{SemanticAdv~\cite{hosseini2018semantic}} & \footnotesize{ColorFool~\cite{shamsabadi2019colorfool}} \\
        \includegraphics[width=0.30\columnwidth,height=0.25\columnwidth]{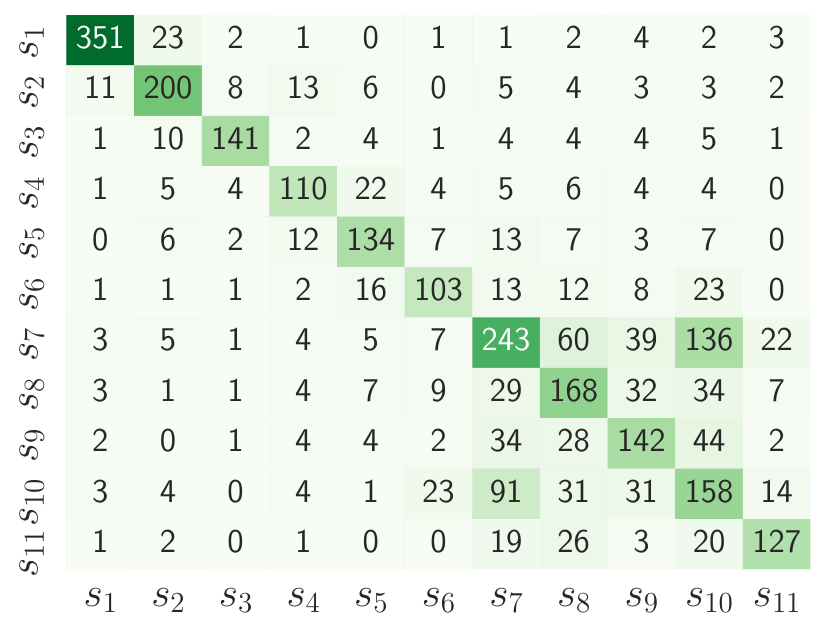}&
        \includegraphics[width=0.30\columnwidth,height=0.25\columnwidth]{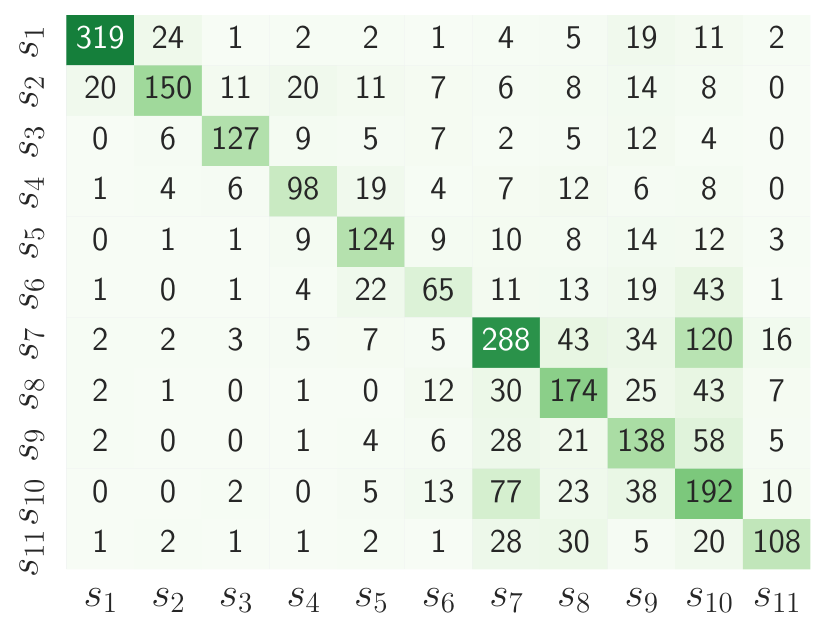}&
        \includegraphics[width=0.30\columnwidth,height=0.25\columnwidth]{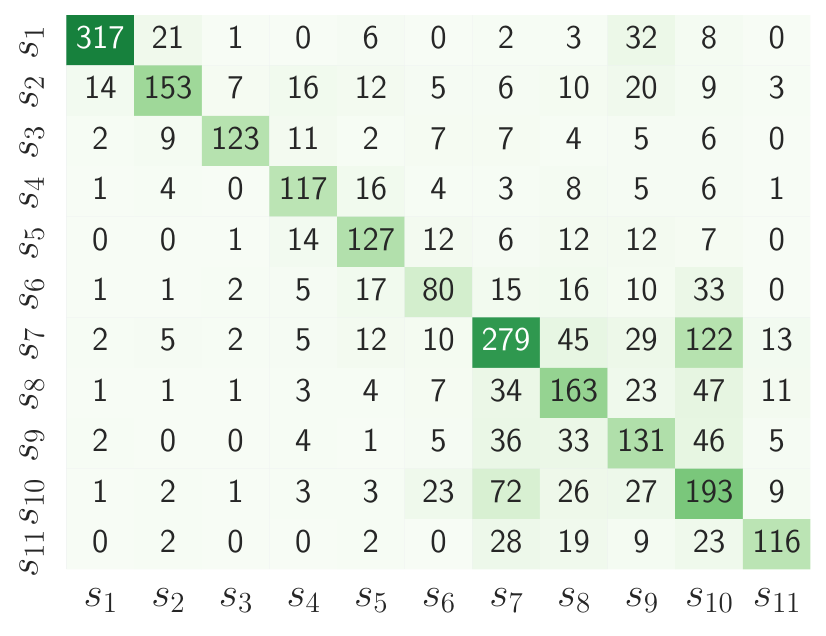}\\
        \footnotesize{Trad. filter (ND)} &  \footnotesize{FilterFool-c (ND)}& \footnotesize{FilterFool (ND)} \\
        \includegraphics[width=0.30\columnwidth,height=0.25\columnwidth]{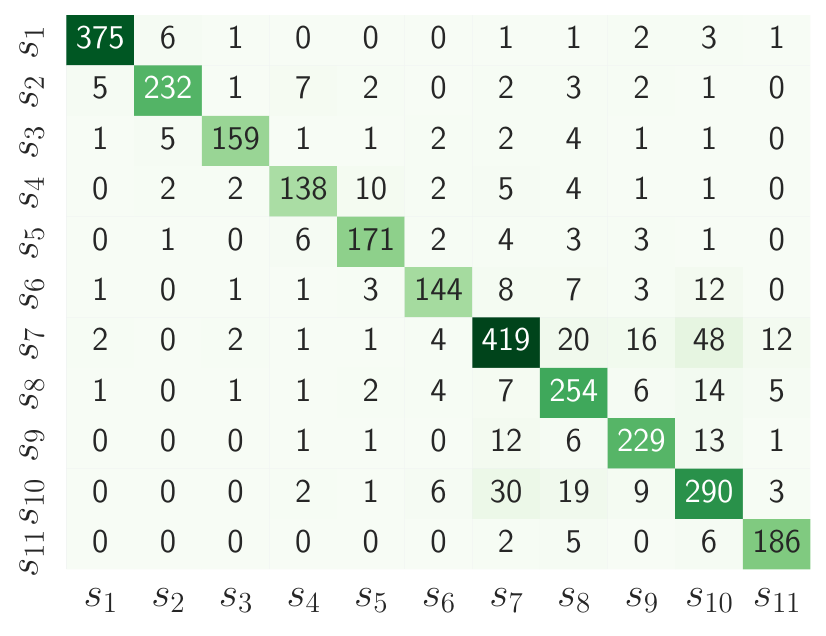}&
         \includegraphics[width=0.30\columnwidth,height=0.25\columnwidth]{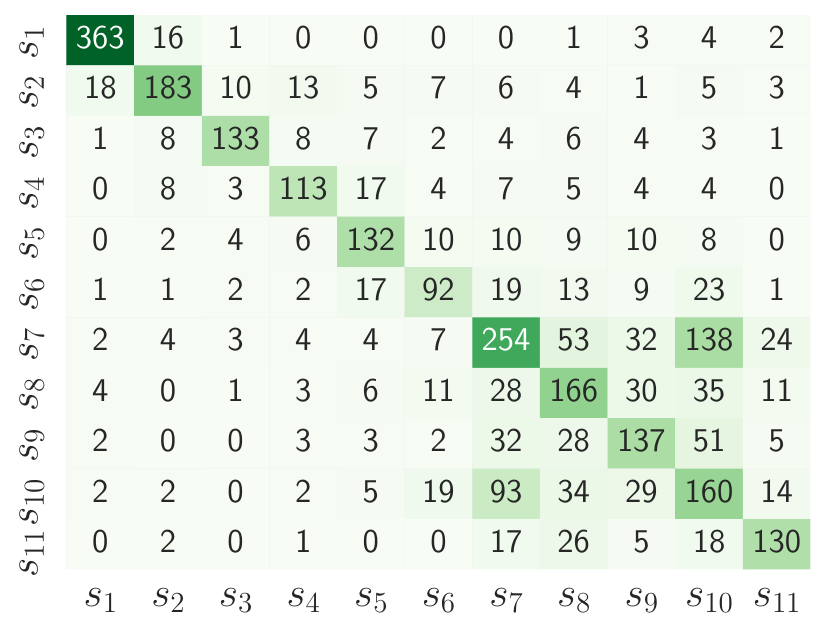}&
         \includegraphics[width=0.30\columnwidth,height=0.25\columnwidth]{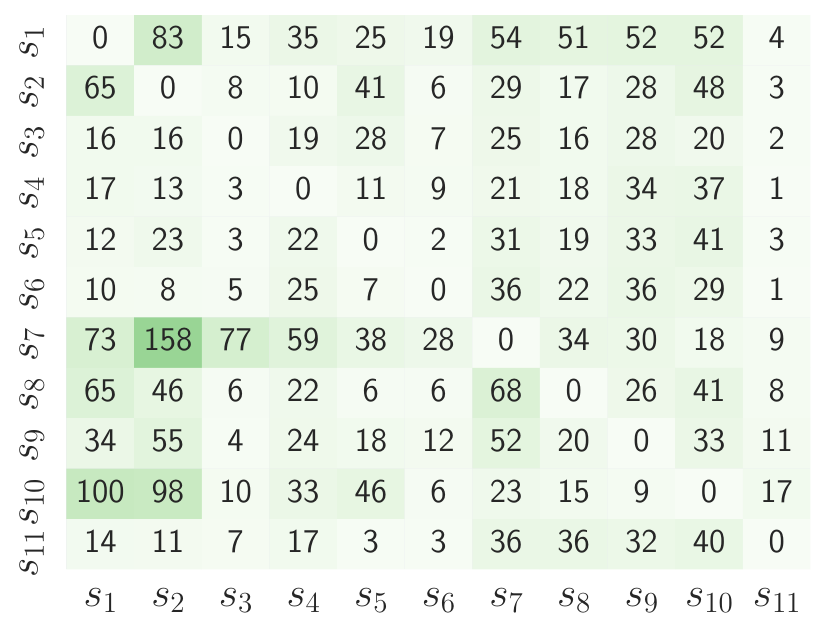}\\

    \end{tabular}
    \caption{Confusion matrices showing the success rate of adversarial attacks per semantic class against ResNet50. Rows and columns show the semantic class of the original and adversarial images, namely Dogs ($s_1$), other mammals ($s_2$), Bird ($s_3$), Reptiles, fish, amphibians ($s_4$), Invertebrates ($s_5$), Food, plants, fungi ($s_6$), Devices ($s_7$), Structures, furnishing ($s_8$), Clothes, covering ($s_9$), Implements, containers, misc. objects ($s_{10}$), Vehicles ($s_{11}$) generated by BIM, {LL-BIM, LL-FGSM, P-FGSM}, DeepFool, SparseFool, EdgeFool, SemanticAdv, ColorFool, Nonlinear Detail enhancement (ND) of traditional {Trad.} filter, ND of FilterFool-c, and ND of FilterFool. {The on-diagonal elements of the matrices show that ResNet50 is still able to classify adversarial images of state-of-the-art adversarial attacks, Filter and FilterFool-c with semantically similar classes to the classes of the original images.} }
    \label{fig:VisConfusionMatrix}
    \vspace{-9pt}
    
\end{figure}
\begin{figure}[t!]
\subfloat{
\begin{tikzpicture}
\begin{axis}[cycle list name=color list,
                  footnotesize,
                  axis lines=left, 
                  width=5cm,
                  height=4cm,
                  bar width=0.12cm,
                  ybar stacked,
                  ymin=0.00,
                  ymax=1.05,
                  title=\footnotesize ResNet50,
                  enlarge x limits=0.04,
                  ylabel={\footnotesize Semantic damage},
                  ytick={0,.2,.4,.6,.8, 1},
                  yticklabels={0,.2,.4,.6,.8, 1},
                  symbolic x coords={BI, l-BI, l-FG, P-BI, DF, SF, EF, SA, CF, FF-c, FF},
                  xtick=data,
                  x axis line style={draw=none},
                  x tick label style={rotate=45,anchor=east},
                  ]

\addplot[fill=darkgray,draw=darkgray] coordinates {(BI,0.1850) (l-BI,0.2533) (l-FG,0.1547) (P-BI,0.2200) (DF,0.1797) (SF,0.1800)(EF,0.1757) (SA,0.1773)(CF,0.1938)(FF-c,0.1707)(FF,0.2136)};\label{tr3}
\addplot[fill=gray,draw=gray]           coordinates {(BI,0.1763) (l-BI,0.4797) (l-FG,0.2026) (P-BI,0.4590) (DF,0.1633) (SF,0.1643)(EF,0.1706) (SA,0.1827)(CF,0.1851)(FF-c,0.1603)(FF,0.5149)};\label{tr5}
\addplot[fill=lightgray,,draw=lightgray] coordinates {(BI,0.2767) (l-BI,0.2423) (l-FG,0.3160) (P-BI,0.2687) (DF,0.2627) (SF,0.2824)(EF,0.2727) (SA,0.2553)(CF,0.2365)(FF-c,0.2767)(FF,0.2434)};\label{tr7}
;
\end{axis}
\end{tikzpicture}}
\subfloat{
\begin{tikzpicture}
\begin{axis}[cycle list name=color list,
                  footnotesize,
                  axis lines=left, 
                  width=5cm,
                  height=4cm,
                  bar width=0.12cm,
                  ybar stacked,
                  ymin=0.00,
                  ymax=1.05,
                  title=\footnotesize AlexNet,
                  enlarge x limits=0.04,
                  ytick={0,.2,.4,.6,.8, 1},
                  yticklabels={0,.2,.4,.6,.8, 1},
                  symbolic x coords={BI, l-BI, l-FG, P-BI,  DF, SF, EF, SA, CF, FF-c, FF},
                  xtick=data,
                  x axis line style={draw=none},
                  x tick label style={rotate=45,anchor=east},
                  ]

\addplot[fill=darkgray,draw=darkgray] coordinates {(BI,0.1840) (l-BI,0.2603) (l-FG,0.1820) (P-BI,0.2110) (DF,0.1791) (SF,0.1797)(EF,0.1797) (SA,0.1767)(CF,0.1931)(FF-c,0.1767)(FF,0.2175)};\label{tr3}
\addplot[fill=gray,draw=gray]           coordinates {(BI,0.1973) (l-BI,0.5390) (l-FG,0.3867) (P-BI,0.4517) (DF,0.168) (SF,0.1763)(EF,0.1846) (SA,0.2363)(CF,0.1898)(FF-c,0.1893)(FF,0.5884)};\label{tr5}
\addplot[fill=lightgray,,draw=lightgray] coordinates {(BI,0.3187) (l-BI,0.1844) (l-FG,0.3510) (P-BI,0.2906) (DF,0.2961) (SF,0.313)(EF,0.323) (SA,0.325)(CF,0.3122)(FF-c,0.3083)(FF,0.1886)};\label{tr7}
;
\end{axis}
\end{tikzpicture}}

\caption{Semantic damage incurred by Basic Iterative method (BI), {least-likely BI (l-BI), least-likely Fast Gradient Sign Method (l-FG), Private BI (P-BI),} DeepFool (DF), SparseFool (SF), EdgeFool (EF), SemanticAdv (SA), ColorFool (CF), Nonlinear detail enhancement of FilterFool-c (FF-c) and FilterFool (FF)  using three values for $T_s$: 0.3 \ref{tr3}, 0.5 \ref{tr5} and 0.7 \ref{tr7} against ResNet50 and AlexNet trained on the ImageNet dataset. Note that ResNet18 results are similar to ResNet50 results. For each pair of original and adversarial image, the semantic damage is 1 only when the semantic similarity, measured by the word similarity metric (Eq.~\ref{eq:SemD}),  between the adversarial class and original class is less than the chosen $T_s$.}
\label{fig:semDamage}
\vspace{-9pt}
\end{figure}
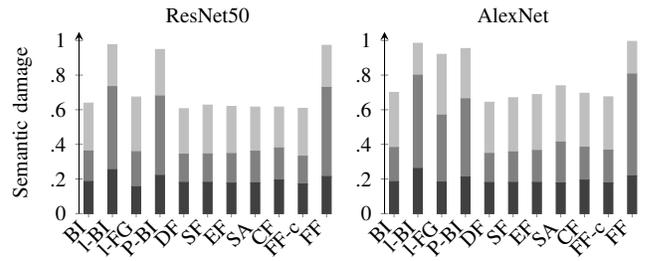

{Next, we relate  the magnitude of the perturbations to their categorical success rate. Figure~\ref{fig:q-sr} shows the categorical success rate of adversarial images with respect to their SSIM, $l_2$ and $l_{\infty}$. The $l_{\infty}$ of SparseFool is the biggest as SparseFool perturbs a few pixels but with a large magnitude. BIM changes all the pixels  controls the maximum change of each pixel, thus resulting in the smallest $l_{\infty}$.
DeepFool has the smallest $l_2$ norm, which indeed it minimises. DeepFool, SparseFool, FilterFool (Log) and BIM achieve the highest SSIM values of 1, 0.98, 0.97 and 0.95, respectively.}

{The magnitude of content-based perturbations are larger than thos of norm-bounded perturbations. The standard deviation of the magnitudes of the perturbations generated by ColorFool and SemanticAdv  are bigger than those of content-based perturbations ($5 \times$ FilterFool's), as their perturbations  are chosen randomly from a range that gradually increases until the misleading property is satisfied.}
{FilterFool and the corresponding target filter achieve similar SSIM and $l_2$ values with respect to the original images. In addition to mimicking the effect of filters, FilterFool can achieve high categorical success rate. For example, the SSIM values of both FilterFool (Log) and Traditional filter (Log) are .97, while the success rate of the former is 100\% and the latter is only 6\%. 
In general, $l_2$ and $l_{\infty}$ of FilterFool and traditional filters are larger than $l_p$ norm-bounded attacks and smaller than other content-based attacks, e.g. ColorFool and SemanticAdv.}

Table~\ref{tab:STR-SOA} shows the semantic success rate and transferability of state-of-the-art attacks. The predicted classes of more than 50\% of untargeted adversarial images generated by each state-of-the-art attack are semantically similar to the predicted classes of their corresponding original images. While the semantic success rate of LL-FGSM is low, LL-BIM is highly effective in pushing adversarial images to reach the least-likely class, thus resulting in more than $97\%$ semantic success rate, as not all the least-likely classes are semantically different than the original class.

Table~\ref{tab:STR-FF} compares the semantic success rate and transferability  of FilterFool.
In general, the semantic transferability of adversarial attacks is lower than the categorical transferability. Hence, improving the semantic transferability of adversarial attacks is an important direction for future research.

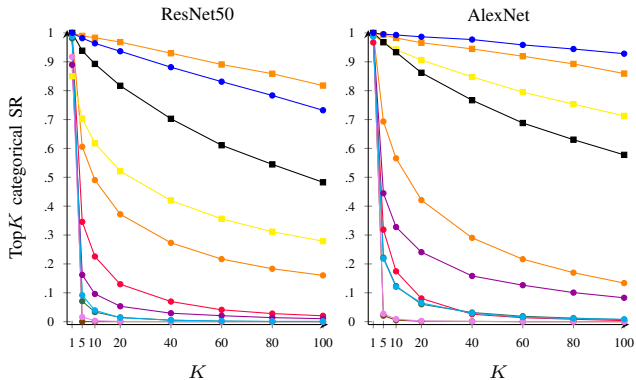
\begin{figure}[t!]
          \centering
          \subfloat{
          \begin{tikzpicture}
          \begin{axis}[cycle list name=color list,
                  footnotesize,
                  axis lines=left, 
                  width=5.05cm,
                  height=5.5cm,
                  xmin=1,
                  ymin=0.00,
                  ymax=1.00,
                  title=\scriptsize ResNet50,
                  enlarge y limits=0.01,
                  enlarge x limits=0.02,
                  xlabel={\scriptsize $K$},
                  ylabel={\scriptsize Top$K$ categorical SR},
                  y tick label style={
                  /pgf/number format/.cd,
                  fixed,
                  fixed zerofill,
                  precision=2,
                  /tikz/.cd
                  },
                  ytick={0,.1,.2,.3,.4,.5,.6,.7,.8,.9,1},
                  yticklabels={\tiny{0},\tiny{.1},\tiny{.2},\tiny{.3},\tiny{.4},\tiny{.5},\tiny{.6},\tiny{.7},\tiny{.8}, \tiny{.9}, \tiny{1}},
                  xtick={1,5,10,20,40,60,80,100},
                   xticklabels={\tiny{1},\tiny{5},\tiny{10},\tiny{20},\tiny{40},\tiny{60},\tiny{80},\tiny{100}},
                  ]
              \addplot[orange,mark=*,mark size=1] table[x index=0,y index=1]{./tikz/ImageNet/TopK/N-FGSM.txt};\label{TopK-N-FGSM}
              \addplot[yellow,mark=square*,mark size=1] table[x index=0,y index=1]{./tikz/ImageNet/TopK/FGSM.txt};\label{TopK-ll-FGSM}
               \addplot[darkorange,mark=square*,mark size=1] table[x index=0,y index=1]{./tikz/ImageNet/TopK/BIM.txt};\label{TopK-ll-BIM}
               \addplot[black,mark=square*,mark size=1] table[x index=0,y index=1]{./tikz/ImageNet/TopK/P-BIM.txt};\label{TopK-P-BIM}
              \addplot[americanrose,mark=*,mark size=1] table[x index=0,y index=1]{./tikz/ImageNet/TopK/DF.txt};\label{TopK-DF}
              \addplot[chocolate(traditional),mark=*,mark size=1] table[x index=0,y index=1]{./tikz/ImageNet/TopK/SF.txt};\label{TopK-SF}
              \addplot[darkgreen,mark=*,mark size=1] table[x index=0,y index=1]{./tikz/ImageNet/TopK/EF.txt};\label{TopK-EF}
              \addplot[viola,mark=*,mark size=1] table[x index=0,y index=1]{./tikz/ImageNet/TopK/SA.txt};\label{TopK-SA}
              \addplot[violet(web),mark=*,mark size=1] table[x index=0,y index=1]{./tikz/ImageNet/TopK/CF.txt};\label{TopK-CF}
              \addplot[cyan,mark=*,mark size=1] table[x index=0,y index=1]{./tikz/ImageNet/TopK/FFC_ND_.txt};\label{TopK-FFC}
              \addplot[blue,mark=*,mark size=1] table[x index=0,y index=1]{./tikz/ImageNet/TopK/HF_ND_.txt};\label{TopK-HF}
           \end{axis}
          \end{tikzpicture}}\hspace*{-0.9em}
          \subfloat{
          \begin{tikzpicture}
          \begin{axis}[cycle list name=color list,
                  footnotesize,
                  axis lines=left, 
                  width=5.05cm,
                  height=5.5cm,
                  xmin=1,
                  ymin=0.00,
                  ymax=1.00,
                  title=\scriptsize AlexNet,
                  enlarge y limits=0.01,
                  enlarge x limits=0.02,
                  xlabel={\scriptsize $K$},
                  y tick label style={
                  /pgf/number format/.cd,
                  fixed,
                  fixed zerofill,
                  precision=2,
                  /tikz/.cd
                  },
                    ytick={0,.1,.2,.3,.4,.5,.6,.7,.8,.9,1},
                  yticklabels={\tiny{0},\tiny{.1},\tiny{.2},\tiny{.3},\tiny{.4},\tiny{.5},\tiny{.6},\tiny{.7},\tiny{.8}, \tiny{.9}, \tiny{1}},
                  xtick={1,5,10,20,40,60,80,100},
                   xticklabels={\tiny{1},\tiny{5},\tiny{10},\tiny{20},\tiny{40},\tiny{60},\tiny{80},\tiny{100}},
                  ]
              \addplot[orange,mark=*,mark size=1] table[x index=0,y index=3]{./tikz/ImageNet/TopK/N-FGSM.txt};
              \addplot[yellow,mark=square*,mark size=1] table[x index=0,y index=3]{./tikz/ImageNet/TopK/FGSM.txt};
               \addplot[darkorange,mark=square*,mark size=1] table[x index=0,y index=3]{./tikz/ImageNet/TopK/BIM.txt};
                \addplot[black,mark=square*,mark size=1] table[x index=0,y index=3]{./tikz/ImageNet/TopK/P-BIM.txt};
              \addplot[americanrose,mark=*,mark size=1] table[x index=0,y index=3]{./tikz/ImageNet/TopK/DF.txt};
              \addplot[chocolate(traditional),mark=*,mark size=1] table[x index=0,y index=3]{./tikz/ImageNet/TopK/SF.txt};
              \addplot[darkgreen,mark=*,mark size=1] table[x index=0,y index=3]{./tikz/ImageNet/TopK/EF.txt};
              \addplot[viola,mark=*,mark size=1] table[x index=0,y index=3]{./tikz/ImageNet/TopK/SA.txt};
              \addplot[violet(web),mark=*,mark size=1] table[x index=0,y index=3]{./tikz/ImageNet/TopK/CF.txt};
              \addplot[cyan,mark=*,mark size=1] table[x index=0,y index=3]{./tikz/ImageNet/TopK/FFC_ND_.txt};
              \addplot[blue,mark=*,mark size=1] table[x index=0,y index=3]{./tikz/ImageNet/TopK/HF_ND_.txt};
           \end{axis}
          \end{tikzpicture}}
           \caption{Top$K$ categorical success rate (SR) of BIM~\ref{TopK-N-FGSM}, {LL-BIM~\ref{TopK-ll-BIM}, LL-FGSM~\ref{TopK-ll-FGSM}, P-FGSM~\ref{TopK-P-BIM}}, DeepFool~\ref{TopK-DF}, SparseFool~\ref{TopK-SF}, EdgeFool~\ref{TopK-EF}, SemanticAdv~\ref{TopK-SA}, ColorFool~\ref{TopK-CF}, Nonlinear detail enhancement of FilterFool-c~\ref{TopK-FFC} and FilterFool~\ref{TopK-HF} against ResNet50 and AlexNet trained on ImageNet.  ResNet18 results are similar to ResNet50 results.}
            \label{fig:TopKSR}
            \vspace{-9pt}
\end{figure}	
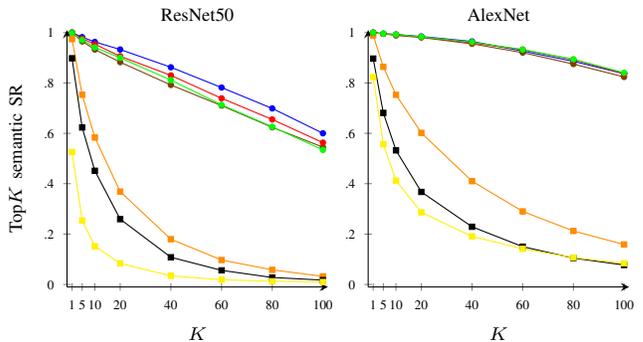
\begin{figure}[h]
          \centering
          \subfloat{
          \begin{tikzpicture}
          \begin{axis}[cycle list name=color list,
                  footnotesize,
                  axis lines=left, 
                  width=5.05cm,
                  height=5cm,
                  xmin=1,
                  ymin=0.00,
                  ymax=1.00,
                  title=\scriptsize ResNet50,
                  enlarge y limits=0.01,
                  enlarge x limits=0.02,
                  xlabel={\scriptsize $K$},
                  ylabel={\scriptsize Top$K$ semantic SR},
                  y tick label style={
                  /pgf/number format/.cd,
                  fixed,
                  fixed zerofill,
                  precision=2,
                  /tikz/.cd
                  },
                  ytick={0,.2,.4,.6,.8, 1},
                  yticklabels={\tiny{0},\tiny{.2},\tiny{.4},\tiny{.6},\tiny{.8}, \tiny{1}},
                  xtick={1,5,10,20,40,60,80,100},
                   xticklabels={\tiny{1},\tiny{5},\tiny{10},\tiny{20},\tiny{40},\tiny{60},\tiny{80},\tiny{100}},
                  ]
              \addplot[darkorange,mark=square*,mark size=1] table[x index=0,y index=1]{./tikz/ImageNet/TopKsemantic/BIM.txt};\label{TopkSSR-BIM}
              \addplot[black,mark=square*,mark size=1] table[x index=0,y index=1]{./tikz/ImageNet/TopKsemantic/P-BIM.txt};\label{TopkSSR-P-BIM}
              \addplot[yellow,mark=square*,mark size=1] table[x index=0,y index=1]{./tikz/ImageNet/TopKsemantic/FGSM.txt};\label{TopkSSR-FG}
              \addplot[blue,mark=*,mark size=1] table[x index=0,y index=1]{./tikz/ImageNet/TopKsemantic/Nonlinear_Detail.txt};\label{TopkSSR-ND}
              \addplot[chocolate(traditional),mark=*,mark size=1] table[x index=0,y index=1]{./tikz/ImageNet/TopKsemantic/Log.txt};\label{TopkSSR-L}
            \addplot[red,mark=*,mark size=1] table[x index=0,y index=1]{./tikz/ImageNet/TopKsemantic/Linear_Detail1.txt};\label{TopkSSR-LD1}
            \addplot[green,mark=*,mark size=1] table[x index=0,y index=1]{./tikz/ImageNet/TopKsemantic/Gamma05.txt};\label{TopkSSR-G05}
           \end{axis}
          \end{tikzpicture}}\hspace*{-0.9em}
          \subfloat{
          \begin{tikzpicture}
          \begin{axis}[cycle list name=color list,
                  footnotesize,
                  axis lines=left, 
                  width=5.05cm,
                  height=5cm,
                  xmin=1,
                  ymin=0.00,
                  ymax=1.00,
                  title=\scriptsize AlexNet,
                  enlarge y limits=0.01,
                  enlarge x limits=0.02,
                  xlabel={\scriptsize $K$},
                  y tick label style={
                  /pgf/number format/.cd,
                  fixed,
                  fixed zerofill,
                  precision=2,
                  /tikz/.cd
                  },
                  ytick={0,.2,.4,.6,.8, 1},
                  yticklabels={\tiny{0},\tiny{.2},\tiny{.4},\tiny{.6},\tiny{.8}, \tiny{1}},
                  xtick={1,5,10,20,40,60,80,100},
                   xticklabels={\tiny{1},\tiny{5},\tiny{10},\tiny{20},\tiny{40},\tiny{60},\tiny{80},\tiny{100}},
                  ]
               \addplot[darkorange,mark=square*,mark size=1] table[x index=0,y index=3]{./tikz/ImageNet/TopKsemantic/BIM.txt};
                \addplot[black,mark=square*,mark size=1] table[x index=0,y index=3]{./tikz/ImageNet/TopKsemantic/P-BIM.txt};
              \addplot[yellow,mark=square*,mark size=1] table[x index=0,y index=3]{./tikz/ImageNet/TopKsemantic/FGSM.txt};
              \addplot[blue,mark=*,mark size=1] table[x index=0,y index=3]{./tikz/ImageNet/TopKsemantic/Nonlinear_Detail.txt};
              \addplot[chocolate(traditional),mark=*,mark size=1] table[x index=0,y index=3]{./tikz/ImageNet/TopKsemantic/Log.txt};
            \addplot[red,mark=*,mark size=1] table[x index=0,y index=3]{./tikz/ImageNet/TopKsemantic/Linear_Detail1.txt};
            \addplot[green,mark=*,mark size=1] table[x index=0,y index=3]{./tikz/ImageNet/TopKsemantic/Gamma05.txt};
           \end{axis}
          \end{tikzpicture}}
           \caption{Top$K$ semantic success rate (SR) of least-likely FGSM~\ref{TopkSSR-FG}, least-likely BIM~\ref{TopkSSR-BIM} and P-FGSM~\ref{TopkSSR-P-BIM} with Log transformation~~\ref{TopkSSR-L}, Gamma correction ($\gamma=0.5$~\ref{TopkSSR-G05}), nonlinear detail enhancement~~\ref{TopkSSR-ND} and linear detail enhancement ($\alpha=1$~~\ref{TopkSSR-LD1}) of FilterFool using ResNet50 and AlexNet trained on ImageNet.  ResNet18 results are similar to ResNet50 results. }
            \label{fig:TopKSemSR}
            \vspace{-9pt}
\end{figure}

\begin{table}[h]
\centering
\caption{Top1 and Top5 classification accuracy ($\downarrow$) with respect to the categorical labels. ResNet50, ResNet18 and AlexNet are evaluated on the original and adversarial images.}
\resizebox{\columnwidth}{!}{
\begin{tabular}{|l|cc|cc|cc|}
    \Xhline{3\arrayrulewidth}
\multirow{2}{*}{Images} & \multicolumn{2}{c|}{ResNet50}& \multicolumn{2}{c|}{ResNet18} & \multicolumn{2}{c|}{AlexNet}\\
               & Top1 & Top5                 & Top1 & Top5                  & Top1 & Top5                 \\
    
    \Xhline{3\arrayrulewidth} 
       Original                                    & 0.726&0.906& 0.650&0.868& 0.517&0.753\\
       \hline
       BIM~\cite{kurakin2016adversarial}        & 0.082&0.482&0.091&0.388&0.094&0.431    \\
       {LL-FGSM~\cite{kurakin2016adversarial}}        & {0.149}& {0.297} & {0.035} & {0.090} & {0.010} & {0.035}    \\
       {LL-BIM~\cite{kurakin2016adversarial}}        & {0.001}& {0.010} & {0.001} & {0.006} & {0.003} & {0.006}    \\
       {P-FGSM~\cite{Li2019}}        & 0.000 & 0.062 & 0.000 & 0.044 & 0.001 & 0.027   \\
       DeepFool~\cite{MoosaviDezfooli16}        & 0.115&0.693&0.109&0.618&0.140&0.641    \\
       SparseFool~\cite{modas2018sparsefool}    & 0.097&0.899&0.097&0.843&0.106&0.716  \\                        
       SemanticAdv~\cite{hosseini2018semantic}  & 0.156&0.760&0.117&0.705&0.057&0.452 \\
       ColorFool~\cite{shamsabadi2019colorfool} & 0.160&0.872&0.141&0.827&0.106&0.707 \\
       EdgeFool~\cite{shamsabadi2019edgefool}   & 0.085&0.840& 0.079&0.784&0.073&0.613 \\
       FilterFool-c {(ND)} & 0.085&0.814&0.086&0.767& 0.084&0.607\\
       FilterFool {(ND)} & 0.006&0.026&0.004&0.024&0.004&0.012\\
    \Xhline{3\arrayrulewidth} 

    \end{tabular}
    }
    \label{tab:Acc}
\end{table}

Figure~\ref{fig:VisConfusionMatrix} visualises the confusion matrix of the semantic success rate for original and adversarial classes. It is possible to note, for example, that 336, 352, 342, 351, 319, 317 of the adversarial images generated by BIM, DeepFool, SparseFool, EdgeFool, SemanticAdv and ColorFool, respectively, for (original) 390 dog images are misclassified as another breed of dogs.  Note also that the 50\% semantic success rates of the state-of-the-art attacks reported in Table~\ref{tab:TR-PC} are due to confusion between devices and containers. 
Moreover, we  observe the effect of the adversarial loss and the semantic adversarial loss by comparing the confusion matrix of the traditional filter, FilterFool-c, FilterFool with nonlinear detail enhancement. The results of traditional filter mostly falls on the diagonal because of the low success rate.

Figure~\ref{fig:semDamage} compares the {\em semantic damage} for three values of the similarity threshold.  
FilterFool, least-likely and private attacks inflict greater semantic damage on  classifiers than other {untargeted} attacks as {their} adversarial labels are semantically different  from the original labels.

\begin{table*}[ht!]
\centering
\caption{Robustness ($\uparrow$) of images generated with various attacks: BIM, LL-BIM, LL-FGSM, P-FGSM, DeepFool (DF), SparseFool (SF), SemanticAdv (SA), ColorFool (CF) and various version of FilterFool (FF) -- Linear with $\alpha=1$ and Nonlinear Detail enhancement (LD) and (ND), Log Transformation (LT) and Gamma Correction with $\gamma=0.5$ (GC) -- on ImageNet for ResNet50 (R50), ResNet18 (R18) and AlexNet (A). }
\label{tab:R-PC}
\begin{tabular}{|l|c|ccccccc|ccc|cccc|}
\Xhline{3\arrayrulewidth}
    
       \multirow{2}{*}{Method} & \multirow{2}{*}{Model} & \multicolumn{7}{c|}{Bit reduction}& \multicolumn{3}{c|}{Median smoothing}& \multicolumn{4}{c|}{JPEG compression}\\
       \multicolumn{1}{|c|}{} &  & \multicolumn{1}{c}{1-bit}&  \multicolumn{1}{c}{2-bit}&  \multicolumn{1}{c}{3-bit}&  \multicolumn{1}{c}{4-bit}&  \multicolumn{1}{c}{5-bit}&  \multicolumn{1}{c}{6-bit}&  \multicolumn{1}{c|}{7-bit}&
      \multicolumn{1}{c}{$2 \times 2$}& \multicolumn{1}{c}{${3 \times 3}$}& \multicolumn{1}{c|}{${5 \times 5}$}& \multicolumn{1}{c}{{q-25}}&\multicolumn{1}{c}{{q-50}}&\multicolumn{1}{c}{{q-75}}&\multicolumn{1}{c|}{{q-100}}\\
      \Xhline{3\arrayrulewidth}
     \multirow{3}{*}{BIM~\cite{kurakin2016adversarial}}  & R50 & 0.933 & 0.622 & 0.527 & 0.687 & 0.809 & 0.859 & 0.870  & 0.509 & 0.465 & 0.462 & 0.381 & 0.402 & 0.504 & 0.837   \\
      & R18 & 0.951  & 0.702 & 0.697 & 0.832 & 0.911 & 0.935 & 0.945 & 0.663 & 0.591 & 0.553 & 0.466 & 0.533 & 0.686 & 0.936   \\
      & A    & 0.927 & 0.780 & 0.784 & 0.880 & 0.924 & 0.939 & 0.943  & 0.649 & 0.656 & 0.663 & 0.569 & 0.653 & 0.771 & 0.927   \\
     \cline{1-16}
      \multirow{3}{*}{{LL-BIM~\cite{kurakin2016adversarial}}}  & {R50} & {0.939}&{0.648}&{0.613}&{0.952}&{0.999}&{1.000}&{1.000}&{0.568}&{0.468}&{0.477}&{0.370}&{0.375}&{0.453}&{0.997}   \\
      &{R18}&{0.952}&{0.760}&{0.798}&{0.991}&{1.000}&{1.000}&{1.000}&0.687&0.579&{0.564}&{0.459}&{0.473}&{0.667}&1.000   \\
      &A&0.946&0.905&0.962&0.995&0.998&0.998&0.998&0.806&0.830&0.751&0.598&0.747&0.918&0.995   \\
    \cline{1-16}
      \multirow{3}{*}{P-FGSM~\cite{Li2019}}  &R50& 0.938&0.636&0.601&0.972&1.000&1.000&1.000&0.559&0.443&0.462&0.368&0.342&0.429&0.998 \\
      &R18& 0.953&0.731&0.818&0.995&1.000&1.000&1.000&0.706&0.563&0.529&0.432&0.434&0.694&1.000\\
      &A& 0.942&0.873&0.973&0.999&1.000&1.000&1.000&0.814&0.801&0.722&0.573&0.771&0.960&1.000\\
     \cline{1-16}
      \multirow{3}{*}{LL-FGSM~\cite{kurakin2016adversarial}}  &R50&0.947&0.760&0.790&0.830&0.846&0.850&0.850&0.719&0.688&0.650&0.573&0.682&0.758&0.839  \\
      &R18&0.965&0.897&0.930&0.961&0.965&0.965&0.966&0.864&0.846&0.781&0.754&0.867&0.927&0.961 \\
      &A&0.960&0.959&0.982&0.988&0.992&0.992&0.992&0.943&0.943&0.896&0.912&0.959&0.979&0.989   \\
     \cline{1-16}
     \multirow{3}{*}{DF~\cite{MoosaviDezfooli16}} & R50 &0.937 &0.625 &0.371 &0.329 &0.345 &0.357 &0.362 &0.304 &0.342 &0.412 &0.348 &0.377 &0.362 &0.364   \\
      & R18 &0.953 &0.683 &0.465 &0.455 &0.481 &0.495 &0.499  &0.388 &0.399 &0.481 &0.426 &0.466 &0.488 &0.499   \\
      & A    &0.922 &0.733 &0.553 &0.584 &0.605 &0.614 &0.614  &0.431 &0.506 &0.602 &0.472 &0.527 &0.600 &0.600   \\
     \cline{1-16}
     \multirow{3}{*}{SF~\cite{modas2018sparsefool}}  & R50    &0.941 &0.619 &0.386 &0.346 &0.422 &0.548 &0.714 &0.180 &0.226 &0.372 &0.363 &0.312 &0.293 &0.328 \\
                                                     & R18    &0.956 &0.690 &0.455 &0.403 &0.497 &0.638 &0.786 &0.210 &0.280 &0.432 &0.423 &0.358 &0.333 &0.397 \\
                                                     & A       &0.929 &0.783 &0.650 &0.719 &0.831 &0.911 &0.959 &0.382 &0.462 &0.605 &0.502 &0.477 &0.521 &0.698 \\
     \cline{1-16}
     \multirow{3}{*}{SA~\cite{hosseini2018semantic}} & R50 &0.958 &0.809 &0.701 &0.716 &0.768 &0.830 &0.866 &0.652 &0.697 &0.783 &0.676 &0.662 &0.667 &0.737   \\
      & R18 &0.975 &0.873 &0.783 &0.787 &0.835 &0.875 &0.909 &0.715 &0.754 &0.813 &0.730 &0.712 &0.712 &0.782   \\
      & A    &0.971 &0.923 &0.903 &0.919 &0.949 &0.977 &0.987  &0.869 &0.888 &0.922 &0.840 &0.846 &0.869 &0.914   \\
     \cline{1-16}
      \multirow{3}{*}{CF~\cite{shamsabadi2019colorfool}} & R50  &0.946 &0.785 &0.654 &0.618 &0.675 &0.763 &0.831 &0.546 &0.622 &0.740 &0.664 &0.607 &0.606 &0.614 \\
                                     & R18 &0.967 &0.824 &0.686 &0.642 &0.683 &0.768 &0.843 &0.609 &0.667 &0.771 &0.684 &0.637 &0.613 &0.617\\
                                     & A  &0.951 &0.861 &0.761 &0.743 &0.808 &0.888 &0.929 &0.684 &0.740 &0.828 &0.732 &0.693 &0.703 &0.735 \\
     \cline{1-16}
    \multirow{3}{*}{EF~\cite{shamsabadi2019edgefool}} & R50  & 0.950&0.610&0.560&0.640&0.806&0.917&0.966&0.397&0.433&0.490&0.516&0.512&0.519&0.691 \\
                                     & R18 & 0.963&0.677&0.633&0.723&0.861&0.942&0.975&0.461&0.483&0.545&0.581&0.563&0.586&0.756\\
                                     & A  & 0.929&0.819&0.820&0.904&0.959&0.981&0.990&0.585&0.598&0.628&0.714&0.749&0.795&0.888 \\
     \cline{1-16}
      \multirow{3}{*}{FF (ND)} & R50  & 0.953&0.642&0.667&0.880&0.986&0.999&1.000&0.474&0.485&0.520&0.507&0.541&0.621&0.990  \\
                           & R18     & 0.967&0.735&0.802&0.954&0.996&0.999&0.999&0.567&0.560&0.591&0.592&0.631&0.774&0.998 \\
                           & A        & 0.942&0.942&0.991&0.999&1.000&1.000&1.000&0.803&0.815&0.753&0.846&0.946&0.989&1.000 \\   
     \cline{1-16}
      \multirow{3}{*}{FF (LD)} & R50  & 0.925&0.569&0.515&0.782&0.960&0.998&1.000&0.539&0.537&0.567&0.396&0.413&0.470&0.977  \\
                              & R18  & 0.949&0.657&0.668&0.890&0.987&0.999&1.000&0.633&0.637&0.654&0.472&0.536&0.663&0.994 \\
      & A     & 0.935&0.881&0.973&0.998&1.000&1.000&1.000&0.855&0.880&0.826&0.695&0.799&0.912&1.000 \\ 
      \cline{1-16}              
          
      \multirow{3}{*}{FF (LT)} & R50  & 0.926&0.592&0.413&0.616&0.902&0.992&0.999&0.437&0.428&0.501&0.357&0.330&0.363&0.965  \\
                              & R18  & 0.951&0.701&0.562&0.802&0.971&0.998&1.000&0.567&0.538&0.591&0.433&0.450&0.538&0.989 \\
                              & A     & 0.937&0.879&0.952&0.994&1.000&0.999&0.999&0.885&0.916&0.866&0.635&0.746&0.884&0.999  \\  
                              
      \cline{1-16}              
      \multirow{3}{*}{FF (GC)}      & R50  &  0.955&0.658&0.454&0.561&0.840&0.986&0.997&0.514&0.497&0.558&0.457&0.418&0.436&0.960 \\
                             & R18  &  0.968&0.728&0.579&0.729&0.946&0.995&0.998&0.656&0.637&0.651&0.516&0.514&0.572&0.985\\
    & A     &  0.969&0.887&0.928&0.994&0.999&1.000&1.000&0.902&0.915&0.875&0.728&0.794&0.898&0.999\\

\Xhline{3\arrayrulewidth}

\end{tabular}
\end{table*}

Figure~\ref{fig:TopKSR} shows the Top$K$ categorical success rate of the attacks.  
{The categorical success rate of untargeted attacks substantially decreases as $K$ increases, since untargeted attacks mostly shift the original label from the most probable one to another one with high probability. 
This results in, for instance, the decreased success rate of DeepFool from 98\% with $K=1$ to 35\% with $K=5$. The least-likely targeted attack and FilterFool maintain high success rate even for $K=100$ as a consequence of targeting semantically different classes and the proposed semantic adversarial loss function, respectively. For example, the Top$K$ categorical success rate of FilterFool only drops to 73\%, 80\% and 93\% for ResNet50, ResNet18 and AlexNet, respectively, when the $K$ changes from 1 to 100. }
{Furthermore, Figure~\ref{fig:TopKSemSR}
compares the Top$K$ semantic success rate of FilterFool and targeted attacks. 
The Top$K$ semantic success rate of FilterFool and the least-likely attacks decreases as $K$ increases. However, this drop is higher in least-likely targeted attacks than FilterFool, as FilterFool performs on a group of labels as opposed to one label considered in targeted attacks. For example, Top$1$, Top$5$ and Top$10$ semantic success rates of LL-BIM is 95\%, 75\% and 58\%, while 100\%, 97\% and 95\% for FilterFool with detail enhancement ($\alpha=1$).}

Table~\ref{tab:Acc} shows the Top1 and Top5 accuracy of ResNet50, ResNet18, AlexNet on the original and adversarial images. Most of the attacks achieve low Top1 accuracy, but those of untargeted attacks with the categorical adversarial loss show high Top5 accuracy. {The targeted attacks with the least-likey label, LL-FGSM and LL-BIM, can avoid this performance decrease as the least-likely class selected for the perturbations can be semantically different from the original label. FilterFool, even with the untargeted attack, can still  mislead Top5 classes.}

Table~\ref{tab:R-PC} reports the robustness of attacks to bit reduction, median filtering and JPEG compression. Content-based attacks are more robust than norm-bounded attacks, as high-frequency norm-bounded perturbations can be easily removed by these input-based transformations. For example, the most effective parameter of bit reduction, median smoothing and JPEG compression drop the 89\% success rate of SemanticAdv against ResNet50 to 70.1\%, 65.2\% and 66.2\%, respectively. The reason is that SemanticAdv, similarly to other content-based attacks such as ColorFool, generates large, low-frequency perturbations as opposed to the high-frequency perturbations of BIM, in which the categorical success rates drop to 52.7\%, 50.9\% and 38.1\%, respectively. Some variants of the FilterFool adversarial perturbations even improve the robustness of existing content-based attacks.
Reducing the number of bits from 8 to 3 decreases the categorical success rate of adversarial images. However, the categorical success rate increases again for 2 bits and 1 bit because of the quality of the resulting images.

\begin{figure}[t!]
\centering
\begin{tikzpicture}
\begin{axis}[cycle list name=color list,
                  footnotesize,
                  axis lines=left, 
                  width=4cm,
                  height=4cm,
                  bar width=0.2cm,
                  ybar stacked,
                  ymin=0.00,
                  ymax=0.85,
                  title=\footnotesize Norm-bounded,
                  enlarge x limits=0.04,
                  ylabel={\footnotesize Success rate},
                  ytick={0,.2,.4,.6,.8},
                  yticklabels={0,.2,.4,.6,.8},
                  symbolic x coords={BIM, L-BIM, P-FG, L-FG, DF, SF},
                  xtick=data,
                  x axis line style={draw=none},
                  x tick label style={rotate=45,anchor=east},
                  ]

\addplot[fill=blue,draw=blue] coordinates {(BIM,0.010) (L-BIM,0.0173) (L-FG,0.066) (P-FG,0.0146) (DF,0.004) (SF,0.117)};
\end{axis}
\end{tikzpicture}
\begin{tikzpicture}
\begin{axis}[cycle list name=color list,
                  footnotesize,
                  axis lines=left, 
                  width=5cm,
                  height=4cm,
                  bar width=0.2cm,
                  ybar stacked,
                  ymin=0.00,
                  ymax=0.85,
                  title=\footnotesize Content-based,
                  enlarge x limits=0.04,
                  ylabel={\footnotesize Success rate},
                  ytick={0,.2,.4,.6,.8},
                  yticklabels={0,.2,.4,.6,.8},
                  symbolic x coords={EF, SA, CF, FF-LT, FF-ND, FF-L1, FF-G.5 },
                  xtick=data,
                  x axis line style={draw=none},
                  x tick label style={rotate=45,anchor=east},
                  ]
\addplot[fill=blue,draw=blue] coordinates {(EF,0.423) (SA,0.584)(CF,0.54)(FF-LT,0.145)(FF-ND,0.4239)(FF-L1,0.326)(FF-G.5,0.455)};
\end{axis}
\end{tikzpicture}

\vspace{-9pt}
\caption{Robustness of norm-bounded attacks -- BIM, least-likely BIM (L-BIM), least-likely FGSM (L-FG), Private FGSM (P-FG), DeepFool (DF), SparseFool (SF) -- and content-based attacks -- EdgeFool (EF), SemanticAdv (SA), ColorFool (CF), FilterFool with Log Transformation (FF-LT), Nonlinear detail enhancement (FF-ND), $\alpha=1$ Linear Detail Enhancement  (FF-LD1), $\gamma=.5$ Gamma correction (FF-GC.5) -- whose adversarial images were generated on ResNet50 against ResNet50 adversarially 
re-trained with BIM images.}
\label{fig:AdvRob}
\vspace{-9pt}
\end{figure}
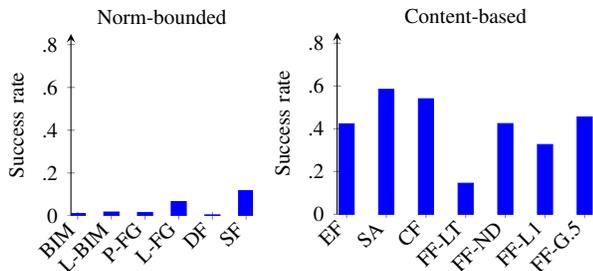
Finally, Figure~\ref{fig:AdvRob} shows the categorical success rate of adversarial images against {\em adversarial training}, when each training mini-batch is augmented with adversarial examples to improve the robustness of the classifier. We use the adversarially re-trained ResNet50 with BIM images~\cite{robustness}, which 
{reduces the success rate of all adversarial attacks. The success  of content-based adversarial images is higher than that of norm-bounded ones  as content-based perturbations generally have larger magnitudes. For example, the success rate of ColorFool, SemanticAdv and FilterFool (Gamma Correction) is above 50\%. However, the success rate of norm-bounded adversarial perturbations generated by DeepFool and FGSM variants substantially  decreases  to below 20\%. }
%
\section{Conclusion}

We proposed FilterFool, an adversarial framework that crafts adversarial perturbations based on the content of an image and on the pre-defined semantics of its label. FilterFool is flexible and can incorporate different image filters to generate various types of adversarial perturbations. These perturbations are larger than those produced by norm-bounded methods, thereby improving the transferability of the attacks to unseen classifiers and their robustness against defences. While in this work we considered the grouping of the labels according to the semantic relationship defined by WordNet~\cite{tsipras2020imagenet}, the framework is general and other groupings can be considered. 
As future work, we will extend FilterFool to cope with multi-label images.

\section*{Acknowledgments}
The authors thank Dmitrii Mukhutdinov and Ashish Alex for their help in developing the online subjective evaluation tool.

\bibliographystyle{IEEEbib}
\bibliography{refs.bib}

\end{document}